\documentclass{article}

\PassOptionsToPackage{numbers, compress}{natbib}

\usepackage[final]{neurips_2022}

\usepackage[flushleft]{threeparttable}
\usepackage{mathtools}

\usepackage{caption}
\usepackage{subcaption}

\usepackage{booktabs}

\newcommand{\para}[1]{\smallskip\noindent{\bf{#1}.}}

\usepackage{amsmath}
\usepackage{amssymb}
\usepackage{mathtools}
\usepackage{amsthm}

\usepackage{graphicx}
\usepackage{textcomp}
\usepackage{multirow}
\usepackage{float}
\usepackage{algorithm}
\usepackage{algorithmicx}
\usepackage[noend]{algpseudocode}

\usepackage{tikz}
\usetikzlibrary{backgrounds}
\usetikzlibrary{arrows,shapes}
\usetikzlibrary{tikzmark}
\usetikzlibrary{calc}
\usepackage{nccmath}
\usepackage{wrapfig}
\usepackage{blindtext}

\usepackage{array}
\usepackage{ragged2e}
\newcolumntype{P}[1]{>{\RaggedRight\hspace{0pt}}p{#1}}
\newcolumntype{X}[1]{>{\RaggedRight\hspace*{0pt}}p{#1}}
\usepackage{tcolorbox}
\usepackage{tikz}
\usetikzlibrary{arrows,shapes,positioning,shadows,trees,mindmap}
\usepackage[edges]{forest}
\usetikzlibrary{arrows.meta}
\colorlet{linecol}{black!75}
\usepackage{xkcdcolors} %
\usepackage{tikz}
\usetikzlibrary{backgrounds}
\usetikzlibrary{arrows,shapes}
\usetikzlibrary{tikzmark}
\usetikzlibrary{calc}
\newcommand{\highlight}[2]{\colorbox{#1!17}{$\displaystyle #2$}}

\definecolor{bittersweet}{rgb}{1.0, 0.44, 0.37}

\renewcommand{\highlight}[2]{\colorbox{#1!17}{#2}}

\usepackage{bbding}

 \usepackage{url}

\usepackage{lipsum,tabularx}

\usepackage{multicol}
\usepackage{multirow}

\usepackage{colortbl}
\usepackage{booktabs}
\usepackage{setspace}

\usepackage[hidelinks,breaklinks,colorlinks]{hyperref}
\hypersetup{
  linkcolor={blue!70!black},
  citecolor={red!70!black},
  urlcolor={blue!70!black}
}

\def\Snospace~{\S{}}

\usepackage{amsmath, amsthm, amssymb}
\theoremstyle{plain}
\newtheorem{theorem}{Theorem}[section]

\theoremstyle{definition}
\newtheorem{definition}[theorem]{Definition}

\theoremstyle{remark}

\usepackage[T1]{fontenc}

\usepackage{balance}

\usepackage{bm}
\usepackage{fp}
\usepackage{siunitx}
\usepackage{amsthm}

\usepackage[labelfont=bf,font=small,skip=5pt]{caption}
\usepackage{subcaption}

\usepackage{comment}

\usepackage{xspace}

\usepackage[super]{nth}

\usepackage{soul}
\usepackage{xcolor}

\algnewcommand\algorithmicinput{\textbf{Input:}}
\algnewcommand\Input{\item[\algorithmicinput]}
\algnewcommand\algorithmicoutput{\textbf{Output:}}
\algnewcommand\Output{\item[\algorithmicoutput]}
\algnewcommand{\LineComment}[1]{\State \(\triangleright\) #1}

\newcommand{\sys}{\mbox{\textsc{NONE}}\xspace}

\def\BibTeX{{\rm B\kern-.05em{\sc i\kern-.025em b}\kern-.08em
    T\kern-.1667em\lower.7ex\hbox{E}\kern-.125emX}}

\title{Training with More Confidence: Mitigating Injected and Natural Backdoors During Training}

\author{%
  Zhenting Wang \\
  Rutgers University \\
  \texttt{zhenting.wang@rutgers.edu}
  \And
  Hailun Ding \\
  Rutgers University \\
  \texttt{hailun.ding@rutgers.edu} \\
  \And
  Juan Zhai \\
  Rutgers University \\
  \texttt{juan.zhai@rutgers.edu} \\
  \And
  Shiqing Ma \\
  Rutgers University \\
  \texttt{sm2283@rutgers.edu} \\
}

\begin{document}

\maketitle

\begin{abstract}

    The backdoor or Trojan attack is a severe threat to deep neural networks (DNNs).
    Researchers find that DNNs trained on benign data and settings can also learn backdoor behaviors, which is known as the natural backdoor.
    Existing works on anti-backdoor learning are based on weak observations that the backdoor and benign behaviors can differentiate during training. An adaptive attack with slow poisoning can bypass such defenses.
    Moreover, these methods cannot defend natural backdoors.
    We found the fundamental differences between backdoor-related neurons and benign neurons: backdoor-related neurons form a hyperplane as the classification surface across input domains of all affected labels.
    By further analyzing the training process and model architectures, we found that piece-wise linear functions cause this hyperplane surface.
    In this paper, we design a novel training method that forces the training to avoid generating such hyperplanes and thus remove the injected backdoors.
    Our extensive experiments on five datasets against five state-of-the-art attacks and also benign training show that our method can outperform existing state-of-the-art defenses.
    On average, the ASR (attack success rate) of the models trained with \sys is 54.83 times lower than undefended models under standard poisoning backdoor attack and 1.75 times lower under the natural backdoor attack.
    Our code is available at \url{https://github.com/RU-System-Software-and-Security/NONE}.

\end{abstract}

\section{Introduction}\label{sec:intro}

Deep Neural Networks (DNNs) are vulnerable to Trojans\footnote{Trojan attack is also known as the backdoor attack in the existing literature.
}.
A Trojaned model makes normal predictions on benign inputs and outputs the target label when the input contains a specific pattern (i.e., Trojan trigger) such as a yellow pad.
To inject a Trojan~\cite{gu2017badnets,jia2021badencoder,carlini2021poisoning,wang2021backdoorl,bagdasaryan2021spinning,chen2021badnl,xie2019dba}, the adversary can poison the training dataset by adding poisoning samples (or Trojan samples): inputs stamped with the Trojan trigger and labeled as the target label.
This is a typical data poisoning attack, and the model can learn the trigger as a strong feature for the target label.
Recently, researchers found the existence of \textit{natural Trojans}.
Namely, a model trained on benign datasets with normal settings (e.g., hyperparameters, optimizers) can also learn Trojans, when there exists a strong input pattern in the training dataset that corresponds to one label~\cite{liu2019abs}.
In such natural Trojan scenarios, the input pattern serves as a Trojan trigger, and its corresponding label is the target label.
By reverse-engineering the trigger, the adversary can leverage it for attacks.
As such, both injected and natural Trojans are severe threats.

There are no existing works for learning a robust DNN against both injected and natural Trojans.
Existing works focus on training benign classifiers when the dataset is poisoned.
For example, ABL~\cite{li2021anti} observes that the model will learn backdoor behavior faster than benign behavior, and proposes a training algorithm to suppress learning the trigger pattern.
DP-SGD is an optimization method that leverages the differential privacy (DP) method and combines it with stochastic gradient descent (SGD) to learn a robust classifier using poisoned datasets.
These methods fail to defend against the natural Trojan.

\begin{wrapfigure}{R}{0.4\textwidth}
		\centering
		\footnotesize
		\includegraphics[width=.4\columnwidth]{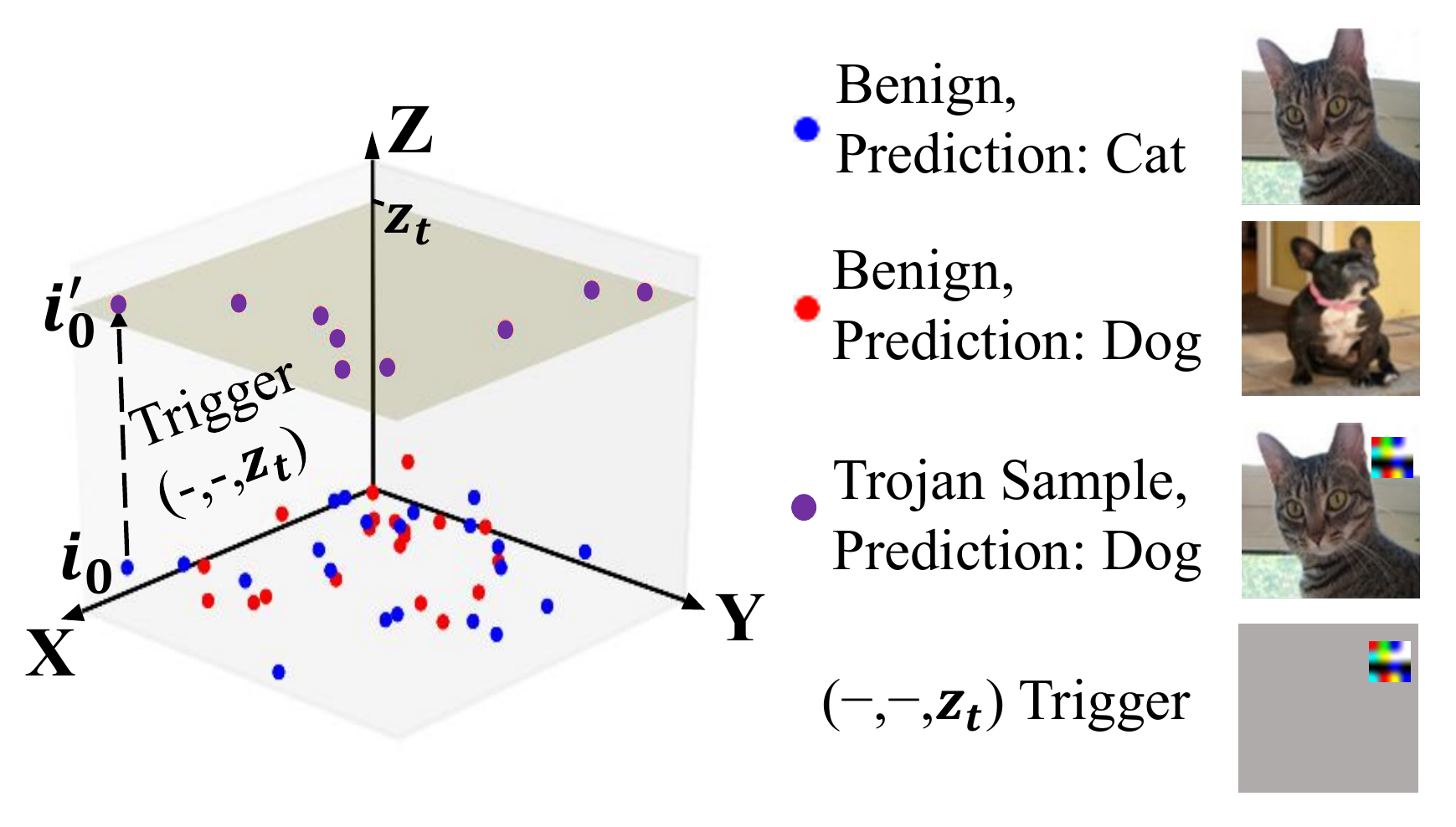}
		\caption{Decision Region of A Trojan Model.}\label{fig:decision_boundary}
\end{wrapfigure}

In this paper, we propose a robust training algorithm that can mitigate both injected and natural Trojans.
For a given Trojan in the form of \( \tilde{\bm{x}} = T(\bm{x}, {\bm m}, {\bm t}) = ({\bm 1}-{\bm m}) {\odot} \bm{x} + {\bm m} {\odot} \bm{t} \), where \(\tilde{\bm{x}}\) and \(\bm{x}\) are respectively the poisoning and benign samples, \(\tilde{\bm{x}}\) is the poisoning sample generation method using the trigger \((\bm{m}, \bm{t})\) with trigger mask matrix \(\bm{m}\) and trigger value matrix \(\bm{t}\), we theoretically prove that there exists one and only one hyperplane in the input space that corresponds to all poisoning samples.
Thus, the trained classifier is mapping this hyperplane to a target label when performing Trojan attacks.
\autoref{fig:decision_boundary} intuitively illustrates our idea using a simple example.
To simplify the problem, each input has three dimensions \((d_x, d_y, d_z)\).
We use red and blue dots to denote inputs belonging to different labels, and the trigger is denoted as \(t = (-, -, z_t)\).
Adding the trigger into an input \( i_0 = (x_0, y_0, z_0) \) to get the corresponding Trojaned input \( i^{\prime}_0 = (x_0, y_0, z_t) \) is equivalent to moving this input to the \(z = z_t\) plane in the input domain.
As shown in \autoref{fig:decision_boundary}, the input \(i_0\) moves along the dashed line and ends up in the wheat colored plane, \(z = z_t\).
Notice that \(i_0\) can be any input, and stamping the trigger will move them to the plane \(z = z_t\).
In other words, the plane \(z = z_t\) contains all Trojan samples.
Likewise, if there exists a plane \(z = z_t\) that is a decision region, its corresponding input pattern \((-, -, z_t)\) is a Trojan trigger.
Stamping such a trigger to any input is essentially projecting the input to this plane, and because all the inputs in this plane have the same label, it is equivalent to performing a Trojan attack.
Extending this to the high dimensional space, the Trojan region will be a hyperplane, and the decision boundaries it shares with other regions will be linear.
In summary, we can say that a Trojan in a DNN always pairs with a hyperplane as its Trojan region.
Considering that modern DNNs are non-linear and non-convex, this rarely happens for benign models.
By further analysis, we found that this is related to the use of activation functions.
Modern DNNs tend to use piece-wise linear functions as their activation functions.
Even though the function itself is linear, its sub-functions are linear.
For example, one of the most popular activation function, ReLU (i.e, \(y = \max(0, x)\)), consists of two linear functions (i.e., \( y = 0, x \leq 0 \) and \( y = x, x > 0 \)).
When a model's weights and biases are trained to specific regions, the neuron values before activation functions will fall into the input domain of only one sub-function (e.g., \(x > 0\)).
As a result, the output and input will form a linear relationship.
Consequently, the model can generate a hyperplane decision region in the input domain.
In other words, we have a hyperplane, denoted as \(<x_0, x_1, x_2> = <a_0, a_1, a_2>\), as a decision region, and for a given input \(i\), if we replace its values in dimensions \(x_0\), \(x_1\) and \(x_2\) to \(a_0\), \(a_1\) and \(a_2\), respectively, we can turn its output label to a desired one.
A model training process includes randomness (e.g., random initiation values, optimization), which we cannot avoid. 
Many possible decision regions will give us the same or similar training/validation accuracy.
Some training will learn a linear decision region while others will not.
This explains the cause of DNN Trojans and answers the former question, which moves forward our understanding of DNN Trojans one more step.

Based on this analysis, we develop a revised training method, \sys{} (\textbf{NON}-Lin\textbf{E}arity) that identifies linear decision regions, filters out inputs that are potentially poisoned, and resets affected neurons to enforce non-linear decision regions during training.
We evaluated our prototype built with Python and PyTorch on MNIST, GTSRB, CIFAR, ImageNet, and the TrojAI dataset.
Compared with SOTA methods, \sys{} is more effective and efficient in mitigating different Trojan attacks (i.e., single-target attack, label-specific attack, label-consistent attack, natural Trojan attack, and the hidden trigger attack).
For example, on average, the ASR (attack success rate) of the models trained with \sys is 54.83 times lower than undefended models under standard poisoning Trojan attack and 1.75 times lower under the natural Trojan attack.

Our contributions in this paper can be summered as follows.
We analyze the cause of (injected and natural) DNN Trojans and conclude that linearity in DNN decision regions is the main reason.
Further, we analyzed the source of linear DNN decision regions and explained why and when it happens using commonly used layers and activation functions.
Then, we propose a novel and general revised training framework, \sys{} that detects and fixes injected and natural Trojans in DNN training.
To the best of our knowledge, we are the first to defend natural Trojans.
We evaluate \sys{} on five different datasets and five different Trojans and compare it with other SOTA techniques.
Results show that \sys{} significantly outperforms these prior works in practice.

\section{Related Work}\label{sec:bg}

\para{Trojans in DNN}
Prior work~\cite{gu2017badnets,chen2017targeted,turner2019label,salem2022dynamic,nguyen2020input,tang2021demon,jia2021badencoder,carlini2021poisoning,saha2021backdoor,li2021hidden, cheng2021deep,wang2022bppattack} demonstrates that the attackers can inject Trojans into the victim models by poisoning the training dataset.
Later, researchers found that a DNN trained on the benign dataset with standard training procedure can also learn such Trojans, which is known as natural Trojans.
By using reverse engineering methods designed for poisoned models, researchers were able to find natural triggers in pretrained models.
For example, in ABS~\cite{liu2019abs}, authors show that a Network in Network (NiN)~\cite{lin2013network} model trained on a benign CIFAR-10 dataset has natural Trojans.
Other works~\cite{tang2021demon, zhao2021deep} also documented similar findings in other models.

\para{Trojan Defense}
One way to defend against Trojan attacks is to filter out poisoning data before training~\cite{tran2018spectral,chen2018detecting}.
Poison suppression defenses~\cite{du2019robust,hong2020effectiveness} restrain the malicious effects of poisoning samples in the training phase.
Du et al.~\cite{du2019robust} and Hong et al.~\cite{hong2020effectiveness} apply DP-SGD~\cite{abadi2016deep} to depress the malicious gradients brought from poisoned samples.
However, existing anti-backdoor learning methods are based on weak observations and they do not find the root cause of Trojans.
They can be bypassed by adaptive attacks.
In addition, they can not defend natural Trojans.
For example, the SOTA anti-backdoor learning method ABL~\cite{li2021anti} is based on the observation that learning of backdoor behaviors and benign behaviors are distinctive.
In detail, the training loss on backdoor examples drops much faster than that on benign examples in the first few epochs.
It can be bypassed by slow poisoning attacks.
More specifically, when the backdoor attack is label-specific (i.e., the samples with different original labels have different target labels) with low poisoning ratios, the model will learn backdoor behavior slower.
We run ABL on the label-specific BadNets attack~\cite{gu2017badnets} with ResNet18 model and GTSRB dataset. The results show that the attack success rate is 93.93\%, meaning that ABL can not defend against such an attack.
This is not surprising because the label-specific attack is more complex, and it is based on the benign behavior of the models (i.e., the model needs to classify the original label of the backdoor examples first, then convert it to corresponding target labels).
Another line of work tries to detect if a model has Trojan or not before its deployment.
Model diagnosis based defenses~\cite{wang2019neural,xu2019detecting,kolouri2020universal,huang2020one,guo2019tabor,shen2021backdoor,liu2022ex,tao2022better,liu2022piccolo,shen2022constrained} determine if a given model has a Trojan or not by inspecting the model behavior.
Model reconstruction based defenses try to eliminate injected Trojans in infected models~\cite{liu2018fine,zhao2020bridging,li2021neural,wu2021adversarial,zeng2021adversarial}, which requires retraining the model with a set of benign data.
Another approach is to defend Trojan attacks at runtime.
Testing-based defenses~\cite{gao2019strip,chou2020sentinet,ma2019nic} judge if the given input contains trigger patterns and reject the ones that are malicious.

\section{Methodology}\label{sec:observation}

\para{Threat Model} In this paper, we consider training time defense against Trojan attacks, which is also used by existing works~\cite{hong2020effectiveness,chen2018detecting,tran2018spectral}.
The training dataset can contain poisoning samples (to inject intended Trojans) or benign (but leads to natural Trojans).
As defenders, we control the training process but do not assume control over training datasets. 
Adopted from existing work~\cite{wang2019neural,gao2019strip,chou2020sentinet,xu2019detecting}, Trojan sample generation can be formalized as \autoref{eq:define_trojan_sample}, where \(\bm{x}\) and \(\tilde{\bm{x}}\) respectively are the benign input and Trojan input. \(\bm{m}\), \(\bm{t}\) respectively represent, mask of the trigger (i.e., whether a pixel is in the trigger region) and contents of the trigger. \(\odot\) is the element-wise multiplication operation on two vectors, i.e., the Hadamard product.

\begin{equation}\label{eq:define_trojan_sample}
            \tilde{\bm{x}} = T(\bm{x}, \tikzmarknode{s}{\highlight{blue}{${\bm m}$}}, \tikzmarknode{x}{\highlight{blue}{${\bm t}$}}) = ({\bm 1}-{\bm m}) \tikzmarknode{ts}{\highlight{red}{$\odot$}} \bm{x} + {\bm m} \tikzmarknode{td}{\highlight{red}{$\odot$}} \bm{t}
\end{equation}
\begin{tikzpicture}[overlay,remember picture,>=stealth,nodes={align=left,inner ysep=1pt},<-]
    \path (x.north) ++ (0,0.4\baselineskip) node[anchor=south east,color=blue!67] (scalep){\textbf{\scriptsize Trigger pattern}};
    \draw [color=blue!87](x.north) |- ([xshift=-0.3ex,color=blue]scalep.south west);
    \path (s.south) ++ (0,-0.5em) node[anchor=north west,color=blue!67] (mean){\textbf{\scriptsize Trigger mask}};
    \draw [color=blue!57](s.south) |- ([xshift=-0.3ex,color=blue]mean.south east);
    \path (ts.north) ++ (-1.5em,0.5em) node[anchor=south west,color=red!67] (scalep){\textbf{\scriptsize Hadamard product}};
        \draw[<->,color=red!57] (ts.north) -- ++(0,0.2)  -| node[] {} (td.north);
\end{tikzpicture}

\subsection{Trojan Analysis}\label{sec:formalana}

To facilitate our discussion, we first define \textit{decision region} which includes all samples with the same predicted label.
Formally, we define it as:

\begin{definition}{(Decision Region)}\label{def:dec_reg}
    For a deep neural network \(\mathcal{M}: \mathcal{X}\mapsto \mathcal{Y}\) where \(\mathcal{X}\) is the input domain \(\mathbb{R}^{m}\) and \(\mathcal{Y}\) is a set of labels \(\{1 \ldots k\}\), a decision region is an input space \(\mathcal{R}^{l} \subseteq \mathcal{X}\), s.t., \(\forall \bm x \in \mathcal{R}^{l}, \mathcal{M}(\bm x) = l\).
\end{definition}

In most tasks, decision regions with the same label spread over the whole input space because natural inputs belonging to the same label are naturally distributed in this way.
Similarly, we define \textit{Trojan Decision Region}, or in short, we call it the \textit{Trojan Region}, which is a subregion of the target label decision region:

\begin{definition}{(Trojan Region)}\label{def:dec_reg}
    For a Trojaned deep neural network \(\mathcal{M}: \mathcal{X}\mapsto \mathcal{Y}\) with target label \(l\), its Trojan regions are input spaces where \(\mathcal{T} \subseteq \mathcal{R}^{l}\), s.t., all Trojan inputs \( \tilde{\bm{x}} \in \mathcal{T}\), and all inputs in \(\mathcal{T}\) are Trojan inputs.
\end{definition}

Based on the definitions, we have the following theorem:

\begin{theorem}\label{th:perfect_trojan1}
    Given a model \(\mathcal{M}: \mathcal{X} \mapsto \mathcal{Y}\) with the Trojan trigger \((\bm{m}, \bm{t})\), if the attack is complete (100\% attack success rate) and precise (no other triggers), there exists one and only one hyperplane \(\{{\bm A}{\bm x} - {\bm b} = 0\}\) Trojan region, where \(i \in \{1\ldots m\}\), diagonal matrix \({{\bm A}_{i,i}} = {\bm m}_i, \bm b = {\bm A}{\bm t}\).
\end{theorem}

The proof for \autoref{th:perfect_trojan1} and empirical results are in Appendix (\autoref{sec:proof}).
The theorem shows that when a model learns a Trojan, it essentially learns a hyperplane as a decision region.
Based on the definition of decision regions, we know that they are inverse functions of the model.
Thus, the inverse function of the Trojan is a hyperplane.
To understand \textbf{how} popular model architectures learn such hyperplanes in practice, we perform further analysis.

We start our discussion from typical Convolutional Neural Networks (CNNs).
A convolutional layer with activation functions can be represented as
\(y_j = \sigma (\mathbf{W}_{j}^\mathbf{T}x_j+\mathbf{b}^\mathbf{T}_j)\), where \(x_j\) and \(y_j\) are the inputs and outputs of layer \(j\), \(\mathbf{W}_j\) and \(\mathbf{b}_j\) are trained weights and bias values, and \(\sigma\) represents the activation function which is used to introduce non-linearity in this layer.
Most commonly used activation functions, e.g., ReLU, are piece-wise linear.
For example, ReLU is defined as \(\sigma(x) = \max(0, x)\), which consists of two linear pieces separated at \(x = 0\).
As pointed out by Goodfellow et al.~\cite{goodfellow2014explaining}, even non-piece-wise linear functions are trained to semi-piece-wise linear.
This helps resolve the gradient explosion/vanishing problem and makes training DNNs more feasible.

We make a key observation that \textit{DNN Trojans will increase linearity of a convolutional layer with activation functions by introducing a large percentage of neurons activating on one piece of the activation function}.
Specifically, we observe that when the Trojan behavior happens, the neuron values before activation functions fall into one input range of the activation function, which makes it linear.
Recall that most well-trained activation functions are piece-wise linear, and if the inputs are in one input range, it regresses to a linear function.
For example, layer \(j\) using ReLU as its activation function will be a linear layer if \(\mathbf{W}_j^\mathbf{T}x_j+\mathbf{b}^\mathbf{T}_j \geq 0 \) for all \(x_j\).
As such, the reverse function of the Trojan can be a hyperplane or overlaps significantly with the hyperplane.

\begin{wrapfigure}{R}{0.6\textwidth}

     \centering
     \footnotesize
     \begin{subfigure}[t]{0.28\columnwidth}
         \centering     
         \footnotesize
         \includegraphics[width=\columnwidth]{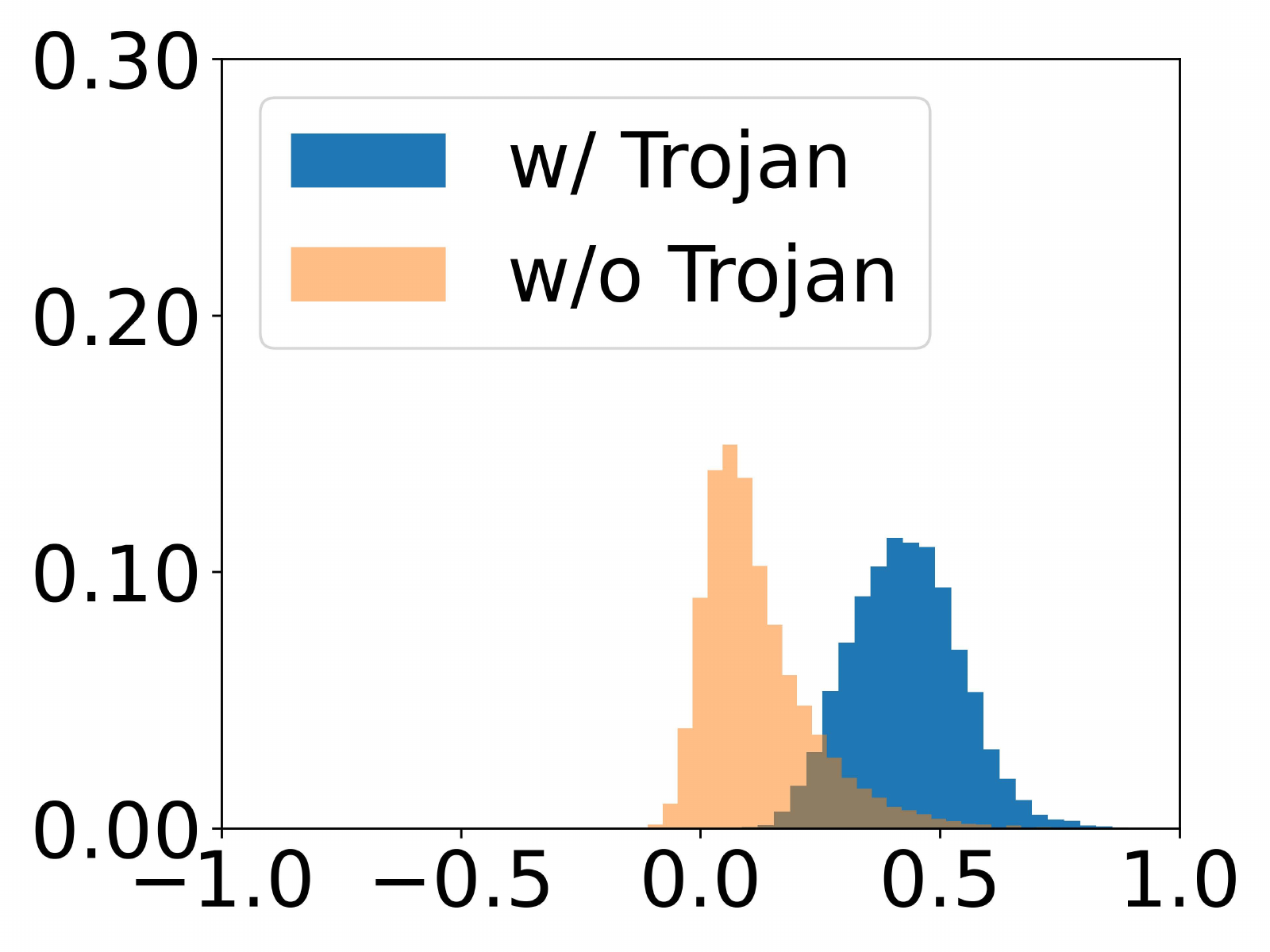}
         \caption{Compromised}\label{fig:comchannel}
     \end{subfigure}
     \begin{subfigure}[t]{0.28\columnwidth}
         \centering     
         \footnotesize
         \includegraphics[width=\columnwidth]{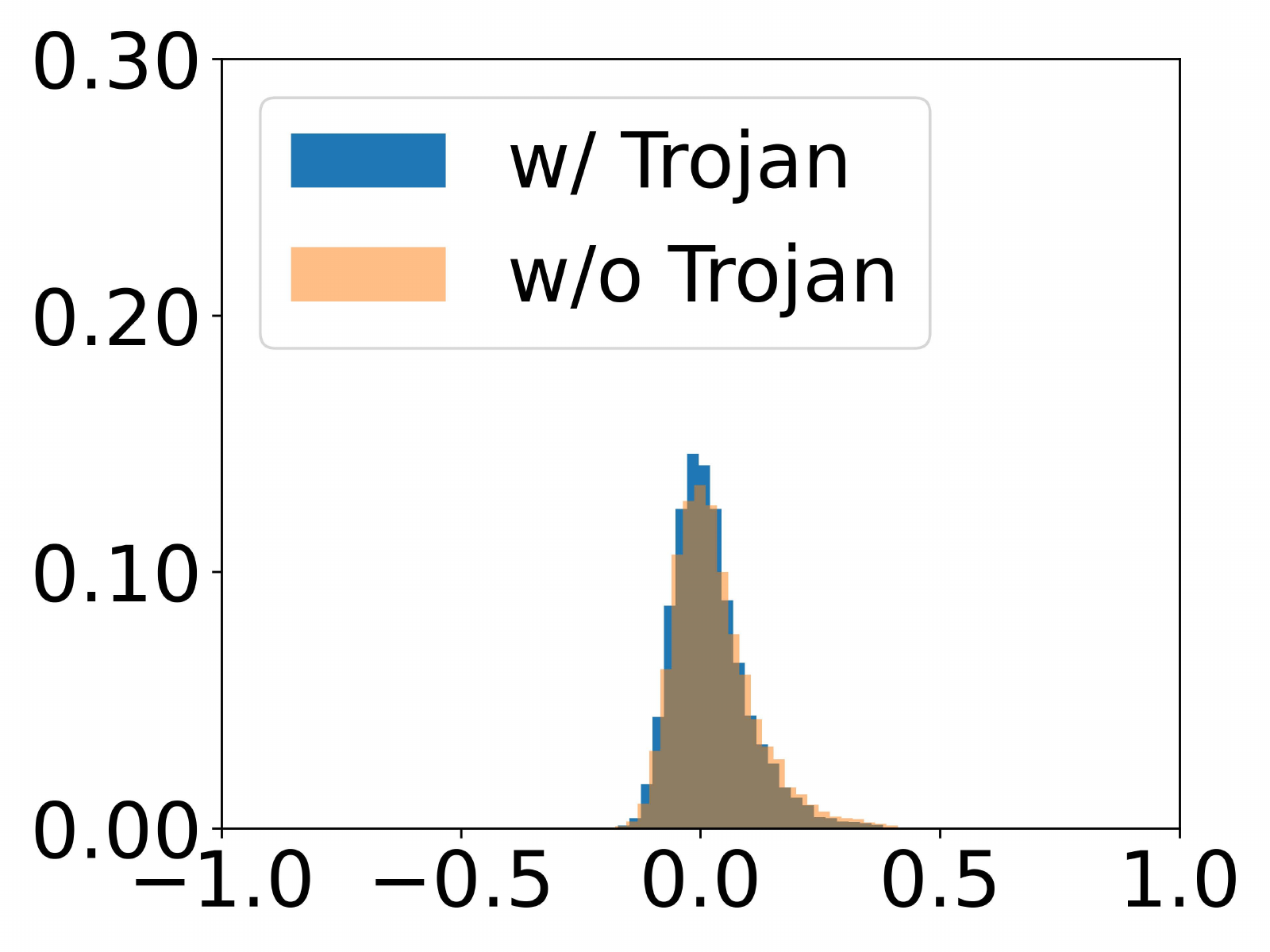}
     	 \caption{Benign}\label{fig:benign_channel}
     \end{subfigure}
     \caption{Comparison of Activation Values.}\label{fig:comparision_compromised_and_benign}
\end{wrapfigure}

\autoref{fig:comparision_compromised_and_benign} shows the empirical comparison
of a benign model and a Trojan model.
In our experiment, we train a benign model \(\mathcal{M}\) and a Trojan model \(\mathcal{M}'\) using ResNet18 on CIFAR-10.
While training \(\mathcal{M}'\), we adopt the TrojanNet~\cite{tang2020embarrassingly} training method which guarantees that only certain neurons will contain the Trojan, which we call Trojan neurons. 
The x-axis in \autoref{fig:comparision_compromised_and_benign} shows the value of \(\mathbf{W}^\mathbf{T}x+\mathbf{b}^\mathbf{T}\), and y-axis shows the percentage of Trojan neurons whose activation value is the corresponding value on x-axis.
We use blue color to denote experiments when inputs have the trigger and orange to denote when inputs do not contain the trigger.
As we can see, the model that contains the Trojan will activate the \(x>0\) region when the input contains the trigger.
By contrary, all the other three cases do not have such a phenomenon.
We also conduct similar experiments on other different models and different types of layers including non-piece-wise linear layers, and all empirical results confirm our observation here. Details are in Appendix (\autoref{sec:appendix_ee_activation}). %

\section{Training Algorithm}\label{sec:defense}

\begin{algorithm}[tb]
 	\caption{Training}\label{alg:training}
    {\bf Input:} %
    \hspace*{0.05in} Training Data: \(D\), Maximal epoch: \(E\)\\
    {\bf Output:} %
    \hspace*{0.05in} Model: \(M\)
	\begin{algorithmic}[1]
	     \Function {Training}{$D$}
      \While{\(e \leq E\) and (not \({\rm terminate}(M)\))}
      \LineComment{Train and gather activation values}
      \State \(M = {\rm train}(D, e)\)
        \State \(A = M.{\rm predict}(D)\)
        \LineComment{Identify compromised neurons}
        \State \(C = \emptyset\)
        \For{{\rm neuron} \(n\) {\rm in} \(M\)}
          \If{ \( \mathbb{P}(A_n \geq 0) \geq \theta \) }
            \State \(C = C \cup \{n\}\)
          \EndIf
        \EndFor
        \LineComment{Identify biased or poisoning samples}
        \For{neuron \(n \in C\)}
        \State \(B_{n},O_{n} = {\rm separate}(A_n)\)
        \State \(\mu, \sigma = {\rm norm(} B_n {\rm )}\)
        \For{activation value of input \(i: i_{n} \in O_n\)}
        \If {\(\left|\frac{i_n-\mu}{\sigma}\right|\geq \lambda \)}
        \State \( D = D - {i}\)
        \EndIf
        \EndFor
        \EndFor
        \LineComment{Resetting compromised neurons}
        \For {neuron \(c \in C\)}
        \State \(c = M.{\rm init}(c)\)
        \EndFor
      \EndWhile{}
         \EndFunction
	\end{algorithmic}
\end{algorithm}

To solve the Trojan problem caused by high linearity neurons, 
we propose \autoref{alg:training} to enforce non-linearity in individual layers by resetting potentially linear neurons and removing data samples that cause such linearity.
This is a revised training process, which is an iterative process that trains the model until the maximal epochs (line 2).
It starts by training the model using the standard backward propagation method (line 4).
We also gather all activation values of individual neurons \(A\) for all training samples in the training dataset \(D\).
Then, the process contains three steps: identifying compromised neurons (lines 6 to 10), identifying biased or poisoning samples (lines 11 to 17), and lastly, resetting the neurons for retraining (lines 18 to 20).

The first step is to identify comprised neurons, namely the neurons that carry Trojans.
Based on our discussion on \autoref{sec:formalana}, we do this by checking the activation values of neuron \(n\), denoted as \(A_n\), to see if its function is highly linear using the condition \(\mathbb{P}(A_n\geq 0)\geq \theta \).
If so, we make the neuron \(n\) as potentially compromised and add it to the candidate set \(C\).
The second step is to identify highly biased samples or poisoning samples.
The overall design is a statistical testing process:
we first find a reference distribution of a particular neuron 
and then mark all inputs whose activation values do not follow such distributions as potentially biased or poisoning samples. 
The idea of finding a benign distribution is that inputs whose activation values in a layer are non-linear will be considered as benign and the distribution that describes their activation values is used as the reference distribution.
In \autoref{alg:training}, line 13 uses a function to test the linearity of activation values and separate all activation values into a benign set \(B\) and others \(O\).
Then, we normalize the distribution of our reference distribution \(B\) to a normal distribution and obtain its mean value \(\mu\) and standard deviation value \(\sigma\) (line 14).
For a single input \(i\), we perform a statistical test to see the probability of \(i\) being a potentially biased or poisoning sample (lines 15 to 16).
We exclude it from the training dataset (line 17).
The last step of our training algorithm is to remove the compromised neuron effects from the model by resetting them.
We reuse the initialization method in our training setting to set a new value for all identified compromised neurons (line 19 and 20).
Then, we continue the training until terminating conditions are met, or the training budget is used up (line 2).
This algorithm uses ReLU as an example, and because our theory still holds for other activation functions (see \autoref{sec:appendix_ee_activation}), the algorithm can generalize to other models by modifying corresponding parameters (line 9). %

To identify compromised neurons and biased/poisoning images, we need to determine the threshold value \(\theta \) (line 9) and separate the activation values into non-linear and linear ones.
In our implementation, we perform the Fisher's linear discriminant analysis (its binary version, Otsu's method~\cite{otsu1979threshold}) and leverage the Jenks natural breaks optimization algorithm to find the separations of non-linear and linear activation values.
Sets \(B\) and \(O\) are outputs of such algorithms, and we use the standard value that evaluates the quality of such separation to compute the value of our threshold \(\theta \), i.e., \(0.95\) in our case.
Notice such values can affect the accuracy of identifying compromised neurons and inputs.
We also choose alternative ways to separate the sets and present the results in the Appendix to evaluate our approach.
Similar to model pruning and fine-tuning, when we reset different numbers of neurons (lines 19 and 20), the accuracy on benign and Trojan samples can be different.
We present more details of the algorithm and evaluate the sensitivity of \sys to the number of neurons reset during training and identify malicious samples in Appendix (\autoref{sec:RQ3}).

\section{Experiments}\label{sec:evaluation}
\sys is implemented in Python 3.8 with PyTorch 1.7.0 and CUDA 11.0. 
If not specified, all experiments are done on a Ubuntu 18.04 machine equipped with six GeForce RTX 6000 GPUs, 64 2.30GHz CPUs, and 376 GB memory.
We first introduce the experiment setup (\autoref{sec:setup}).
Then, we evaluate the overall effectiveness of \sys (\autoref{sec:RQ1}) and investigate its robustness against different attack settings (\autoref{sec:RQ2}).
We also evaluate the generalization of \sys on real-world applications (\autoref{sec:RQ4}).
Finally, we measure the precision and the recall in the poisoned samples identification stage (\autoref{sec:precision_recall}).
Other results and discussions are in Appendix.
\subsection{Experiment Setup.}

\label{sec:setup}

\begin{wrapfigure}{R}{0.5\textwidth}

     \centering
     \footnotesize
     \scalebox{0.7}{
     \begin{subfigure}[t]{0.17\columnwidth}
         \centering     
         \footnotesize
         \includegraphics[width=\columnwidth]{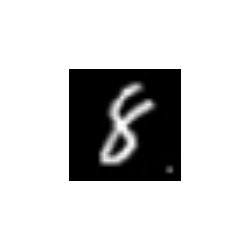}
         \caption{Single Pixel}
         \label{fig:trigger_mnist}
     \end{subfigure}
     \hfill
     \begin{subfigure}[t]{0.17\columnwidth}
         \centering
     	   \footnotesize
     	   \includegraphics[width=\columnwidth]{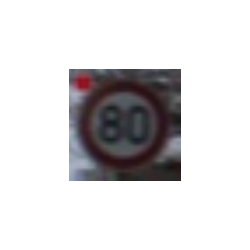}
     	   \caption{Static}
     	   \label{fig:trigger_gtsrb}
     \end{subfigure}
     \hfill
     \begin{subfigure}[t]{0.17\columnwidth}
         \centering     
         \footnotesize
         \includegraphics[width=\columnwidth]{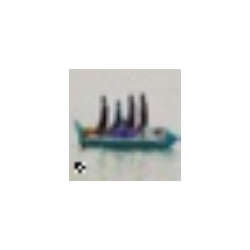}
     	   \caption{Dynamic}
         \label{fig:trigger_cifar10}
     \end{subfigure}
     \hfill
     \label{fig:natural_cifar_compare}
     \begin{subfigure}[t]{0.17\columnwidth}
         \centering
     	   \footnotesize
     	   \includegraphics[width=\columnwidth]{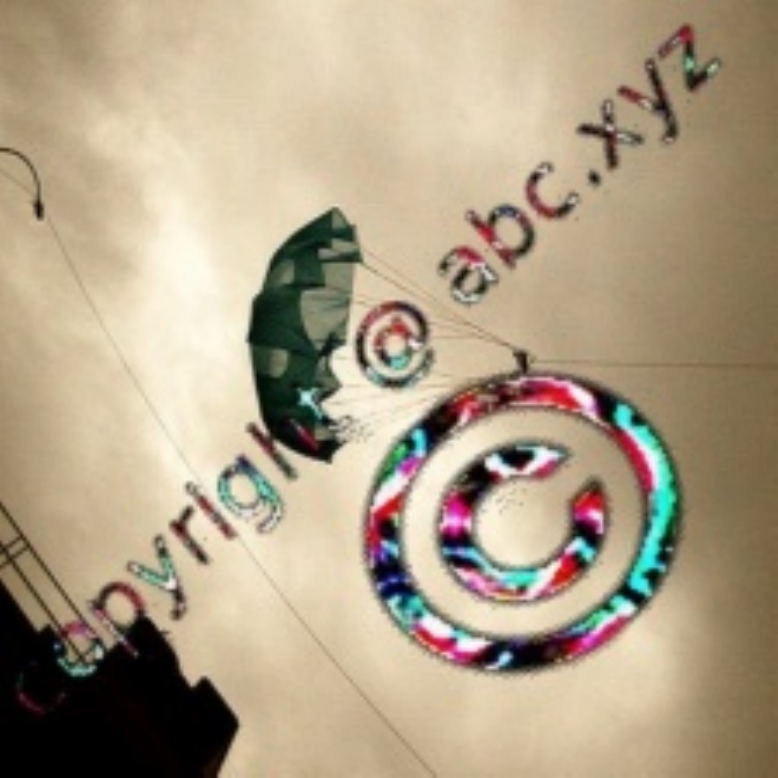}
     	   \caption{Watermark}
     	   \label{fig:trigger_imagenette}
     \end{subfigure}}
     \caption{Examples of Using Different Trigger Patterns.}
     \label{fig:triggers}
\end{wrapfigure}

\noindent
\textbf{Datasets and Models.}
We evaluate \sys on five publicly available datasets: MNIST~\cite{lecun1998gradient}, GTSRB~\cite{stallkamp2012man}, CIFAR-10~\cite{krizhevsky2009learning}, ImageNet-10~\cite{krizhevsky2012imagenet} and TrojAI~\cite{karra2020trojai}.
The overview of our datasets and more details can be found in \autoref{sec:details_datasets}.
\autoref{fig:triggers} illustrates different trigger patterns used in the experiments
(i.e., a single pixel located in the right bottom corner of~\autoref{fig:trigger_mnist}, a fixed red patch in~\autoref{fig:trigger_gtsrb}, a black-and-white pattern whose location is random in~\autoref{fig:trigger_cifar10} and a colorful watermark in ~\autoref{fig:trigger_imagenette}). 
The default trigger pattern for MNIST, CIFAR-10, GTSRB, and ImageNet-10 are Single Pixel (\autoref{fig:trigger_mnist}), Static Patch (\autoref{fig:trigger_cifar10}), Dynamic Patch (\autoref{fig:trigger_gtsrb}) and Watermark (\autoref{fig:trigger_imagenette}), respectively.
Besides the default triggers for each dataset, we also measure the impacts of using other triggers.
The results are included in~\autoref{sec:RQ2}.
We evaluate \sys and other defense methods on AlexNet~\cite{krizhevsky2012imagenet}, NiN (Network in Network)~\cite{lin2013network}, VGG11, VGG16~\cite{simonyan2014very} and ResNet18~\cite{he2016deep}.
These models are representative and are commonly used in existing Trojan related studies~\cite{liu2019abs,wang2019neural}.

\noindent
\textbf{Evaluation Metrics.}
We use benign accuracy (BA) and attack success rate (ASR)~\cite{veldanda2020nnoculation} as evaluation metrics, which is a common practice~\cite{doan2020februus,gao2019strip,wang2019neural}.
BA is defined as the number of correctly classified benign samples 
over the number of all benign samples. It implies model's capability on its original task.
ASR evaluates the success rate of backdoor attacks.
It is calculated as the number of backdoor samples that can successfully attack the model over the number of all generated backdoor samples.

\noindent
\textbf{Attack Settings.}
As we introduced in~\autoref{sec:bg}, Trojans are classified into two categories: \textit{Injected Trojans} and \textit{Natural Trojans}.
We implement both of them to evaluate \sys and other defense methods.
For Injected Trojans, we implement both single target and label specific BadNets~\cite{gu2017badnets}, label-consistent Trojan attack~\cite{turner2019label} and hidden trigger Trojan attack~\cite{saha2020hidden}. 
Following the original paper, we use images from a pair of classes (class theater curtain and class plunger) in ImageNet for hidden trigger Trojan attack.
For natural Trojans, we follow the previous work~\cite{liu2019abs} to reproduce the attack.
Due to the fact that we do not know if the data is poisoned or not in advance, \sys keeps detecting Trojan attacks and training the model even if there is no attack activity.
We evaluate the additional costs of \sys on benign models in such non-attack settings.
More details for our attack settings are included in \autoref{sec:details_attack}.
Besides above attacks, We also evaluate the generalization of \sys on more attacks in \autoref{sec:resistance_more_attacks}.

\noindent
\textbf{Comparison.}
We compare \sys with 4 state-of-the-art defense methods: DP-SGD~\cite{hong2020effectiveness}, Neural Attention Distillation
(NAD)~\cite{li2021neural}, Activation Clustering
(AC)~\cite{chen2018detecting}
and Anti-backdoor Learning (ABL)~\cite{li2021anti}.
We use their official code and default hyperparameters specified in the original papers.

\subsection{Effectiveness of \sys}
\label{sec:RQ1}

\begin{table*}[]
    \centering
    \scriptsize
    \setlength\tabcolsep{1.3pt}
    \caption{Comparisons on Injected Trojans.
    }
    \label{tab:standard backdoor attack}
    \scalebox{0.9}{
        \begin{tabular}{@{}ccccrrrrrrrrrrrrrrrrr@{}}
\toprule
\multirow{2}{*}{Attack Type}                                                         & \multirow{2}{*}{Dataset}     & \multirow{2}{*}{Network} &  & \multicolumn{2}{c}{Undefended}                   & \multicolumn{1}{c}{} & \multicolumn{2}{c}{DP-SGD}                       & \multicolumn{1}{c}{} & \multicolumn{2}{c}{NAD}                          & \multicolumn{1}{c}{} & \multicolumn{2}{c}{AC}                           & \multicolumn{1}{c}{} & \multicolumn{2}{c}{ABL}                          & \multicolumn{1}{c}{} & \multicolumn{2}{c}{NONE}                         \\ \cmidrule(lr){5-6} \cmidrule(lr){8-9} \cmidrule(lr){11-12} \cmidrule(lr){14-15} \cmidrule(lr){17-18} \cmidrule(l){20-21} 
                                                                                     &                              &                          &  & \multicolumn{1}{c}{BA} & \multicolumn{1}{c}{ASR} & \multicolumn{1}{c}{} & \multicolumn{1}{c}{BA} & \multicolumn{1}{c}{ASR} & \multicolumn{1}{c}{} & \multicolumn{1}{c}{BA} & \multicolumn{1}{c}{ASR} & \multicolumn{1}{c}{} & \multicolumn{1}{c}{BA} & \multicolumn{1}{c}{ASR} & \multicolumn{1}{c}{} & \multicolumn{1}{c}{BA} & \multicolumn{1}{c}{ASR} & \multicolumn{1}{c}{} & \multicolumn{1}{c}{BA} & \multicolumn{1}{c}{ASR} \\ \midrule
\multirow{12}{*}{\begin{tabular}[c]{@{}c@{}}BadNets\\ (single target)\end{tabular}}  & \multirow{3}{*}{MNIST}       & NiN                      &  & 99.65\%                & 99.96\%                 &                      & 92.94\%                & 0.48\%                  &                      & 99.26\%                & 0.06\%                  &                      & 98.47\%                & 0.51\%                  &                      & 98.25\%                & 99.70\%                 &                      & \textbf{99.58\%}                & \textbf{0.06\%}                  \\
                                                                                     &                              & VGG11                    &  & 99.35\%                & 100.00\%                &                      & 97.20\%                & 1.61\%                  &                      & 98.85\%                & 2.12\%                  &                      & 97.44\%                & 10.93\%                 &                      & 98.48\%                & 0.64\%                  &                      & \textbf{99.11\%}                & \textbf{0.14\%}                  \\
                                                                                     &                              & ResNet18                 &  & 99.61\%                & 99.98\%                 &                      & 97.31\%                & 0.13\%                  &                      & 98.83\%                & 0.12\%                  &                      & 98.04\%                & 0.25\%                  &                      & 99.34\%                & \textbf{0.04\%}                  &                      & \textbf{99.57\%}                & 0.37\%                  \\ \cmidrule(l){2-21} 
                                                                                     & \multirow{3}{*}{CIFAR-10}    & NiN                      &  & 90.52\%                & 100.00\%                &                      & 36.41\%                & 93.01\%                 &                      & 80.99\%                & 52.97\%                 &                      & 83.77\%                & 100.00\%                &                      & \textbf{90.13\%}                & 100.00\%                &                      & 90.11\%                & \textbf{2.32\%}                  \\
                                                                                     &                              & VGG16                    &  & 90.46\%                & 100.00\%                &                      & 55.61\%                & 99.32\%                 &                      & 88.70\%                & 98.71\%                 &                      & 88.14\%                & 99.84\%                 &                      & \textbf{89.96\%}                & 100.00\%                &                      & 89.70\%                & \textbf{4.91\%}                  \\
                                                                                     &                              & ResNet18                 &  & 94.10\%                & 100.00\%                &                      & 52.29\%                & 99.99\%                 &                      & 88.74\%                & 1.28\%                  &                      & 89.57\%                & 57.43\%                 &                      & 92.40\%                & 1.66\%                  &                      & \textbf{93.62\%}                & \textbf{1.07\%}                  \\ \cmidrule(l){2-21} 
                                                                                     & \multirow{3}{*}{GTSRB}       & NiN                      &  & 95.95\%                & 99.72\%                 &                      & 31.54\%                & 74.78\%                 &                      & \textbf{96.53\%}                & 26.15\%                 &                      & 95.36\%                & 99.52\%                 &                      & 96.14\%                & 99.54\%                 &                      & 95.51\%                & \textbf{0.87\%}                  \\
                                                                                     &                              & VGG16                    &  & 95.43\%                & 99.93\%                 &                      & 54.60\%                & 86.83\%                 &                      & \textbf{95.94\%}                & 86.10\%                 &                      & 91.28\%                & 5.28\%                  &                      & 94.06\%                & 95.79\%                 &                      & 94.66\%                & \textbf{0.96\%}                  \\
                                                                                     &                              & ResNet18                 &  & 96.67\%                & 99.84\%                 &                      & 55.73\%                & 90.88\%                 &                      & 95.95\%                & 13.18\%                 &                      & 94.64\%                & 98.57\%                 &                      & 96.24\%                & 0.93\%                  &                      & \textbf{96.39\%}                & \textbf{0.76\%}                  \\ \cmidrule(l){2-21} 
                                                                                     & \multirow{3}{*}{ImageNet-10} & NiN                      &  & 82.17\%                & 99.64\%                 &                      & 21.22\%                & 65.21\%                 &                      & 76.99\%                & 92.61\%                 &                      & 74.32\%                & 92.41\%                 &                      & \textbf{81.35\%}                & 99.92\%                 &                      & 80.31\%                & \textbf{0.14\%}                  \\
                                                                                     &                              & VGG16                    &  & 88.97\%                & 100.00\%                &                      & 18.06\%                & 44.62\%                 &                      & 84.03\%                & 10.93\%                 &                      & 81.58\%                & 100.00\%                &                      & 79.16\%                & 100.00\%                &                      & \textbf{87.11\%}                & \textbf{0.14\%}                  \\
                                                                                     &                              & ResNet18                 &  & 89.83\%                & 99.07\%                 &                      & 40.31\%                & 29.58\%                 &                      & 86.29\%                & 25.89\%                 &                      & 82.39\%                & 98.44\%                 &                      & 83.72\%                & 6.83\%                  &                      & \textbf{86.34\%}                & \textbf{0.08\%}                  \\ \midrule
\multirow{12}{*}{\begin{tabular}[c]{@{}c@{}}BadNets\\ (label specific)\end{tabular}} & \multirow{3}{*}{MNIST}       & NiN                      &  & 99.64\%                & 98.86\%                 &                      & 93.02\%                & 1.10\%                  &                      & 99.24\%                & 0.09\%                  &                      & 99.60\%                & 0.03\%                  &                      & 98.40\%                & 89.32\%                 &                      & \textbf{99.42\%}                & \textbf{0.03\%}                  \\
                                                                                     &                              & VGG11                    &  & 99.05\%                & 98.99\%                 &                      & 97.17\%                & 0.41\%                  &                      & 99.06\%                & 32.06\%                 &                      & 97.59\%                & 0.10\%                  &                      & \textbf{99.21\%}                & 98.51\%                 &                      & 98.63\%                & \textbf{0.11\%}                  \\
                                                                                     &                              & ResNet18                 &  & 99.57\%                & 99.49\%                 &                      & 96.81\%                & 0.38\%                  &                      & 99.19\%                & 0.19\%                  &                      & \textbf{99.35\%}                & \textbf{0.03\%}                  &                      & 99.01\%                & 98.81\%                 &                      & 99.09\%                & 0.20\%                  \\ \cmidrule(l){2-21} 
                                                                                     & \multirow{3}{*}{CIFAR-10}    & NiN                      &  & 90.50\%                & 78.56\%                 &                      & 38.42\%                & 6.43\%                  &                      & 82.70\%                & 13.65\%                 &                      & 84.74\%                & 56.57\%                 &                      & \textbf{89.66\%}                & 77.69\%                 &                      & 89.51\%                & \textbf{1.27\%}                  \\
                                                                                     &                              & VGG16                    &  & 90.73\%                & 96.86\%                 &                      & 55.12\%                & 5.23\%                  &                      & 88.65\%                & 39.99\%                 &                      & 88.18\%                & 82.87\%                 &                      & \textbf{90.13\%}                & 86.91\%                 &                      & 89.64\%                & \textbf{1.22\%}                  \\
                                                                                     &                              & ResNet18                 &  & 94.37\%                & 92.00\%                 &                      & 52.19\%                & 12.17\%                 &                      & 88.21\%                & 1.50\%                  &                      & 92.68\%                & 5.29\%                  &                      & 83.83\%                & 83.48\%                 &                      & \textbf{93.05\%}                & \textbf{1.04\%}                  \\ \cmidrule(l){2-21} 
                                                                                     & \multirow{3}{*}{GTSRB}       & NiN                      &  & 96.06\%                & 93.74\%                 &                      & 24.70\%                & 7.39\%                  &                      & \textbf{96.46\%}                & 9.90\%                  &                      & 94.02\%                & 6.17\%                  &                      & 96.14\%                & 99.54\%                 &                      & 95.99\%                & \textbf{0.96\%}                  \\
                                                                                     &                              & VGG16                    &  & 95.71\%                & 94.57\%                 &                      & 53.90\%                & 6.71\%                  &                      & \textbf{96.33\%}                & 81.43\%                 &                      & 94.96\%                & 69.22\%                 &                      & 94.32\%                & 80.21\%                 &                      & 95.49\%                & \textbf{1.65\%}                  \\
                                                                                     &                              & ResNet18                 &  & 96.93\%                & 97.40\%                 &                      & 60.59\%                & 6.61\%                  &                      & 95.49\%                & 11.13\%                 &                      & 95.74\%                & 1.16\%                  &                      & 90.24\%                & 93.93\%                 &                      & \textbf{96.63\%}                & \textbf{0.91\%}                  \\ \cmidrule(l){2-21} 
                                                                                     & \multirow{3}{*}{ImageNet-10} & NiN                      &  & 82.65\%                & 66.37\%                 &                      & 23.44\%                & 10.06\%                 &                      & 76.31\%                & 23.31\%                 &                      & 78.85\%                & 14.39\%                 &                      & \textbf{80.20\%}                & 56.82\%                 &                      & 79.39\%                & \textbf{6.98\%}                  \\
                                                                                     &                              & VGG16                    &  & 89.04\%                & 76.20\%                 &                      & 17.94\%                & 10.80\%                 &                      & 80.21\%                & 22.17\%                 &                      & 81.22\%                & 59.62\%                 &                      & 78.34\%                & 53.68\%                 &                      & \textbf{84.51\%}                & \textbf{8.31\%}                  \\
                                                                                     &                              & ResNet18                 &  & 88.87\%                & 59.75\%                 &                      & 44.59\%                & 7.95\%                  &                      & 85.76\%                & 15.82\%                 &                      & 79.44\%                & 23.41\%                 &                      & 81.30\%                & 50.52\%                 &                      & \textbf{84.74\%}                & \textbf{6.55\%}                  \\ \midrule
\multirow{3}{*}{Label consistent}                                                         & \multirow{3}{*}{CIFAR-10}    & NiN                      &  & 91.32\%                & 98.98\%                 &                      & 38.83\%                & 5.21\%                  &                      & 82.35\%                & 65.63\%                 &                      & 83.92\%                & 95.96\%                 &                      & 89.95\%                & 95.28\%                 &                      & \textbf{90.11\%}                & \textbf{2.19\%}                  \\
                                                                                     &                              & VGG16                    &  & 90.97\%                & 98.41\%                 &                      & 53.76\%                & 9.24\%                  &                      & 88.88\%                & 92.71\%                 &                      & 88.46\%                & 7.57\%                  &                      & 90.38\%                & 94.94\%                 &                      & \textbf{90.07\%}                & \textbf{4.26\%}                  \\
                                                                                     &                              & ResNet18                 &  & 94.73\%                & 83.42\%                 &                      & 54.00\%                & 11.00\%                 &                      & 90.74\%                & 60.74\%                 &                      & 88.00\%                & 65.86\%                 &                      & 87.34\%                & 2.17\%                  &                      & \textbf{94.01\%}                & \textbf{2.14\%}                  \\ \midrule
Hidden Trigger                                                                       & ImageNet-pair                & AlexNet                  &  & 93.00\%                & 82.00\%                 &                      & 80.00\%                & 62.00\%                 &                      & 91.00\%                & 74.00\%                 &                      & 90.00\%                & 22.00\%                 &                      & 90.00\%                & 54.00\%                 &                      & \textbf{91.00\%}                & \textbf{4.00\%}                  \\ \bottomrule
\end{tabular}}
    \end{table*}
\begin{table}[]
\centering
\scriptsize
\setlength\tabcolsep{3pt}
\caption{Comparisons on Natural Trojan.}
\label{tab:natural}
\scalebox{1}{
    \begin{tabular}{@{}cccccccccccccc@{}}
\toprule
\multirow{2}{*}{Dataset} & \multirow{2}{*}{Network} &  & \multicolumn{2}{c}{Undefended} &  & \multicolumn{2}{c}{DP-SGD-1} &  & \multicolumn{2}{c}{DP-SGD-2} &  & \multicolumn{2}{c}{NONE} \\ \cmidrule(lr){4-5} \cmidrule(lr){7-8} \cmidrule(lr){10-11} \cmidrule(l){13-14} 
                         &                          &  & BA             & ASR           &  & BA            & ASR          &  & BA            & ASR          &  & BA          & ASR        \\ \midrule
CIFAR-10                 & NiN                      &  & 91.02\%        & 87.62\%       &  & 60.22\%       & 98.85\%      &  & 39.19\%       & 87.22\%      &  & \textbf{86.94\%}     & \textbf{34.21\%}    \\
                         & VGG16                    &  & 90.78\%        & 71.88\%       &  & 78.25\%       & 61.11\%      &  & 53.40\%       & 63.58\%      &  & \textbf{81.83\%}     & \textbf{37.49\%}    \\ \midrule
\multirow{2}{*}{TrojAI}  & VGG11                    &  & 99.88\%        & 72.09\%       &  & 84.51\%       & 88.75\%      &  & 6.06\%        & 88.55\%      &  & \textbf{99.04\%}     & \textbf{56.68\%}    \\
                         & Resnet18                 &  & 99.91\%        & 54.33\%       &  & 78.88\%       & 61.77\%      &  & 52.81\%       & 58.94\%      &  & \textbf{98.13\%}     & \textbf{34.98\%}    \\ \bottomrule
\end{tabular}}
\end{table}

{\bf Experiments.}
We measure the effectiveness of \sys by comparing the BA and ASR of models protected by \sys with those of undefended models and models protected by existing defense methods.
The comparison results on injected Trojans, natural Trojans and non-attack settings are shown in \autoref{tab:standard backdoor attack}, \autoref{tab:natural} and \autoref{tab:benign model}, respectively.
In each table, we show the detailed settings including attack settings, dataset names and network architectures., etc.
For the evaluation results on natural Trojans (~\autoref{tab:natural}), the ASR and BA are the average results under different trigger size settings: 2\%, 4\%, 6\%, 8\%, 10\% and 12\% of the whole image.
Notice that, to the best of our knowledge, there is no defense method designed for natural Trojans.
We observe that DP-SGD can potentially mitigate natural Trojans because it reduces the high gradients brought from natural Trojans.
Therefore, we adapt DP-SGD as the baseline method for natural Trojans.
We configure DP-SGD with two settings of parameters following the prior work~\cite{hong2020effectiveness}. 
For DP-SGD-1, we set the clip as 4.0 and the noise as 0.1. The clip and noise of DP-SGD-2 are 1.0 and 0.5. 
For the non-attack settings, we deploy \sys on several benign models and show the decrease of BA in \autoref{tab:benign model}.

{\bf Results on Injected Trojans.}
From the results on Injected Trojan attacks (\autoref{tab:standard backdoor attack}), we observe that applying \sys can better protect models from being attacked by injected Trojans than other defense methods. 
With \sys, the average ASR of models decreases from 93.34\% to 1.91\%, which is much better than other defense methods (DP-SGD,  NAD, AC and ABL can only reduce the average ASR to 30.32\%, 34.08\%, 45.45\% and 68.60\% respectively).
The reason is that, unlike existing methods based only on specific empirical observations, \sys targets the root cause of the Trojans (i.e., the linearity) and reveals the attacks more accurately.
Therefore, \sys can better defend against attacks than other methods.

We also find that \sys almost does not have negative impacts on the original task of models.
From~\autoref{tab:standard backdoor attack}, the BA of \sys is the highest among all methods and is similar to that of undefended models, 
meaning that \sys has a low defense cost.
The reason is that \sys only modifies the compromised neurons that are highly relevant to Trojan but less related to the original task of models.
Therefore, most of the benign knowledge is preserved when applying \sys and the model can still perform well on its original tasks.
Meanwhile, \sys finetunes the model on the purified data after the reset process, further strengthening the capabilities of models and reducing defense costs.

It is worth clarifying that ABL has a poor performance on label-specific attacks and other attacks that use NiN and VGG models.
The possible reason is that the design of ABL requires the model to learn quicker and better on Trojan samples than benign samples~\cite{li2021anti} (i.e., the learning of Trojan samples should have a lower training loss value in the early training stage).
When the attacker uses more complex attacks (e.g., label specific BadNets) or models with limited learning capabilities (e.g., NiN and VGG), the model cannot learn Trojan samples quickly, leading to poor performance.

{\bf Results on Natural Trojans.}
From the results in ~\autoref{tab:natural}, we find that \sys protects the model most effectively and has the lowest defense costs among all defense methods.
Overall, applying \sys achieves 1.75 times lower ASR than undefended models.
DP-SGD methods can only slightly decreases ASR or even increase ASR.
The results show that using \sys is the most effective way for natural Trojan defense.
The loss of BA using \sys is also smaller than the loss caused by applying the DP-SGD methods (3.90\% with \sys and 38.76\% with DP-SGD methods on average), further showing the efficiency of \sys.
The results confirm our analysis: reducing the linearity of models reduces the ASR of natural Trojans without posing much additional cost.

{\bf Results on Non-attack Setting.}
We also explore whether \sys affects the performance of benign models on their original tasks.
From~\autoref{tab:benign model}, we find that \sys has a low effect on benign models.
Applying \sys only decreases 2.33\% BA on average.
Because only a few neurons are detected as compromised neurons and are reset by \sys when there is no Trojan activity, \sys does not affect the learned benign knowledge.
Moreover, the subsequent training process further reduces the costs.
Therefore, we conclude that \sys does not impose high additional costs on benign models.

\begin{table*}[]
    \begin{minipage}{0.5\linewidth}
    \centering
	\scriptsize
	\setlength\tabcolsep{3pt}
	\caption{Benign Accuracy in Non-attack
		Settings.}\label{tab:benign model}
    \begin{tabular}{@{}cccc@{}}
    \toprule
    Dataset                      & Network      & Without \sys & With \sys \\ \midrule
    \multirow{3}{*}{CIFAR-10}    & NiN          & 91.02\%    & 89.40\% \\
                                 & VGG16        & 90.78\%    & 89.62\% \\
                                 & ResNet18     & 94.83\%    & 93.92\% \\ \midrule
    \multirow{3}{*}{GTSRB}       & NiN          & 95.68\%    & 95.36\% \\
                                 & VGG16        & 94.67\%    & 94.08\% \\
                                 & ResNet18     & 96.89\%    & 96.87\% \\ \midrule
    \multirow{3}{*}{ImageNet-10} & NiN & 83.34\%    & 79.18\% \\
                                 & VGG16        & 88.84\%    & 83.41\% \\
                                 & ResNet18     & 89.81\%    & 85.25\% \\ \bottomrule
    \end{tabular}
    \end{minipage}
    \begin{minipage}{0.5\linewidth}
    \caption{Results on Different Trigger Patterns.}
    \label{tab:different trigger}
    \centering
	\scriptsize
	\setlength\tabcolsep{3pt}
    \scalebox{0.9}{
    \begin{tabular}{@{}cccrrlr@{}}
    \toprule
    \multirow{2}{*}{Trigger Pattern} & \multirow{2}{*}{Network} & \multicolumn{2}{c}{Undefended}    & \multicolumn{1}{c}{} & \multicolumn{2}{c}{\sys}          \\ \cmidrule(lr){3-4} \cmidrule(l){6-7} 
                                     &                          & BA      & \multicolumn{1}{c}{ASR} & \multicolumn{1}{c}{} & \multicolumn{1}{c}{BA} & \multicolumn{1}{c}{ASR} \\ \midrule
    \multirow{3}{*}{Dynamic Patch}   & NiN                      & 90.52\% & 100.00\%                &                      & 90.11\%                & 2.32\%                  \\
                                     & VGG16                    & 90.46\% & 100.00\%                &                      & 89.70\%                & 4.91\%                  \\
                                     & ResNet18                 & 94.10\% & 100.00\%                &                      & 93.62\%                & 1.07\%                  \\ \midrule
    \multirow{3}{*}{Static Patch}    & NiN                      & 90.92\% & 100.00\%                &                      & 89.93\%                & 2.61\%                  \\
                                     & VGG16                    & 90.12\% & 100.00\%                &                      & 89.48\%                & 4.03\%                  \\
                                     & ResNet18                 & 94.24\% & 99.99\%                 &                      & 93.93\%                & 1.37\%                  \\ \midrule
    \multirow{3}{*}{Watermark}       & NiN                      & 90.88\% & 99.99\%                 &                      & 87.74\%                & 3.27\%                  \\
                                     & VGG16                    & 90.64\% & 99.99\%                 &                      & 89.14\%                & 5.36\%                  \\
                                     & ResNet18                 & 94.28\% & 100.00\%                &                      & 92.27\%                & 5.99\%                  \\ \bottomrule
    \end{tabular}}
    \end{minipage}
\end{table*}

\begin{table*}[]
    \begin{minipage}{0.5\linewidth}
    \centering
    \scriptsize
    \caption{Results on Different Trigger Sizes.}\label{tab:different trigger size}
    \setlength\tabcolsep{3pt}
    \scalebox{0.9}{
    \begin{tabular}{@{}crrcrr@{}}
    \toprule
    \multirow{2}{*}{Trigger Size} & \multicolumn{2}{c}{Undefended}                         &  & \multicolumn{2}{c}{\sys}                \\ \cmidrule(lr){2-3} \cmidrule(l){5-6} 
                                  & \multicolumn{1}{c}{BA} & \multicolumn{1}{c}{ASR} &  & \multicolumn{1}{c}{BA} & \multicolumn{1}{c}{ASR} \\ \midrule
    3*3                           & 94.10\%                      & 100.00\%                &  & 93.62\%                      & 1.07\%                  \\
    5*5                           & 94.27\%                      & 100.00\%                &  & 93.51\%                      & 1.34\%                  \\
    7*7                           & 94.10\%                      & 99.98\%                 &  & 93.70\%                      & 1.53\%                  \\
    9*9                           & 94.34\%                      & 100.00\%                &  & 93.19\%                      & 5.18\%                  \\
    15*15                           & 94.44\%                      & 100.00\%                &  & 92.18\%                      & 32.27\%                  \\
    20*20                           & 94.58\%                      & 100.00\%                &  & 92.55\%                      & 99.82\%                  \\
    \bottomrule
    \end{tabular}}
    \end{minipage}
    \begin{minipage}{0.5\linewidth}
    \centering
    \scriptsize
    \caption{Results on Different Poisoning Rates.}
    \label{tab:different poisoning fraction}
    \setlength\tabcolsep{3pt}
    \scalebox{1}{
    \begin{tabular}{@{}crrcrr@{}}
    \toprule
    \multirow{2}{*}{Poisoning rate} & \multicolumn{2}{c}{Undefended}                   &  & \multicolumn{2}{c}{\sys}          \\ \cmidrule(lr){2-3} \cmidrule(l){5-6} 
                                    & \multicolumn{1}{c}{BA} & \multicolumn{1}{c}{ASR} &  & \multicolumn{1}{c}{BA} & \multicolumn{1}{c}{ASR} \\ \midrule
    0.50\%                          & 94.46\%                & 100.00\%                &  & 93.50\%                & 2.52\%                  \\
    5.00\%                             & 94.10\%                & 100.00\%                &  & 93.62\%                & 1.07\%                  \\
    10.00\%                            & 93.82\%                & 100.00\%                &  & 93.13\%                & 1.04\%                  \\
    20.00\%                            & 92.70\%                & 100.00\%                &  & 92.14\%                & 1.39\%                  \\ \bottomrule
    \end{tabular}}
    \end{minipage}
\end{table*}

\subsection{Robustness of \sys}
\label{sec:RQ2}
We evaluate the robustness of \sys against various attack settings (e.g., different trigger sizes, trigger patterns and poisoning rates).
If not specified, the model used in the evaluation is ResNet18.
The dataset is CIFAR-10 and the evaluated attack is the single target BadNets attack.

{\bf Trigger Sizes.}
To study the effects of trigger size, we use triggers of different sizes (from 3*3 to 20*20) to attack models and collect the ASR and BA of applying \sys on these compromised models.
The results are shown in~\autoref{tab:different trigger size}.
Overall, the BA of undefended models and models protected by \sys is insensitive to the change of trigger size.
The difference between the highest and lowest BA on the unprotected and protected models is 0.48\% and 1.52\%, respectively, which is very small.
We believe that the BA does not change significantly because triggers usually do not affect the learning of benign features used for original tasks, as discussed in previous work~\cite{gu2017badnets}.

\begin{figure}[]
     \centering
     \footnotesize
     \scalebox{0.95}{
     \begin{subfigure}[t]{0.18\columnwidth}
         \centering     
         \footnotesize
         \includegraphics[width=\columnwidth]{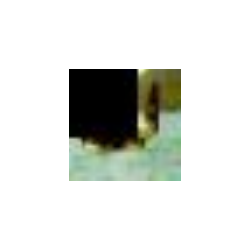}
         \label{fig:large_size0}
     \end{subfigure}
     \hfill
     \begin{subfigure}[t]{0.18\columnwidth}
         \centering
     	   \footnotesize
     	   \includegraphics[width=\columnwidth]{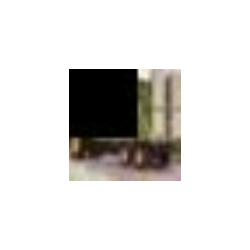}
     	   \label{fig:large_size1}
     \end{subfigure}
     \hfill
     \begin{subfigure}[t]{0.18\columnwidth}
         \centering     
         \footnotesize
         \includegraphics[width=\columnwidth]{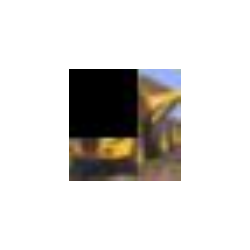}
         \label{fig:large_size2}
     \end{subfigure}
     \hfill
     \label{fig:natural_cifar_compare}
     \hspace{-.06in}
     \begin{subfigure}[t]{0.18\columnwidth}
         \centering
     	   \footnotesize
     	   \includegraphics[width=\columnwidth]{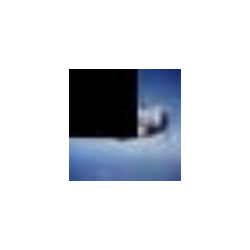}
     	   \label{fig:large_size3}
     \end{subfigure}
     }
     \caption{Trojan Inputs with Large Triggers.}
     \label{fig:large_size}
     \vspace{-0.3cm}
\end{figure}

On the other hand, the trigger size affects the ASR of protected models.
When the trigger size becomes larger, the ASR of the protected model increases dramatically from 1.07\% to 99.82\% and \sys fails.
The results are understandable because a large trigger size modifies more pixels in the original image, making the triggers obvious and easy to learn.
When the trigger is large, it almost covers the whole image and becomes the majority of the image.
In such a scenario, models easily capture trigger features and are compromised.
A detailed example is shown in \autoref{fig:large_size}.
Currently, the sensitivity to trigger sizes is a common limitation for Trojan defense methods~\cite{wang2019neural,liu2019abs}.
Considering that a large trigger size is almost impractical because it makes the trigger too obvious to be detected directly by administrators, we consider \sys robust to most trigger sizes.

{\bf Trigger Patterns.}
To measure the robustness of \sys against different attack trigger patterns, we use \sys to protect models from being attacked by Dynamic Patch trigger, Static Patch trigger and Watermark trigger.
The results are shown in \autoref{tab:different trigger}.
We observe that \sys always achieves low ASR and high BA under different trigger settings.
The results demonstrate that \sys is effective against different trigger pattern settings.
Moreover, we notice that the ASR of using the watermark trigger is particularly larger compared with using other triggers. The reason is that the watermark triggers are large and more complex, as shown in \autoref{fig:triggers}.

{\bf Poisoning Rates.}
To measure the impacts of different poisoning rates, we collect the ASR and BA of models being compromised at different poisoning rates from 0.50\% to 20.00\%.
The results are summarized in ~\autoref{tab:different poisoning fraction}.
Based on the results, we find that increasing the poisoning rate slightly decreases both the BA and ASR.
Specifically, the BA of the model decreases by 1.36\% and the ASR of the model decreases by 1.13\%.
This is because increasing the poisoning rate reduces the number of benign samples used for training, and the BA of the model naturally decreases.
Meanwhile, a large poisoning rate makes Trojans easy to be detected, which leads to a lower ASR.
Since the changes in ASR and BA are quite small, \sys is considered robust to most poisoning rates.

\begin{table}[]
\centering
\scriptsize
\caption{Results on Federated Learning.}
\label{tab:federal learning}
\vspace{0.1cm}
\scalebox{1}{
\begin{tabular}{@{}ccrlcr@{}}
\toprule
\multirow{2}{*}{Dataset} & \multicolumn{2}{c}{Undefended}                        &  & \multicolumn{2}{c}{\sys}                              \\ \cmidrule(lr){2-3} \cmidrule(l){5-6} 
                         & BA                          & \multicolumn{1}{c}{ASR} &  & BA                          & \multicolumn{1}{c}{ASR} \\ \midrule
MNIST                    & \multicolumn{1}{r}{99.22\%} & 99.15\%                 &  & \multicolumn{1}{r}{98.60\%} & 0.13\%                  \\
CIFAR-10                  & \multicolumn{1}{r}{80.31\%} & 43.23\%                 &  & \multicolumn{1}{r}{78.67\%} & 4.44\%                  \\ \bottomrule
\end{tabular}}
\end{table}

\subsection{Generalization on Complex Applications}\label{sec:RQ4}
Attackers may conduct attacks in a more complex scenario.
To measure the generalization of \sys on complex applications, we evaluate \sys on two federated learning applications trained on different datasets (i.e., MNIST and CIFAR-10) and a transfer learning application.
Each federated learning application has 10 participants, of which 4 of them are malicious participants who conduct the distributed Trojan attacks~\cite{xie2019dba} jointly to inject Trojan triggers into the global model.
We assume that attackers train their local models on the poisoned training data and contribute to the global model without scaling the original weight of the poisoned local models.
We then apply \sys on the global model to defend against the Trojan attack from malicious local models.
Specifically, \sys requires the participants to use their data to test the global model and upload the activation values of the global model to identify compromised neurons.
To measure the defense performance, we measure the BA and ASR of the global models (i.e., the original model and model deployed with \sys).
The results are shown in \autoref{tab:federal learning}.
As shown in the table, on average, \sys achieve 31.16 times lower ASR than undefended models with a slight decrease (i.e., 1.13\%) in the BA.
The results show that \sys can defend against the Trojan attacks effectively in real-world federated learning applications at a low cost.
Besides federated learning, we have also discussed the performance of \sys in transfer learning settings.
Hidden Trigger Trojan Attack~\cite{saha2020hidden} in~\autoref{sec:RQ1} of the main paper is conducted in transfer learning scenarios and the results are shown in~\autoref{tab:standard backdoor attack} of the main paper.
As the results show, \sys achieves low ASR (i.e., 4.00\%) and high BA (i.e., only 2.00\% lower than undefended models), proving the generalization of \sys on transfer learning settings.

\subsection{Precision and Recall of the Poisoned Sample Identification}\label{sec:precision_recall}
In line 12-17 of \autoref{alg:training}, we identify the poisoned samples in the training data.
To evaluate the effectiveness of the identification process, we measure the precision and the recall of detecting poisoned samples on the CIFAR-10 dataset and three different models (i.e., NiN, VGG16, and ResNet18).
Results in \autoref{tab:precision_recall} demonstrate the precision and the recall of \sys are always above 98\% on different settings.
For example, on ResNet18 and Label-consistent attack, both the precision and the recall of \sys are 100.00\%.
Thus, \sys can detect poisoned samples accurately.

\begin{table}[]
\centering
\scriptsize
\caption{Precision and Recall of Poisoned Samples Identification.}
\vspace{0.1cm}

\label{tab:precision_recall}
\setlength\tabcolsep{3pt}
\begin{tabular}{@{}cccccccccc@{}}
\toprule
\multirow{2}{*}{Attack} &  & \multicolumn{2}{c}{NiN} &  & \multicolumn{2}{c}{VGG16} &  & \multicolumn{2}{c}{ResNet18} \\ \cmidrule(lr){3-4} \cmidrule(lr){6-7} \cmidrule(l){9-10} 
                        &  & Precision   & Recall    &  & Precision    & Recall     &  & Precision     & Recall       \\ \midrule
BadNets                 &  & 99.60\%     & 99.64\%   &  & 99.96\%      & 100.00\%   &  & 99.84\%       & 99.92\%      \\
Label-consistent        &  & 98.80\%     & 99.20\%   &  & 100.00\%     & 100.00\%   &  & 100.00\%      & 100.00\%     \\ \bottomrule
\end{tabular}
\end{table}

\section{Discussion}\label{sec:discussion}
In this paper, we focus the discussion on image classification tasks, which is the focus of many existing works~\cite{gu2017badnets,saha2020hidden,turner2019label,liu2019abs,wang2019neural,bagdasaryan2021blind}.
Expanding our work, including the theory and system to other problem domains, such as natural language processing and reinforcement learning, other computer vision tasks, e.g., object detection, will be our future work.

Research on adversarial machine learning potentially has ethical concerns.
In this research, we propose a theory to explain existing phenomena and attacks, and propose a new training method that removes Trojans in a DNN model.
We believe this is beneficial to society.

In our current threat model, the adversary can only inject poisoning data into the training dataset, and there is no existing adaptive attacks.
However, adaptive adversaries can still conduct attacks under other threat models, and we discuss such case in \autoref{sec:appendix_adaptive_attack}.

\section{Conclusion}\label{sec:conclusion}
In this paper, we present an analysis on DNN Trojans and find relationships between decision regions and Trojans with a formal proof.
Moreover, we provide empirical evidence to support our theory.
Furthermore, we analyzed the reason why models will have such phenomena is because of linearity of trained layers.
Based on this, we propose a novel training method to remove Trojans during training, 
\sys{}, which can effectively and efficiently prevent intended and unintended Trojans. %

\section*{Acknowledgement}
We thank the anonymous reviewers for their valuable comments.
This research is supported by IARPA TrojAI W911NF-19-S-0012.
Any opinions, findings, and conclusions expressed in this paper are those of the authors only and do not necessarily reflect the views of any funding agencies.

\bibliographystyle{unsrtnat}
\bibliography{reference}

\section*{Checklist}

\begin{enumerate}

\item For all authors...
\begin{enumerate}
  \item Do the main claims made in the abstract and introduction accurately reflect the paper's contributions and scope?
    \answerYes{}
  \item Did you describe the limitations of your work?
    \answerYes{See \autoref{sec:discussion}.}
  \item Did you discuss any potential negative societal impacts of your work?
    \answerYes{See \autoref{sec:discussion}.}
  \item Have you read the ethics review guidelines and ensured that your paper conforms to them?
    \answerYes{See \autoref{sec:discussion}.}
\end{enumerate}

\item If you are including theoretical results...
\begin{enumerate}
  \item Did you state the full set of assumptions of all theoretical results?
    \answerYes{See \autoref{sec:formalana}.}
	\item Did you include complete proofs of all theoretical results?
    \answerYes{See Appendix.}
\end{enumerate}

\item If you ran experiments...
\begin{enumerate}
  \item Did you include the code, data, and instructions needed to reproduce the main experimental results (either in the supplemental material or as a URL)?
    \answerYes{See Abstract, \autoref{sec:evaluation}, and Appendix.}
  \item Did you specify all the training details (e.g., data splits, hyperparameters, how they were chosen)?
    \answerYes{See \autoref{sec:evaluation} and Appendix.}
	\item Did you report error bars (e.g., with respect to the random seed after running experiments multiple times)?
    \answerNo{}
	\item Did you include the total amount of compute and the type of resources used (e.g., type of GPUs, internal cluster, or cloud provider)?
    \answerYes{See \autoref{sec:evaluation}.}
\end{enumerate}

\item If you are using existing assets (e.g., code, data, models) or curating/releasing new assets...
\begin{enumerate}
  \item If your work uses existing assets, did you cite the creators?
    \answerYes{See \autoref{sec:evaluation}.}
  \item Did you mention the license of the assets?
    \answerYes{See Appendix.}
  \item Did you include any new assets either in the supplemental material or as a URL?
    \answerYes{URL for our code is included in Abstract.}
  \item Did you discuss whether and how consent was obtained from people whose data you're using/curating?
    \answerYes{See Appendix.}
  \item Did you discuss whether the data you are using/curating contains personally identifiable information or offensive content?
    \answerYes{See Appendix.}
\end{enumerate}

\item If you used crowdsourcing or conducted research with human subjects...
\begin{enumerate}
  \item Did you include the full text of instructions given to participants and screenshots, if applicable?
    \answerNA{}
  \item Did you describe any potential participant risks, with links to Institutional Review Board (IRB) approvals, if applicable?
    \answerNA{}
  \item Did you include the estimated hourly wage paid to participants and the total amount spent on participant compensation?
    \answerNA{}
\end{enumerate}

\end{enumerate}

\newpage
\section{Appendix}\label{sec:appendix}

\noindent
{\bf Roadmap:} In this appendix, we first list the symbols we used in this paper.
We show the proof and the empirical results for Theorem 3.3 (\autoref{sec:proof}), more empirical evidence for model's linearity (\autoref{sec:appendix_ee_activation}), explanation for DP-SGD (\autoref{sec:explain}), and more details about the sample separation (\autoref{sec:more_details_alg1}).
Then, we provide implementation details including details of datasets (\autoref{sec:details_datasets}) and used attacks (\autoref{sec:details_attack}).
We then discuss the the resistance of \sys against more attacks (\autoref{sec:resistance_more_attacks}), and the efficiency of \sys (\autoref{sec:efficiency}).
We measure the sensitivity of \sys against different configurable parameters (\autoref{sec:RQ3}).
We then evaluate \sys on an adaptive attack (\autoref{sec:appendix_adaptive_attack}).
In \autoref{sec:compare_to_dbd}, we compare \sys to another defense DBD~\cite{huang2022backdoor}. We also compare \sys to more defenses on natural Trojan in \autoref{sec:compare_to_more_natural_trojan}. Finally, we evaluate the generalization to larger models (\autoref{sec:generalization_larger_models}) and larger datasets (\autoref{sec:generalization_to_larger_datasets}).

\noindent
{\bf Symbol Table.}
\begin{table}[H]
\centering
\scriptsize
\caption{Summary of Symbols}\label{tab:summary_symbols_alg1}
\vspace{0.1cm}
\begin{tabular}{@{}ccc@{}}
\toprule
Scope                       & Symbol                             & Meaning                                                \\ \midrule
\multirow{17}{*}{Theory}    & $\bm{x}$                           & Benign Sample                                          \\
                            & $\tilde{\bm{x}}$                   & Trojan Sample                                          \\
                            & $T$                                & Trojan Sample Generation Function                      \\
                            & $\bm{m}$                           & Mask of Trojan Trigger                                 \\
                            & $\bm{t}$                           & Pattern of Trojan Trigger                              \\
                            & $\odot$                            & Hadamard product                                       \\
                            & $\mathcal{M}$                      & Model                                                  \\
                            & $\mathcal{X}$                      & Input Domain                                           \\
                            & $\mathcal{Y}$                      & Set of Labels                                          \\
                            & $\mathcal{R}^{l}$                  & Decision Region of Label $l$                           \\
                            & $\mathcal{T}$                      & Trojan Region                                          \\
                            & $m$                                & The Number of Elements in $\bm m$                      \\
                            & $\{{\bm A}{\bm x} - {\bm b} = 0\}$ & Trojan Hyperplane                                      \\
                            & $x_j$                              & Inputs of Layer $j$                                    \\
                            & $y_j$                              & Outputs of Layer $j$                                   \\
                            & $\mathbf{W}_j$                     & Trained Weights of Layer $j$                           \\
                            & $\mathbf{b}_j$                     & Trained Bias of Layer $j$                              \\ \midrule
\multirow{14}{*}{Algorithm} & $D$                                & Training Data                                          \\
                            & $E$                                & Maximal Epoch                                          \\
                            & $e$                                & Current Epoch                                          \\
                            & $M$                                & Model                                                  \\
                            & $n$                                & Neuron                                                 \\
                            & $A$                                & Activation Values                                      \\
                            & $A_n$                              & Activation values on Neuron $n$                        \\
                            & $C$                                & Compromised Neurons                                    \\
                            & $B_n$                              & The Cluster of Smaller Values in $A_n$                 \\
                            & $O_n$                              & The Cluster of Larger Values in $A_n$                  \\
                            & $\mu$                              & Mean Value of $B_n$                                    \\
                            & $\sigma$                           & Standard Deviation Value of $B_n$                      \\
                            & $i$                                & Input Sample                                           \\
                            & $i_n$                              & The Activation Value of Input Sample $i$ on Neuron $n$ \\ \bottomrule
\end{tabular}
\end{table}

\subsection{Proof and Empirical Evidence for Theorem 3.3}\label{sec:proof}

We start our analysis from ideal Trojan attacks, which we define as complete and precise Trojans:

\begin{definition}\label{def:definition1}
    \textbf{Complete Trojan}: For a Trojaned model \(\mathcal{M}: \mathcal{X} \mapsto \mathcal{Y}\) with trigger \((\bm{m}, \bm{t})\) and target label \(l\), we say a Trojan is complete if \(\forall \bm x \in T(\mathcal{X}, {\bm m}, {\bm t}), \mathcal{M}(\bm x) = l\).
\end{definition}

\begin{definition}\label{def:definition1}
    \textbf{Precise Trojan}: For a Trojaned model \(\mathcal{M}: \mathcal{X} \mapsto \mathcal{Y}\) with trigger \((\bm{m}, \bm{t})\) and target label \(l\), we say a Trojan is precise if the follow condition is met: \(\forall (\bm{m}^{\prime},\bm{t}^{\prime}) \neq (\bm{m},\bm{t}), \bm{x}^{\prime} = T(\bm{x}, {\bm m}^{\prime}, {\bm t}^{\prime}), \mathcal{M}(\bm{x}) \neq l \Rightarrow \mathcal{M}(\bm{x}^{\prime}) \neq l\).
\end{definition}

Intuitively, a complete Trojan means the attack success rate of this attack is 100\%, and a precise Trojan means that the trigger is unique: if we change the trigger (\(\bm{t}\) or \(\bm{m}\)), it will not trigger the predefined misclassification.

\begin{proof}
    In \autoref{th:perfect_trojan1}, we have \(\textbf{S\(_0\)} \iff \textbf{S\(_1\)} \) where:
    \begin{itemize}
        \item \textbf{S\(_0\)}: Trojan in \(\mathcal{M}\) with trigger being \((\bm{m}, \bm{t})\) and target label being \(l\) is a complete and precise Trojan.
        \item \textbf{S\(_1\)}: The hyperplane \(\{{\bm A}{\bm x} - {\bm b} = 0\}\) is the Trojan region of \(\mathcal{M}\) and the only one, where \(i \in \{1\ldots m\}\), diagonal matrix \({{\bm A}_{i,i}} = {\bm m}_i, \bm b = {\bm A}{\bm t}\).
    \end{itemize}

    In this proof, we first prove \( \bm x \in T(\mathcal{X}, {\bm m}, {\bm t}) \iff \bm x \in \{{\bm A}{\bm{x}} - {\bm b} = 0\}\), and then prove \(\textbf{S\(_0\)} \Rightarrow  \textbf{S\(_1\)} \) and \(\textbf{S\(_1\)} \Rightarrow  \textbf{S\(_0\)}\).

    \smallskip\noindent\textbf{Step 1:}
    \( \bm x \in T(\mathcal{X}, {\bm m}, {\bm t}) \Rightarrow \bm x \in \{{\bm A}{\bm{x}} - {\bm b} = 0\}\).    Let \(\tilde{\bm{x}}\) be a Trojan input generated from \(\bm{x}\) by applying \autoref{eq:define_trojan_sample} (in \autoref{sec:observation} of the main paper), 
    \(\small \tilde{\bm{x}} = T(\bm{x}, {\bm m}, {\bm t}) = ({\bm 1}-{\bm m}) \odot \bm{x} + {\bm m} \odot \bm{t}\), we get:

    \begin{equation}\label{eq:proof_in_plane_0}
        {\bm A}\tilde{\bm{x}} - {\bm b} = {\bm A}(({\bm 1}-{\bm m}) \odot \bm{x} + {\bm m} \odot \bm{t}) - {\bm b}
    \end{equation}

    Then, based on the definition of matrix \(\bm A\), \(\bm b\), the Hadamard product, and the distributive property of matrix multiplication, we can get the following equation, where \({\bm E}\) is the identity matrix:
    \begin{equation}\label{eq:proof_in_plane_1}
        \small
        \begin{aligned}
            {\bm A}(({\bm 1}-{\bm m}) \odot \bm{x} + {\bm m} \odot \bm{t}) - {\bm b}
            &={\bm A}(({\bm E}-{\bm A}) \bm{x} + {\bm A} \bm{t}) - {\bm A}{\bm t}\\
            &={\bm A}({\bm E}-{\bm A}) \bm{x} + {\bm A} {\bm A} \bm{t} - {\bm A}{\bm t}
            \end{aligned}
    \end{equation}

    Since \(\bm A\) is a diagonal matrix, and all elements of \(\bm A\) is 0 or 1 based on the definition of \(\bm A\) and trigger mask \(\bm m\), we can get \(\bm A \bm A = \bm A\) and \(\bm A (\bm E - \bm A) = 0\). 
    Then, according to \autoref{eq:proof_in_plane_0} and \autoref{eq:proof_in_plane_1}, we get: \(\forall \bm x \in T(\mathcal{X}, {\bm m}, {\bm t}),\;\; {\bm A}{\bm x} - {\bm b} = 0\).

    \noindent\textbf{Step 2:}
    \( \bm x \in \{{\bm A}{\bm{x}} - {\bm b} = 0\} \Rightarrow \bm x \in T(\mathcal{X}, {\bm m}, {\bm t})\).
    This step is to prove that any sample in the hyperplane \(\{{\bm A}{\bm{x}} - {\bm b} = 0\}\) can be obtained from pasting Trojan trigger on other samples.
    Let \(\tilde{\bm{x}}\) denote any sample in the hyperplane, and \({\bm x}\) is the sample that is not in the hyperplane, i.e., an external sample. 
    Any external sample \(\bm{x}\) that satisfies \(({\bm E}-{\bm A}) \bm{x} = ({\bm E}-{\bm A}) \tilde{\bm{x}}\) can be transformed to \(\tilde{\bm{x}}\) via the projection specified by \({\bm m}\) and \({\bm t}\).
    Therefore, we conclude that any sample in the hyperplane can be obtained by pasting the Trojan trigger on other samples.

    Steps 1 and 2 prove that \(\bm x\in T(\mathcal{X}, {\bm m}, {\bm t})\) is equivalent to \(\bm x\) in the hyperplane \({\bm A}{\bm x} - {\bm b} = 0\), namely:
    \begin{equation}\label{eq:proof_space}
         \bm x \in T(\mathcal{X}, {\bm m}, {\bm t}) \iff \bm x \in \{{\bm A}{\bm{x}} - {\bm b} = 0\}
    \end{equation}
    
    \noindent\textbf{Step 3:}
    \(\textbf{S\(_0\)} \Rightarrow  \textbf{S\(_1\)}\).
    Based on \(\textbf{S\(_0\)}\), Trojan in \(\mathcal{M}\) is a complete Trojan. 
    Based on \autoref{eq:proof_space} and the definition of complete Trojan (i.e., \autoref{def:definition1}), we get: \(\forall \bm x \in \{{\bm A}\bm{x} - {\bm b} = 0\}, \mathcal{M}(\bm x) = l\), which means \(\{\bm A\bm{x} - {\bm b} = 0\}\) is a Trojan decision region.
    We then prove \(\{\bm A\bm{x} - {\bm b} = 0\}\) is the only Trojan region using proof by contradiction.
    For any other hyperplane \(\{{\bm A}^{\prime}\bm{x} - {\bm b}^{\prime} = 0\}\) where \(({\bm A}^{\prime},{\bm b}^{\prime}) \neq ({\bm A},{\bm b})\), based on \autoref{eq:proof_space}, we can get: \((\bm{m}^{\prime},\bm{t}^{\prime}) \neq (\bm{m},\bm{t}),
    \bm{x}^{\prime} = T(\bm{x}, {\bm m}^{\prime}, {\bm t}^{\prime}) \iff {\bm A}^{\prime}\bm{x}^{\prime} - {\bm b}^{\prime} = 0\).
    According to \(\textbf{S\(_0\)}\), the Trojan is a precise Trojan: \(\forall (\bm{m}^{\prime},\bm{t}^{\prime}) \neq (\bm{m},\bm{t}), {\bm x}^{\prime} = T({\bm x}, {\bm m}^{\prime}, {\bm t}^{\prime}), \mathcal{M}({\bm x}) \neq l \Rightarrow \mathcal{M}({\bm x}^{\prime}) \neq l\).
    Thus, we get that \({\bm A}^{\prime}\bm{x} - {\bm b}^{\prime} = 0\) is not a Trojan region.
    That is, the Trojan region has only one hyperplane, \(\{\bm A\bm{x} - {\bm b} = 0\}\).

    \noindent\textbf{Step 4:}
    \(\textbf{S\(_1\)} \Rightarrow  \textbf{S\(_0\)}\).
    According to \(\textbf{S\(_1\)}\), we have:
    \begin{equation}\label{eq:s1_0}
        \forall \bm x \in \{{\bm A}\bm{x} - {\bm b} = 0\}, \mathcal{M}({\bm x}) = l
    \end{equation}
    \begin{equation}\label{eq:s1_1}
        \small
        \forall {({\bm A}^{\prime},{\bm b}^{\prime}) \neq (\bm A, {\bm b}),
        {\bm A}^{\prime}\bm{x}^{\prime} - {\bm b}^{\prime} = 0}, \mathcal{M}({\bm x}) \neq l \Rightarrow \mathcal{M}({\bm x}^{\prime}) \neq l
    \end{equation}
    
    From \autoref{eq:proof_space} and \autoref{eq:s1_0}, we get \(\forall {\bm x \in T(\mathcal{X}, {\bm m}, {\bm t})}, \mathcal{M}({\bm x}) = l\), which means the Trojan is complete.
    Based on \autoref{eq:proof_space} and \autoref{eq:s1_1}, we can get 
    \(
        \forall (\bm{m}^{\prime} , \bm{t}^{\prime}) \neq (\bm{m},\bm{t}), {\bm x}^{\prime} = T({\bm x}, {\bm m}^{\prime}, {\bm t}^{\prime}), \mathcal{M}({\bm x}) \neq l \Rightarrow  \mathcal{M}({\bm x}^{\prime}) \neq l\), where \(\bm{m}^{\prime} = \bm {A}_{i,i}^{\prime}, \bm{b}^{\prime} = \bm {A}^{\prime}\bm {t}^{\prime},
    \)
    indicating that the Trojan is precise.

    From Step 3 and 4, we can conclude that \(\textbf{S\(_0\)} \iff \textbf{S\(_1\)} \), and complete the proof of \autoref{th:perfect_trojan1}.
\end{proof}

Intuitively, the Trojan is precise means the attack success rate is 100\% which guarantees that all samples with the trigger will be classified as the target label.
The Trojan is complete means that no other input patterns can trigger this trigger, and thus all inputs that activate this Trojan have this trigger.
In the real world, these are hard to achieve.
In practice, a Trojan of model \(\mathcal{M}\) whose trigger is \((\bm{m}, \bm{t})\) and target label is \(l\) has 
\begin{gather}
\exists (\bm{m}^{\prime}, \bm{t}^{\prime}) \approx (\bm{m}, \bm{t}), \mathbb{P}(\mathcal{M}(T(\bm{x}, \bm{m}^{\prime}, \bm{t}^{\prime})) = l) > \lambda \\
\mathbb{P}(\mathcal{M}(T(\bm{x}, \bm{m}, \bm{t})) = l) < 1, \bm x \in \mathcal{D}
\end{gather}
where \(\mathcal{D}\) is the dataset, and \(\lambda\) is a threshold value for the attack success rate (e.g., 90\%).
Namely, in the real world, a Trojan trigger cannot guarantee a 100\% attack success rate and the model can learn a trigger that is different from the intended one.
Consequently, the real Trojan region \(\mathcal{T^{\prime}}\) and the theoretical one \(\mathcal{T}\) satisfy \(\frac{|\mathcal{T^{\prime}}\cap \mathcal{T}|}{|\mathcal{T}|} = \alpha\) where \(\alpha\) is the real attack success rate.

\begin{figure*}[]
    \centering
    \footnotesize
    \begin{subfigure}[t]{0.24\columnwidth}
        \centering     
        \footnotesize
        \includegraphics[width=\columnwidth]{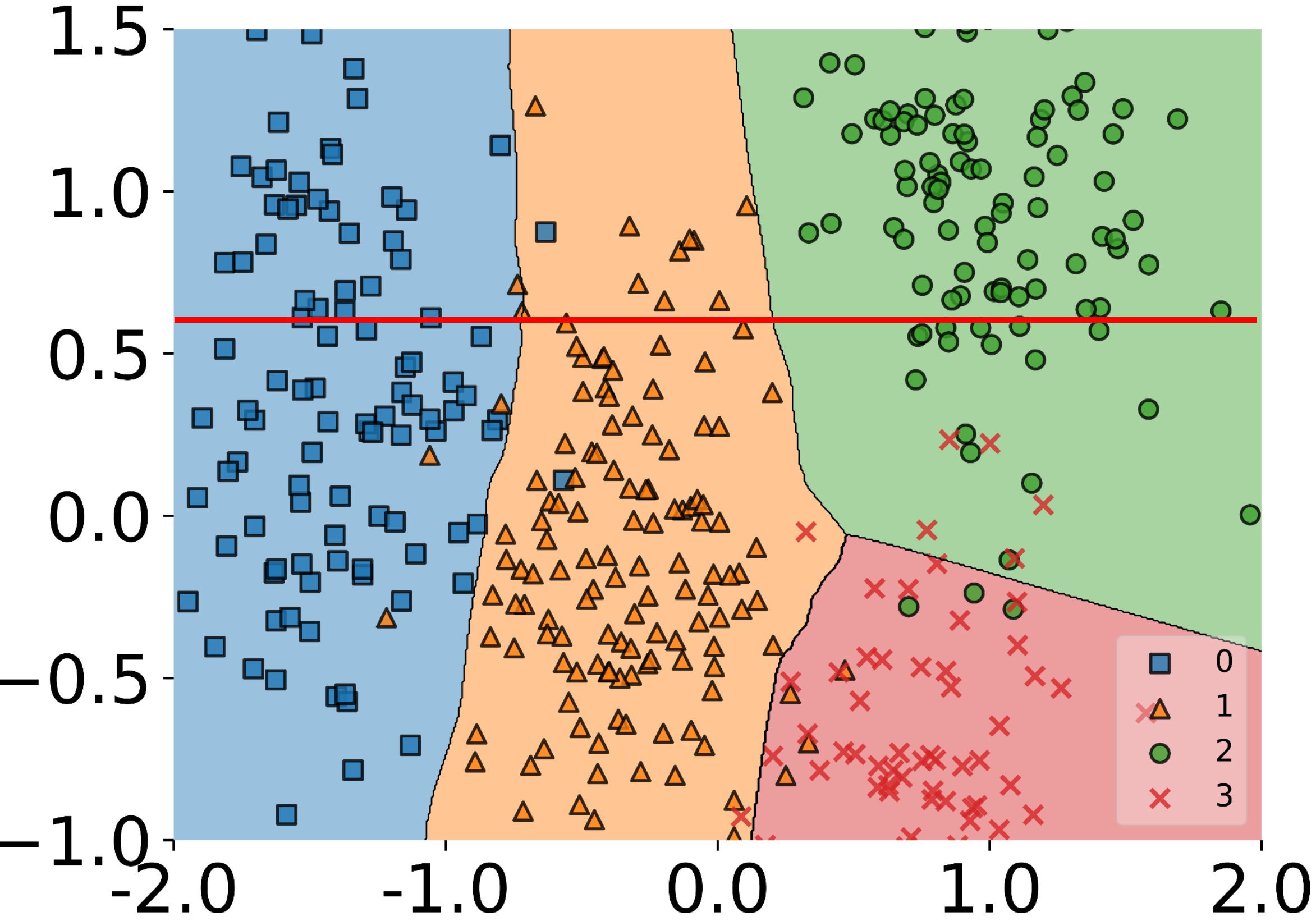}
        \caption{Perfect Trojan}
        \label{fig:perfect_trojan_perfect}
    \end{subfigure}
    \begin{subfigure}[t]{0.24\columnwidth}
        \centering     
        \footnotesize
        \includegraphics[width=\columnwidth]{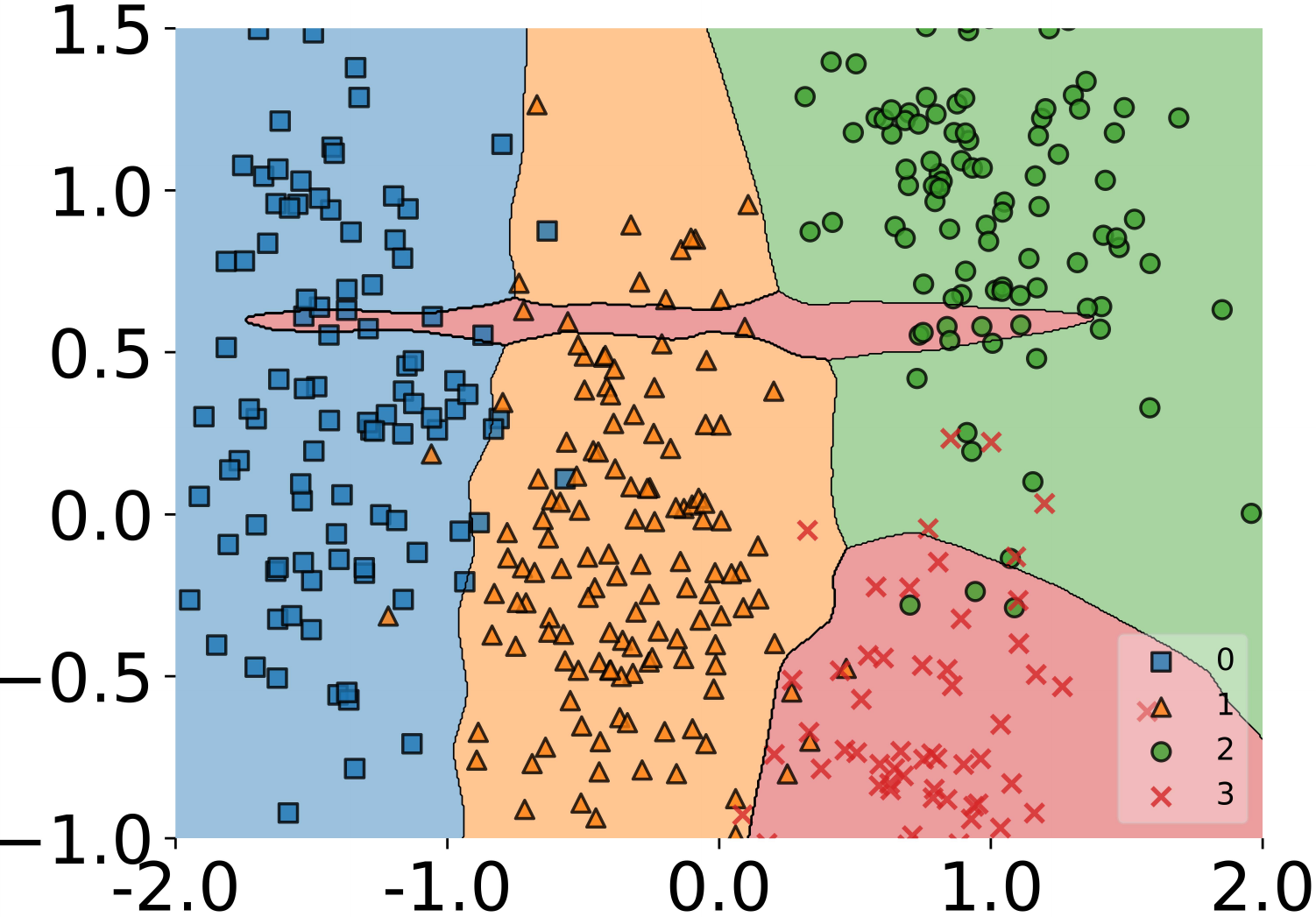}
        \caption{Real Case: 7.5\% Poisoned}
        \label{fig:perfect_trojan_r7.5}
    \end{subfigure}
    \begin{subfigure}[t]{0.24\columnwidth}
        \centering
           \footnotesize
           \includegraphics[width=\columnwidth]{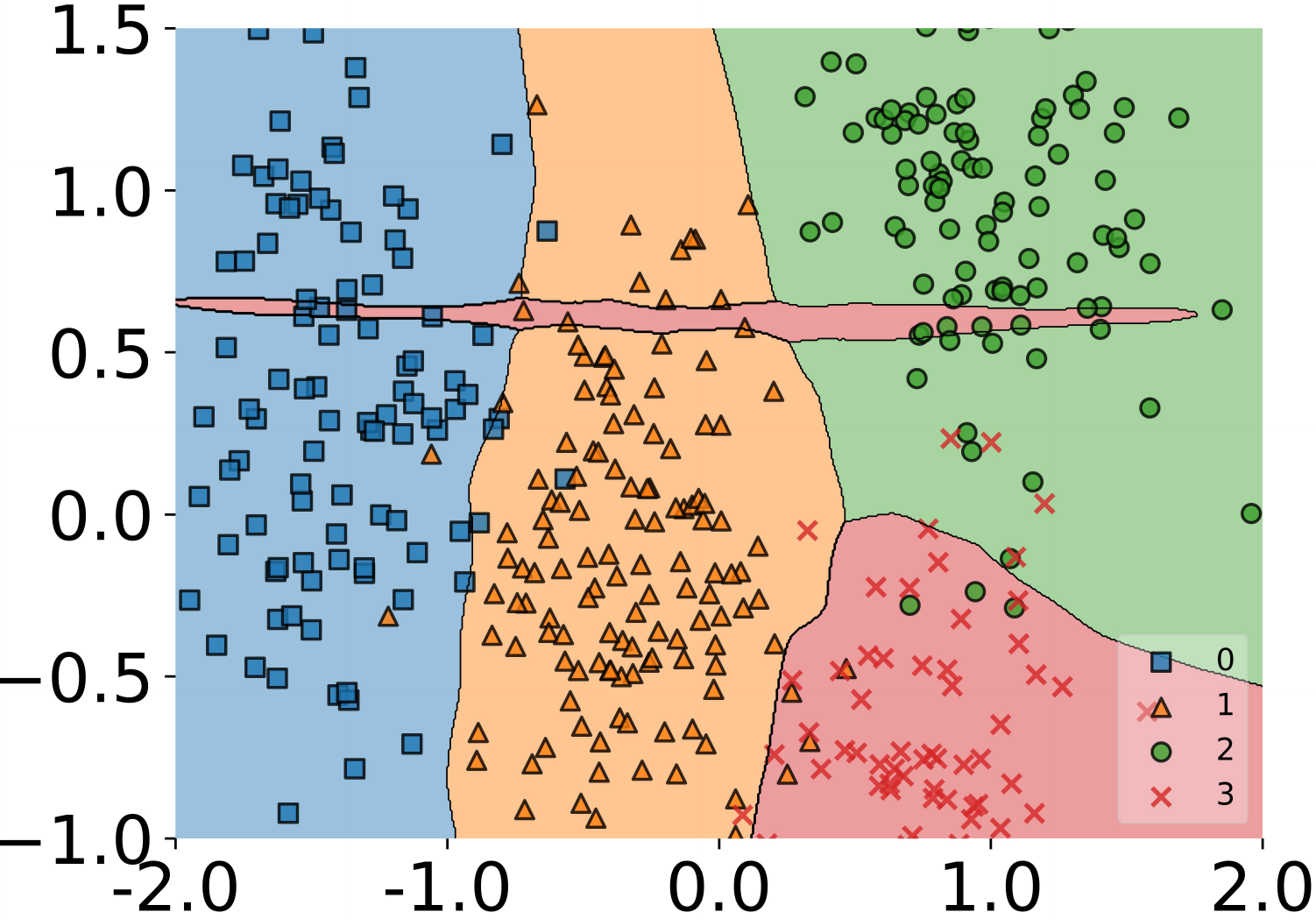}
           \caption{Real Case: 10.0\% Poisoned}
           \label{fig:perfect_trojan_r10}
    \end{subfigure}
    \begin{subfigure}[t]{0.24\columnwidth}
        \centering
           \footnotesize
           \includegraphics[width=\columnwidth]{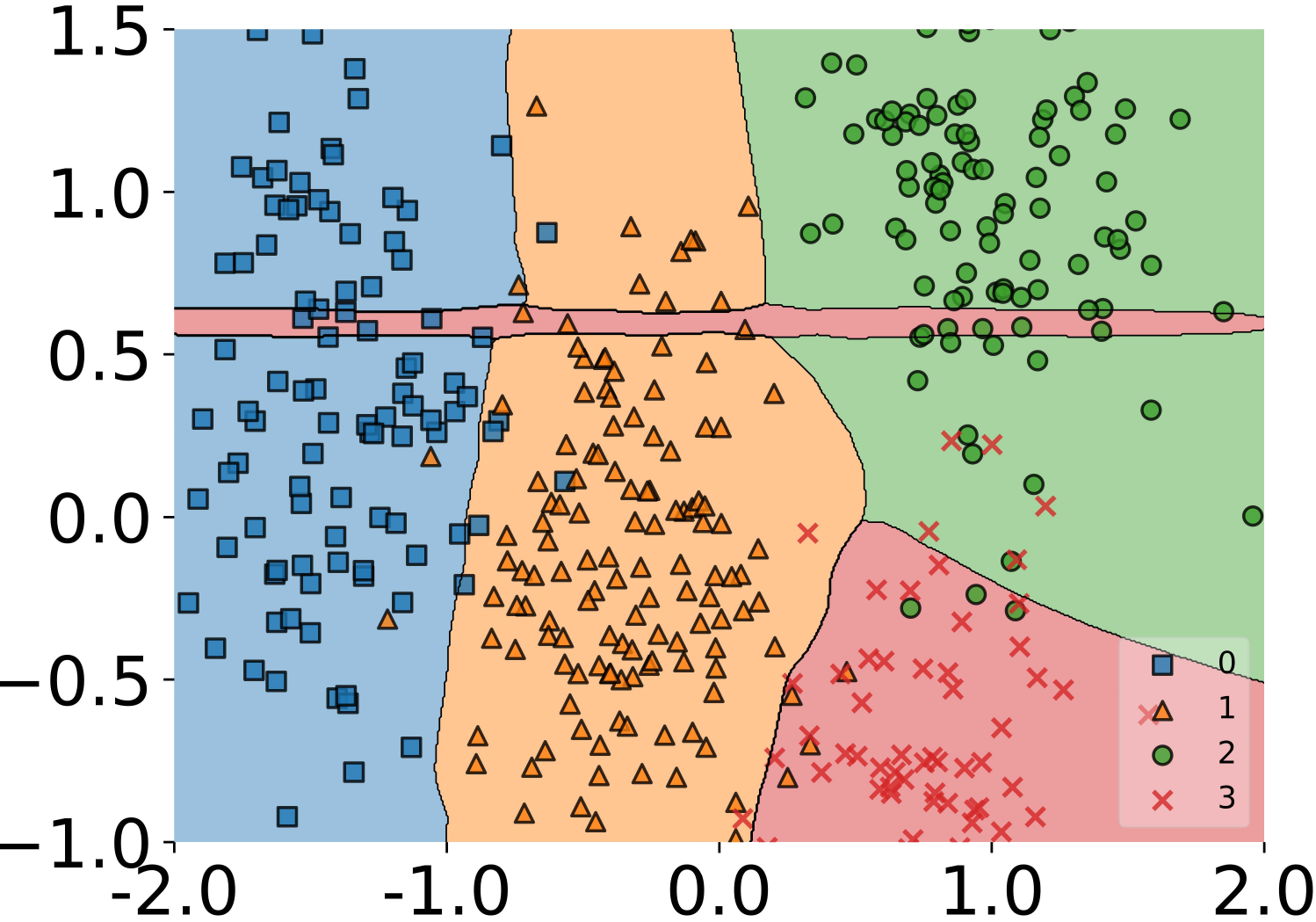}
           \caption{Real Case: 15.0\% Poisoned}
           \label{fig:perfect_trojan_r15}
    \end{subfigure}
    \caption{Perfect Trojans and Relaxations on 2D Data.
    Each sub-figure contains test samples (dots) and the learned decision regions for different labels (in different colors) under a specific setting.
    The Trojan trigger is \(t = (-, 0.6)\), and the target label \(y_t = 3\).
    The red region near (-, 0.6) is the learned Trojan decision region.}    \label{fig:perfect_trojan_relaxation} %
\end{figure*}

To evaluate if the Trojan decision region in real-world data is the relaxation of the Trojan linear hyperplane, we visualize the decision regions of Trojaned neural networks. 

Following Bai et al.~\cite{bai2021targeted}, we visualize the decision region of neural networks on 2d data.
Specifically, We visualize decision regions of compromised Multilayer Perceptrons (MLP) trained on different poisoning rates.
The MLP model has 5 layers and each layer contains 100 neurons, and we use ReLU as the activation function. 
Similar to Bai et al.~\cite{bai2021targeted}, the used dataset contains five isotropic Gaussian 2d blobs, in which each blob represents a class.
In~\autoref{fig:perfect_trojan_relaxation}, we show the complete and precise Trojan decision region (\autoref{fig:perfect_trojan_perfect}) for this model and real-world relaxations with different poisoning rates of BadNets attack (\autoref{fig:perfect_trojan_r7.5}, \autoref{fig:perfect_trojan_r10}, \autoref{fig:perfect_trojan_r15}). 
Each color in the figure denotes one output label.
In our experiments, we set the trigger to \(t = (-, 0.6)\), and the target class \(y_t = 3\) (red).
Thus, the red region close to \(t = (-, 0.6)\) denotes the Trojan region.
We observe that, with the growth of the poisoning ratio, the attacks get a higher attack success rate and become more precise, and the Trojan region also converts to the ideal one shown in \autoref{fig:perfect_trojan_perfect}.
Despite such relaxations, we can also confirm that the Trojan region has a large intersection with the hyperplane and other possible triggers are around the ground truth one.

\begin{figure*}[tb]
    \captionsetup[subfigure]{labelformat=empty}
    \centering
    \subfloat[(a) NiN]
    {
    \subfloat[Compromised]{\includegraphics[width=0.24\columnwidth]{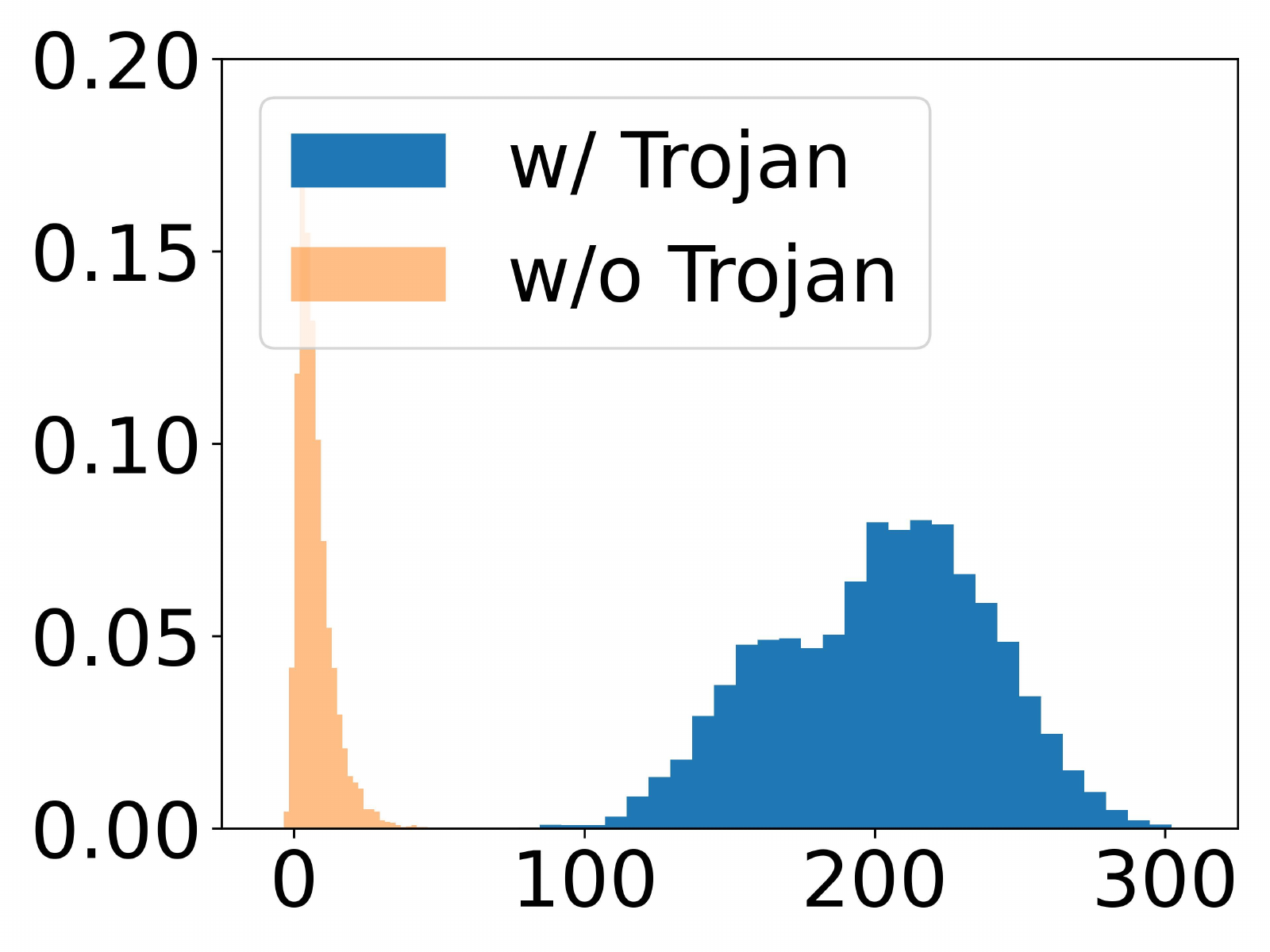}}
    \subfloat[Benign]{\includegraphics[width=0.24\columnwidth]{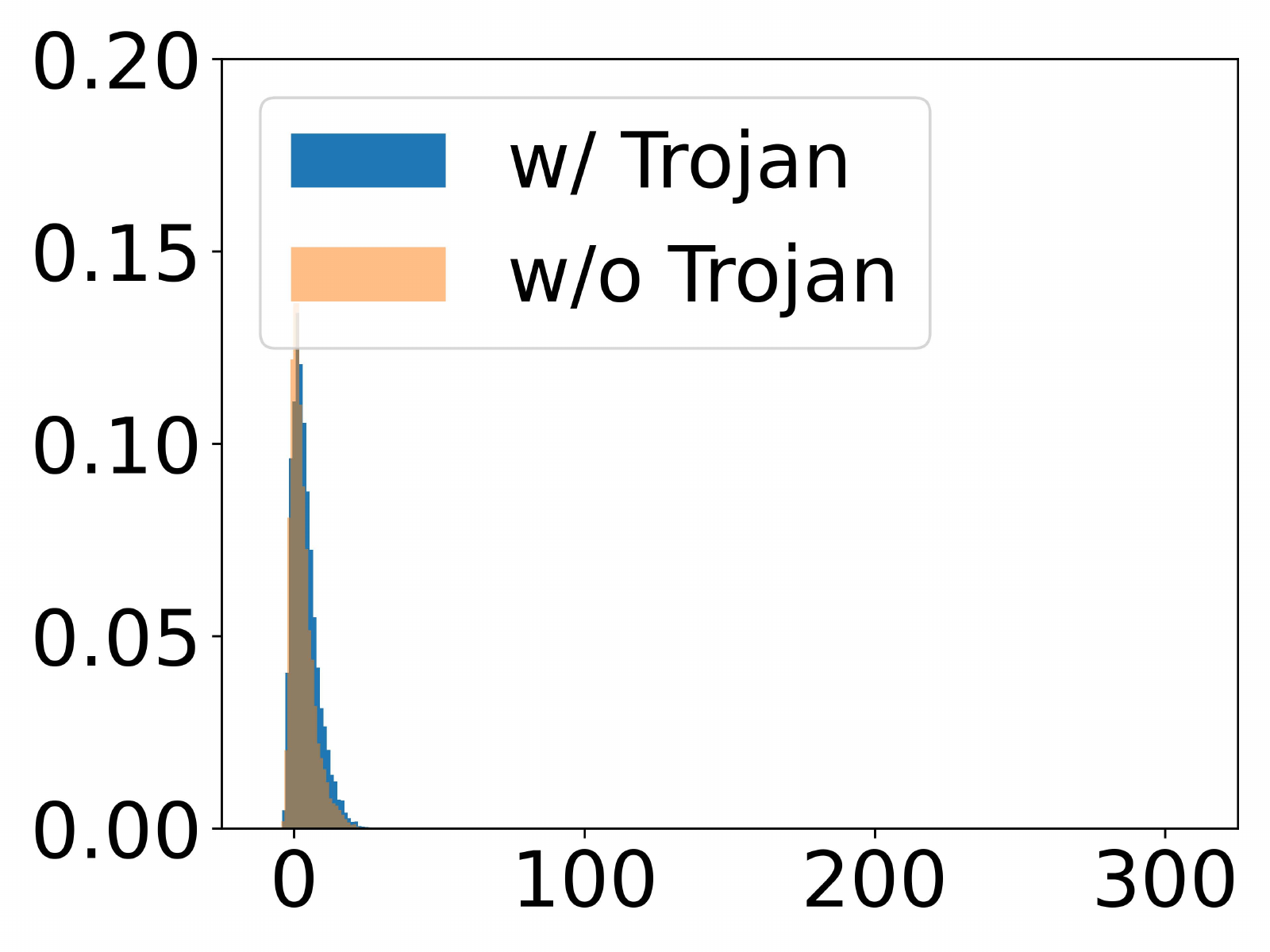}}
    }
    \hfill
    \subfloat[(b) VGG16]
    {\setcounter{subfigure}{0}
    \subfloat[Compromised]{\includegraphics[width=0.24\columnwidth]{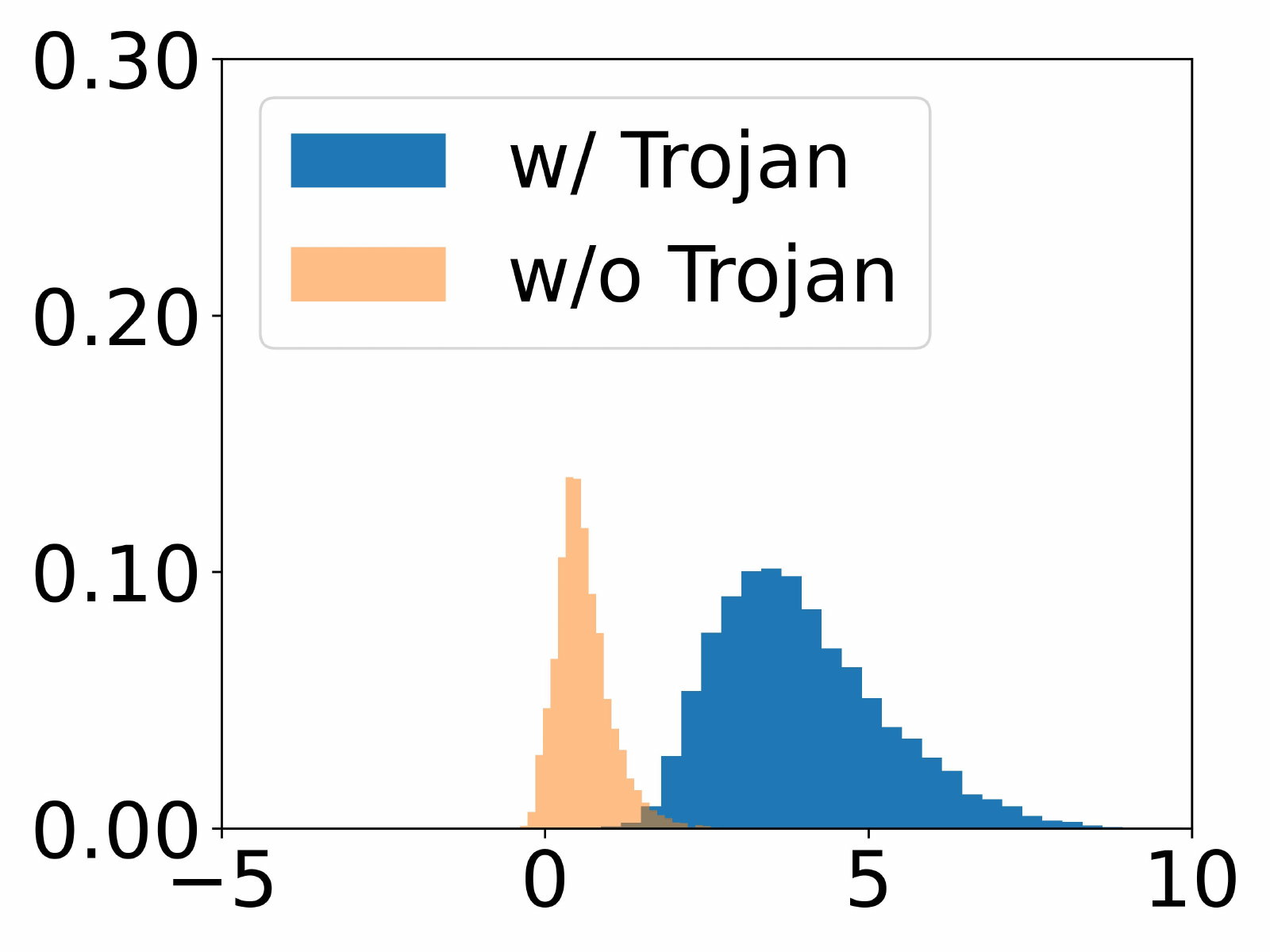}}
    \subfloat[Benign]{\includegraphics[width=0.24\columnwidth]{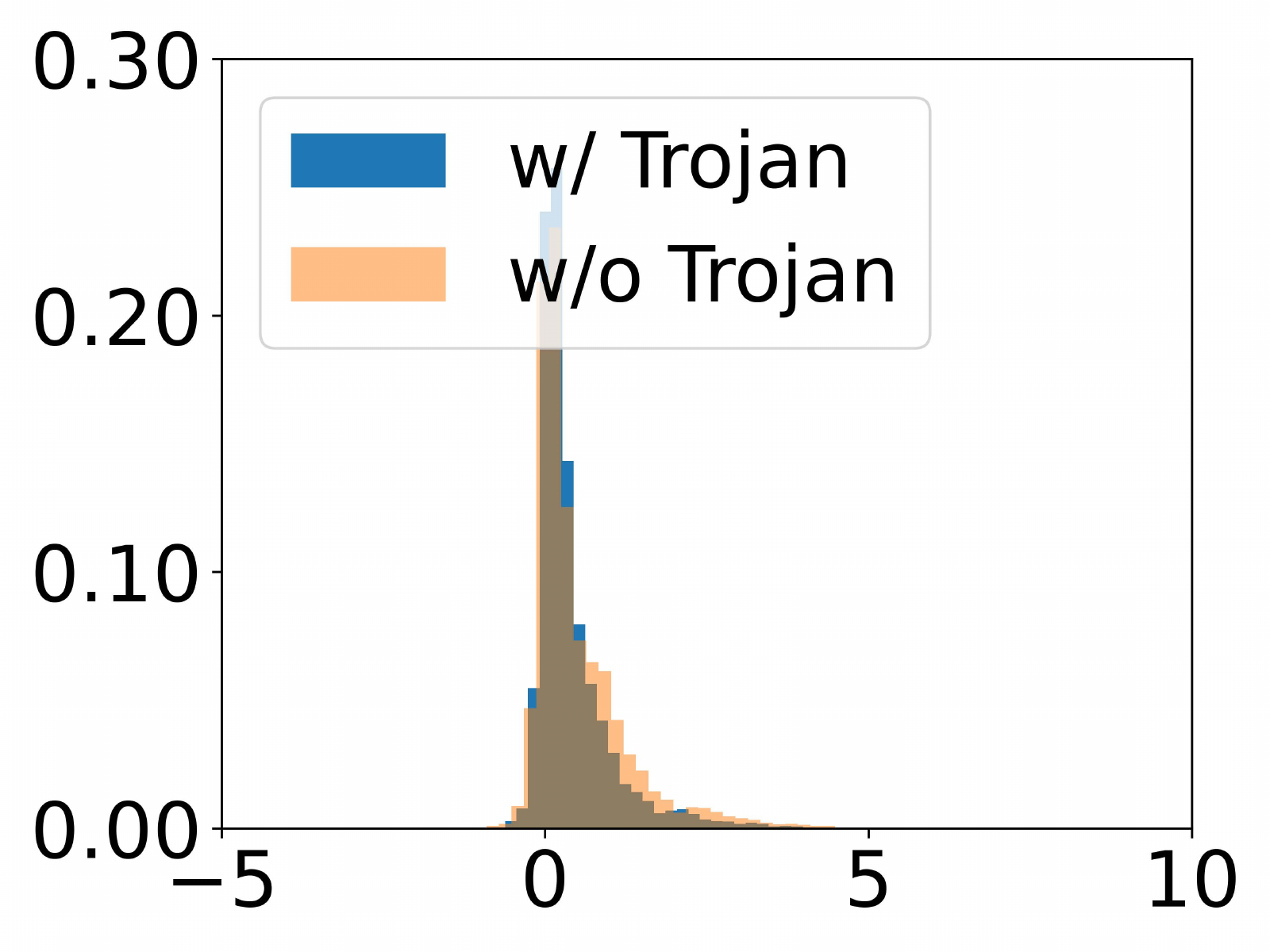}}
    }
    \hfill
    \caption{Comparison of Activation Values on Different Network Architectures.}\label{fig:compare_different_arch}
\end{figure*}

\begin{figure*}[tb]
    \captionsetup[subfigure]{labelformat=empty}
    \centering
    \subfloat[(a) \nth{14} Layer]
    {
    \subfloat[Compromised]{\includegraphics[width=0.24\columnwidth]{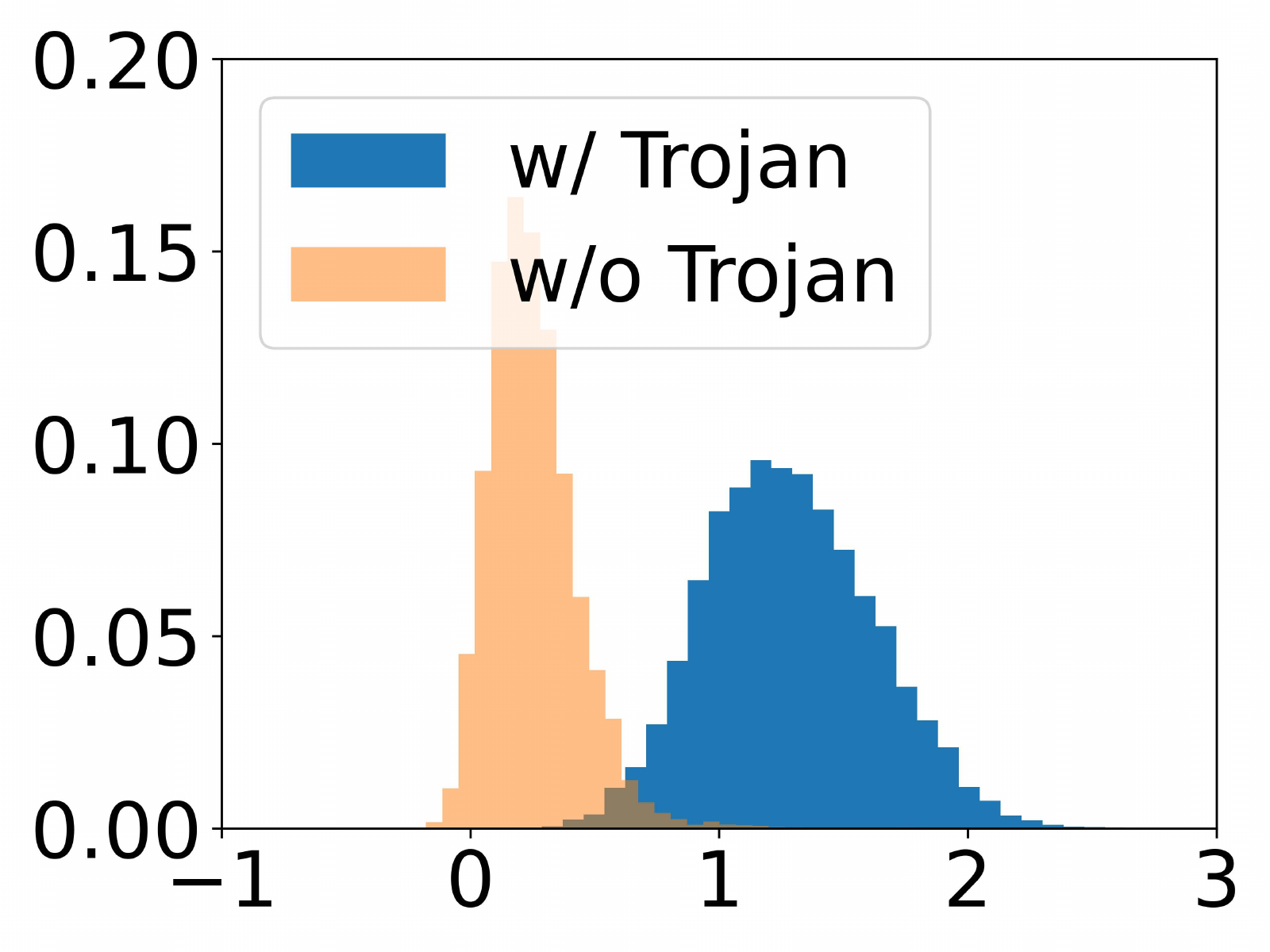}}
    \subfloat[Benign]{\includegraphics[width=0.24\columnwidth]{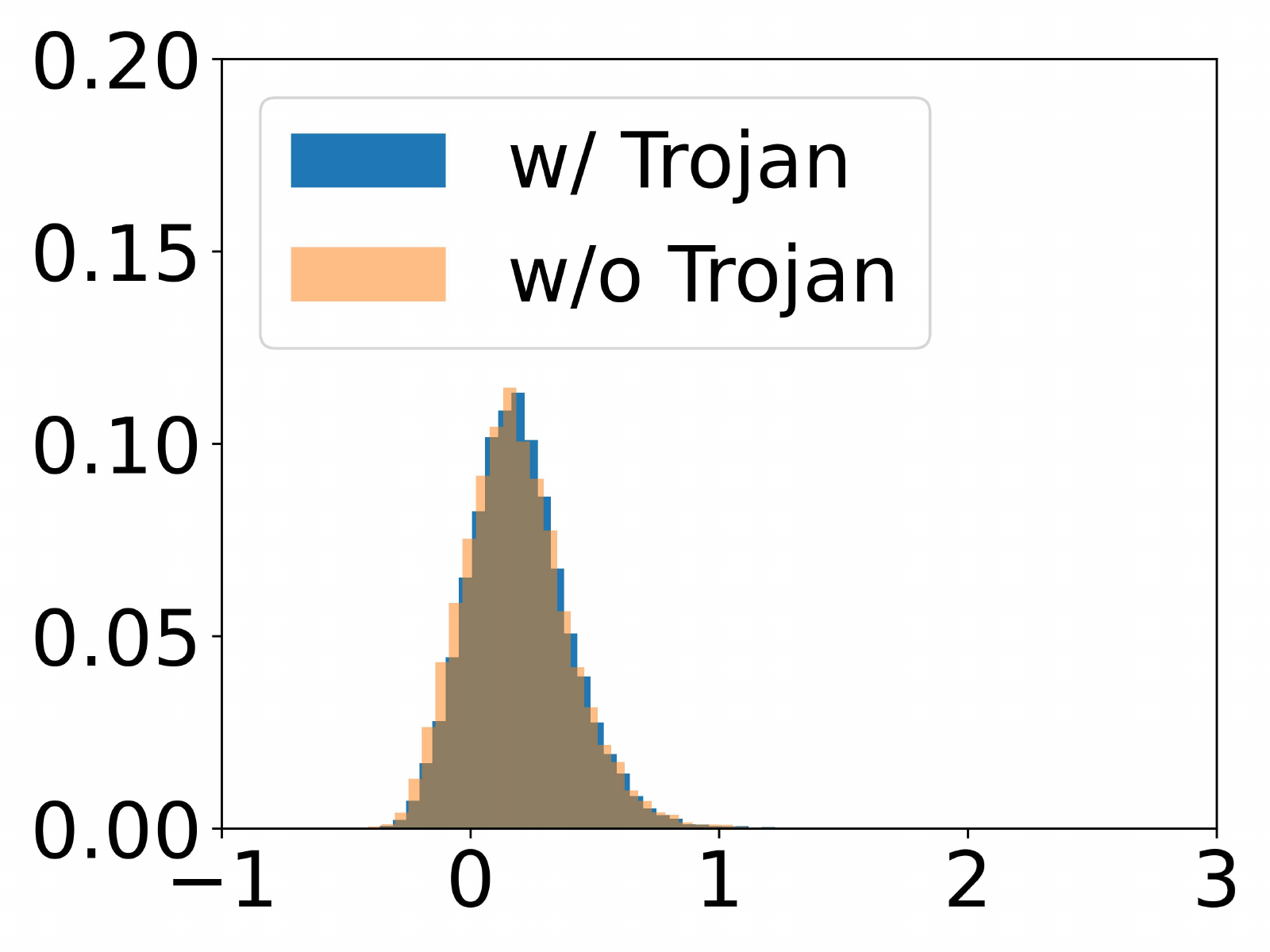}}
    }
    \hfill
    \subfloat[(b) \nth{15} Layer]
    {
    \subfloat[Compromised]{\includegraphics[width=0.24\columnwidth]{fig/cc.pdf}}
    \subfloat[Benign]{\includegraphics[width=0.24\columnwidth]{fig/bc.pdf}}
    }
    \hfill
    \subfloat[(c) \nth{16} Layer]
    {
    \subfloat[Compromised]{\includegraphics[width=0.24\columnwidth]{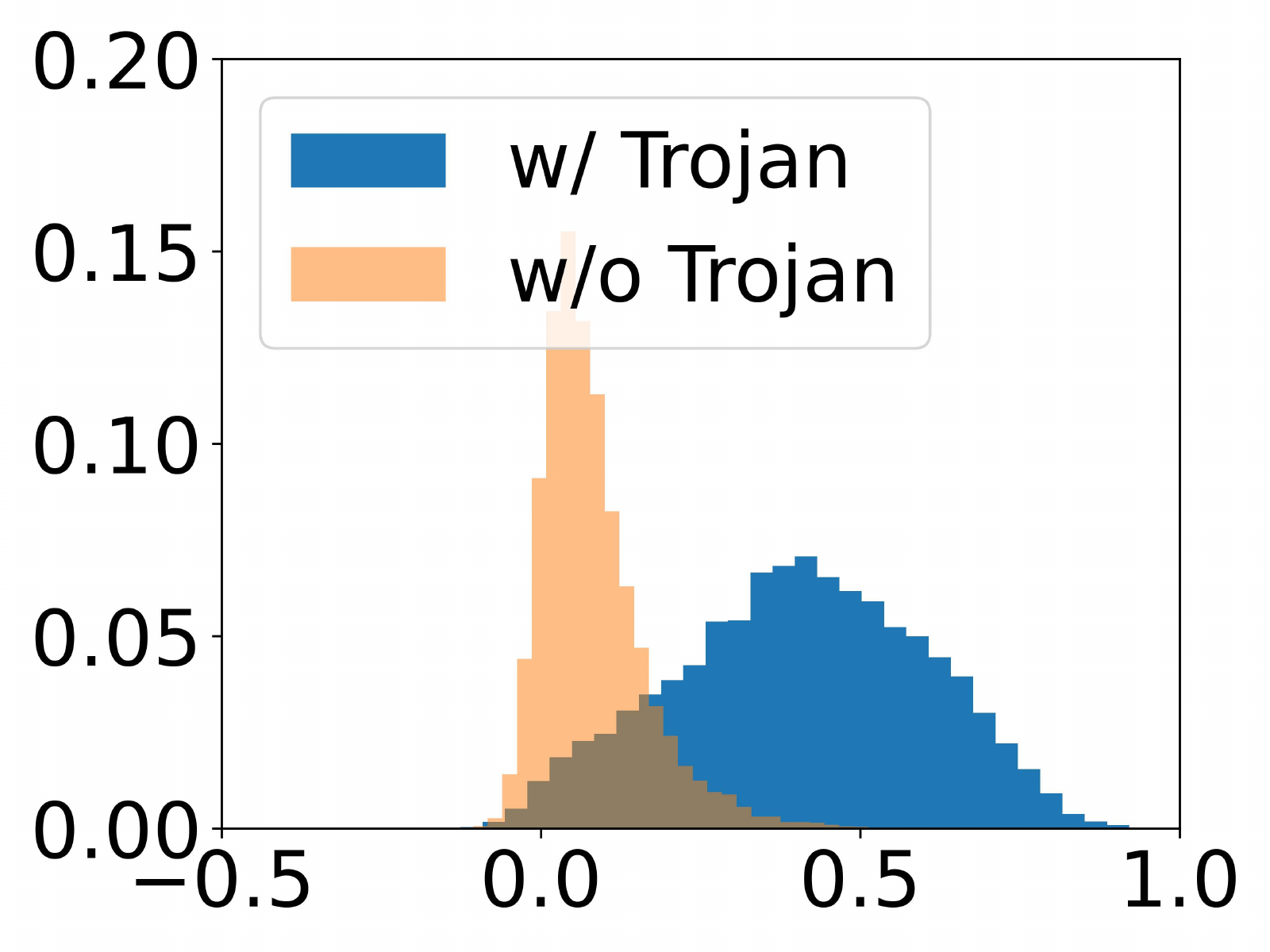}}
    \subfloat[Benign]{\includegraphics[width=0.24\columnwidth]{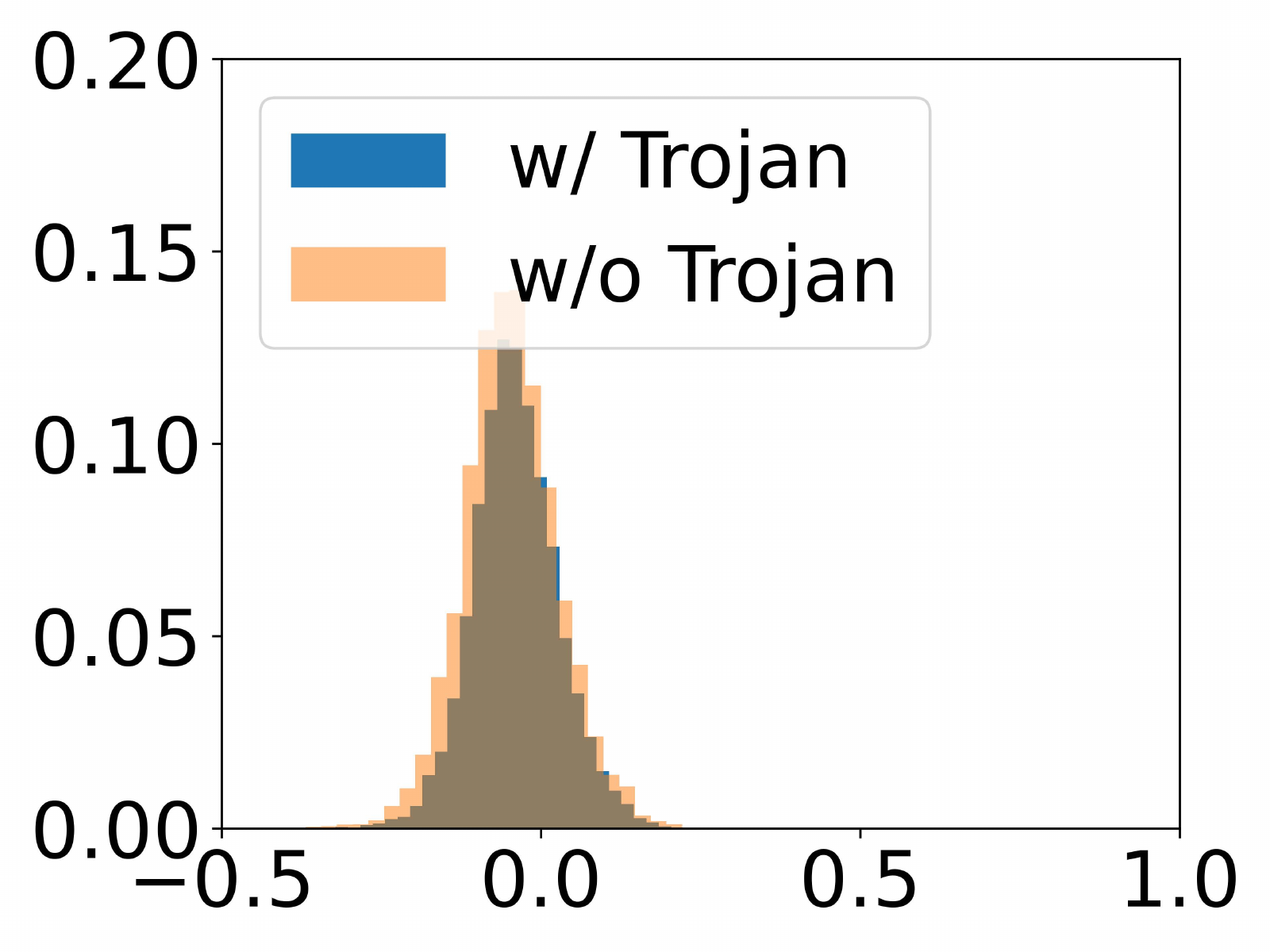}}
    }
    \hfill
    \subfloat[(d) \nth{17} Layer]
    {
    \subfloat[Compromised]{\includegraphics[width=0.24\columnwidth]{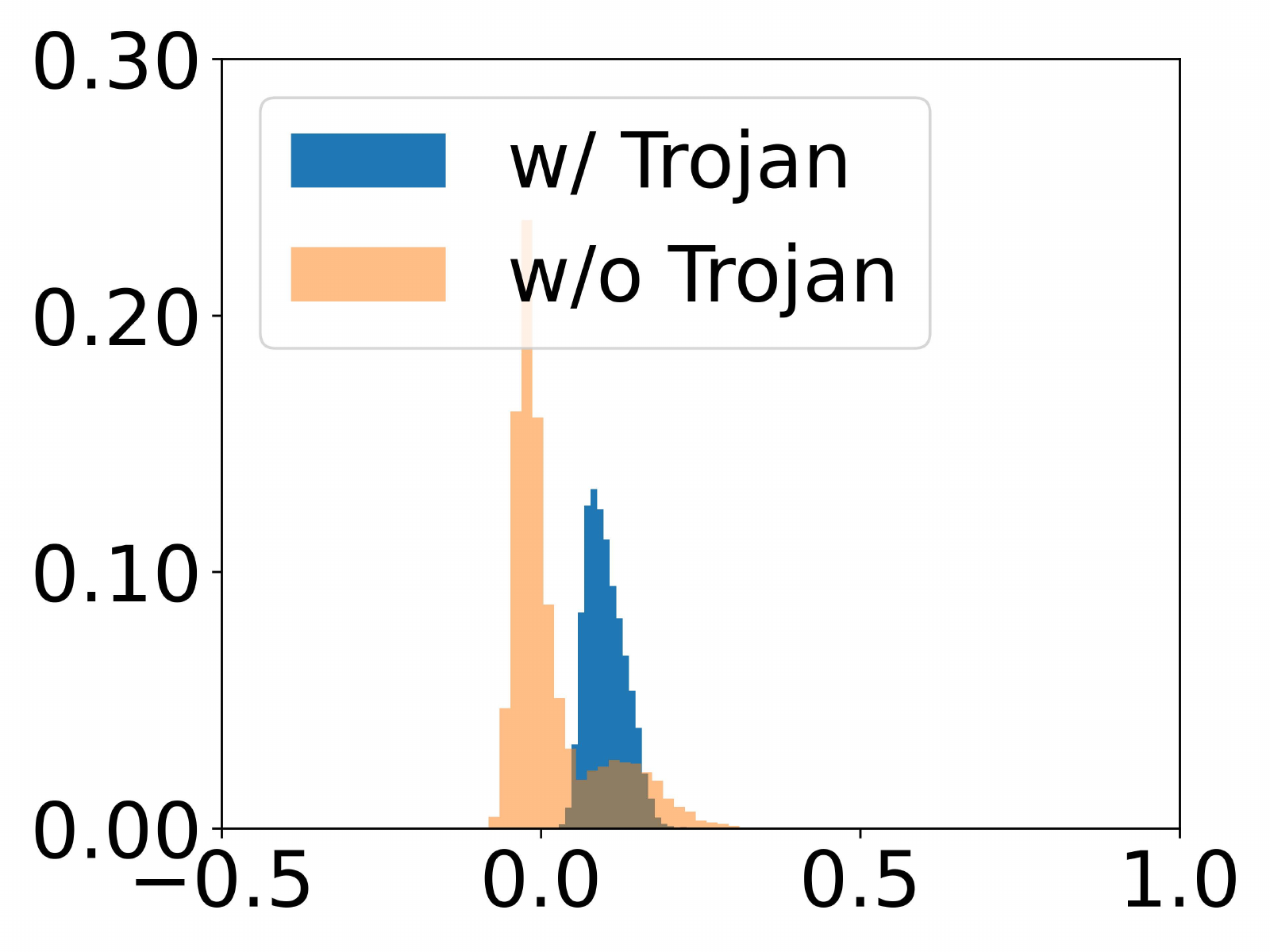}}
    \subfloat[Benign]{\includegraphics[width=0.24\columnwidth]{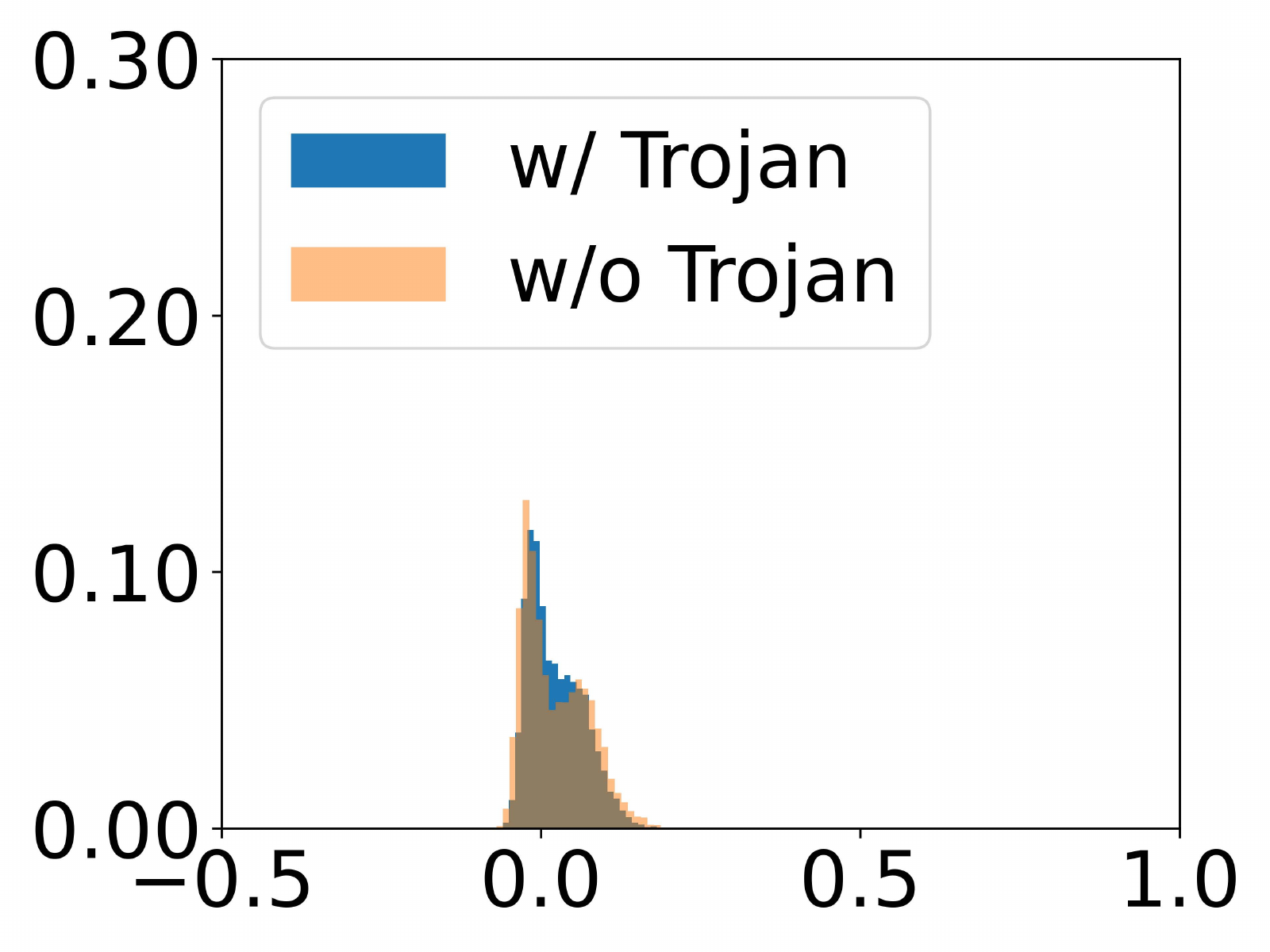}}
    }
    \hfill
    \caption{Comparison of Activation Values on Different CNN Layers in ResNet18 Model.}\label{fig:compare_different_layers}
\end{figure*}

\subsection{Empirical Evidence for \autoref{th:perfect_trojan1} on Other Models}\label{sec:appendix_ee_activation}
\noindent
{\bf Different model architectures.}
To evaluate the linearity of different model architectures, we collect the activation outputs of models with different architectures (i.e., NiN and VGG16). 
Similar to~\autoref{sec:observation} in the main paper, 
we use both benign samples and compromised samples as the input of models and collect their activation outputs.
The results are shown in~\autoref{fig:compare_different_arch}.
The results show that compromised samples always lead to significantly higher activation values than benign samples in different models.
The conclusion is consistent with the linearity theory in~\autoref{sec:observation} of the main paper and proves that our theory can generalize to different model architectures.

\noindent
{\bf Different model layers:}
Besides the linearity on different model architectures, we also evaluate the linearity on different model layers.
\autoref{fig:compare_different_layers} demonstrates the activation outputs of different convolutional layers (i.e., \nth{14} to \nth{17} layers). 
Note that we only show the results on 4 layers due to the space limitation.The results on other layers are similar. 
From the results, we observe that Trojans introduce a large set of high activation values in each layer, leading to the final linearity between input and activation output. The results are consistent with our previous analysis in~\autoref{sec:observation} of the main paper and further confirm that Trojans can introduce linearity at each layer of the DNN model.

\noindent
{\bf Different activation functions.}
To investigate if our theory and \sys can generalize on different activation functions, we train 5 ResNet18 models on CIFAR-10 with 2 common used linear activation functions (i.e., ReLU~\cite{nair2010rectified}, LeakyReLU~\cite{maas2013rectifier}) and 3 non-linear activation functions (i.e., ELU~\cite{clevert2015fast}, Tanhshrink~\cite{Tanhshrink} and Softplus~\cite{zheng2015improving}).
Then we apply \sys to protect these models.
We report the ASR and BA of both protected models and undefended models.
The results are shown in \autoref{tab:different_activations}.
Overall, we find that \sys always achieves a low ASR when using different activation functions, showing the generalization of \sys on different activation functions.
Even with non-linear activation functions, \sys is still effective and we suspect the reason is that even though some activation functions are non-linear, well-trained deep neural networks do fall into the "highly linear" regions. 
The results are also consistent with existing papers~\cite{goodfellow2014explaining}.
\begin{table}[H]
\centering
\scriptsize
\caption{Evaluation Results with Different Activation Functions.}\label{tab:different_activations}
\vspace{0.1cm}
\begin{tabular}{@{}ccrrrrr@{}}
\toprule
\multirow{2}{*}{Activation Function} &  & \multicolumn{2}{c}{Undefended}                   & \multicolumn{1}{c}{} & \multicolumn{2}{c}{NONE}                         \\ \cmidrule(lr){3-4} \cmidrule(l){6-7} 
                                     &  & \multicolumn{1}{c}{BA} & \multicolumn{1}{c}{ASR} & \multicolumn{1}{c}{} & \multicolumn{1}{c}{BA} & \multicolumn{1}{c}{ASR} \\ \midrule
ReLU                                 &  & 94.10\%                & 100.00\%                &                      & 93.62\%                & 1.07\%                  \\
LeakyReLU                            &  & 94.32\%                & 100.00\%                &                      & 93.48\%                & 1.24\%                  \\
ELU                                  &  & 92.99\%                & 99.93\%                 &                      & 91.11\%                & 1.46\%                  \\
Tanhshrink                           &  & 91.68\%                & 99.76\%                 &                      & 90.18\%                & 5.11\%                  \\
Softplus                             &  & 92.81\%                & 100.00\%                &                      & 89.91\%                & 2.07\%                  \\ \bottomrule
\end{tabular}
\end{table}
\subsection{Explaining DP-SGD Defense}\label{sec:explain}

\begin{figure}[]
    \centering
    \footnotesize
    \begin{subfigure}[t]{0.36\columnwidth}
        \centering     
        \footnotesize
        \includegraphics[width=\columnwidth]{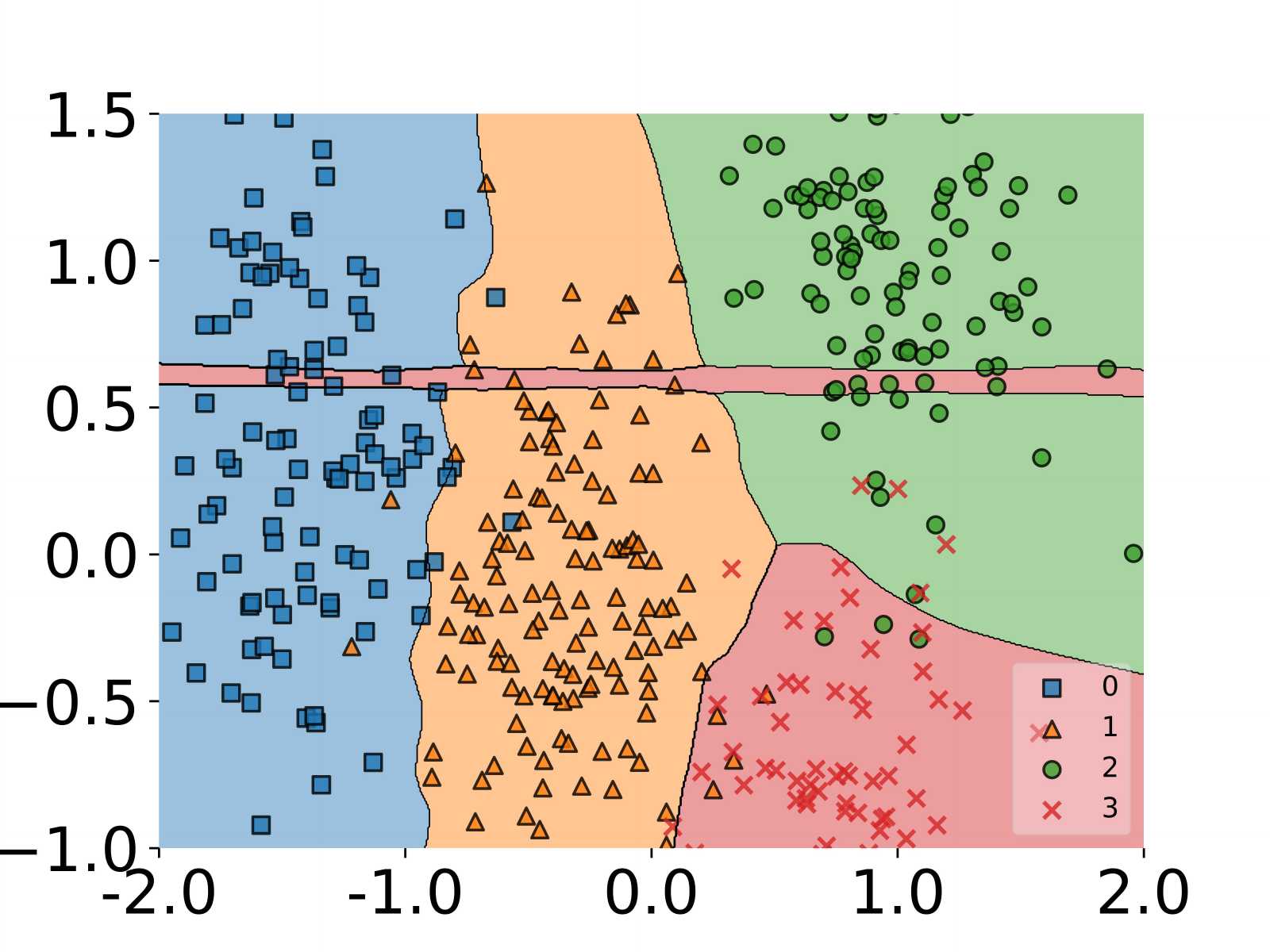}
        \caption{SGD}\label{fig:explain_sgd}
    \end{subfigure}
    \begin{subfigure}[t]{0.36\columnwidth}
        \centering     
        \footnotesize
        \includegraphics[width=\columnwidth]{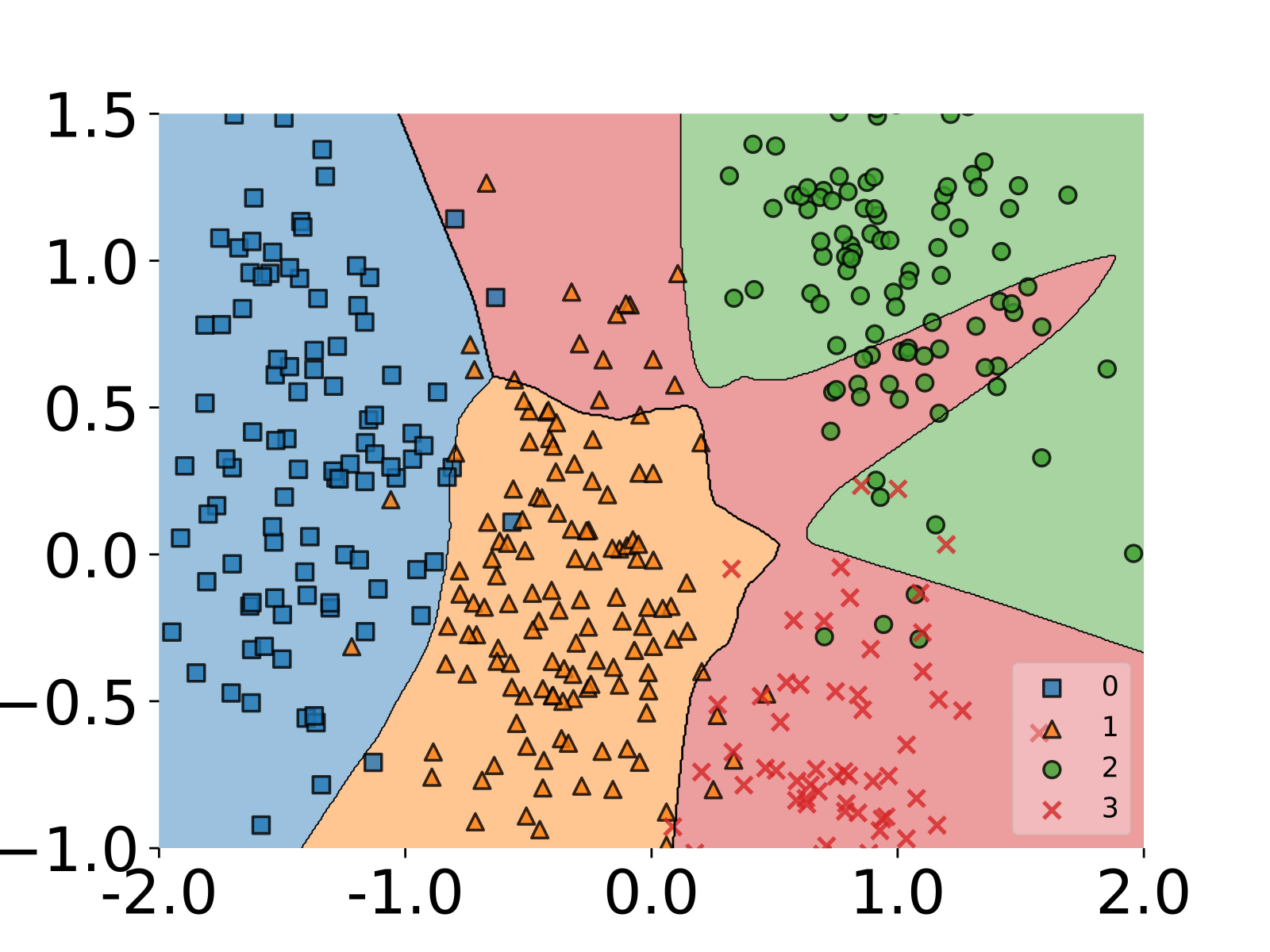}
         \caption{DP-SGD}\label{fig:explain_dpsgd}
    \end{subfigure}
    \caption{Decision Region Generated by SGD and DP-SGD.}\label{fig:comparision_sgd_and_dpsgd}
\end{figure}

DP-SGD~\cite{hong2020effectiveness} improves existing SGD methods by removing the noises added to poisoning training samples and shadows promising results in defending against Trojans.
Here, we explain why it works.
Specifically, we use the same settings with~\autoref{fig:perfect_trojan_relaxation} to train 2 compromised models with vanilla SGD and DP-SGD, and show the comparison results in~\autoref{fig:comparision_sgd_and_dpsgd}. 
Results show that data poisoning can successfully attack the vanilla SGD method.
As a comparison, DP-SGD makes the decision region (red) much more complex, and removes the malicious ``hyperplane'' effects to defense against Trojans.
Recently, Tursynbek et al.~\cite{tursynbek2020robustness} quantitatively measured the curvature of DNN using Curvature Profile~\cite{moosavi2019robustness} and showed that models trained with DP-SGD produce more curved decision boundaries, which is consistent with our results.
By doing so, DP-SGD breaks the ``hyperplane'' Trojans rely on and hence, removes the Trojan effects.
However, this unavoidably affects the accuracy of benign samples.
As shown in \autoref{fig:comparision_sgd_and_dpsgd}, many benign samples got misclassified.

\subsection{Sample Separation}\label{sec:more_details_alg1}

In line 13 of \autoref{alg:training}, we separate the activation values into two clusters via Fisher’s linear discriminant analysis.
In detail, we minimize the variance within clusters \(\sigma_{within}\) and maximize the variance between clusters \(\sigma_{between}\).
The process is implemented by Jenks natural breaks optimization, which is an iterative optimization method that finds the minima/maxima of \(\sigma_{within}/\sigma_{between}\).
The detailed process can be found in \autoref{alg:jenks}.
In line 2 of \autoref{alg:jenks}, it iterates all possible breaks.
In lines 5 to 8, it calculates the value of \(\sigma_{within}\) and \(\sigma_{between}\).
Lines 9 to 12 find the lowest value of \(\sigma_{within}/\sigma_{between}\) and the best separation.

\begin{algorithm}[tb]
 	\caption{Jenks Natural Breaks Optimization}\label{alg:jenks}
    {\bf Input:} %
    \hspace*{0.05in} All Activation Values: \(A_n\)\\
    {\bf Output:} %
    \hspace*{0.05in} Cluster of smaller values: \(B_n\), Cluster of larger values: \(O_n\)
	\begin{algorithmic}[1]
	     \Function {Separation}{$A_n$}
	     
	  \For{{\rm break} \(b\) {\rm in} \(Breaks\)}
	      \State \(B_n^{\prime} = \{x \in A_n \| x \leq b\}\)
	      \State \(O_n^{\prime} = \{x \in A_n \| x \geq b\}\)
	      \State \(\mu_B, \sigma_B = norm(B_n^{\prime})\)
	      \State \(\mu_O, \sigma_O = norm(O_n^{\prime})\)
	      \State \(\sigma_{within}^2 = \sigma_B^2 + \sigma_O^2\)
	      \State \(\sigma_{between}^2 = (\mu_B - \mu_O)^2\)
          \If{ \( \sigma_{within}/\sigma_{between} \leq {\rm lowest} \) }
            \State \(lowest = \sigma_{within}/\sigma_{between}\)
            \State \(O_n = O_n^{\prime}\)
            \State \(B_n = B_n^{\prime}\)
          \EndIf
      \EndFor
         \EndFunction
	\end{algorithmic}
\end{algorithm}

\subsection{Dataset Details}
\label{sec:details_datasets}
The overview of the dataset is shown in~\autoref{tab:dataset}. 
Specifically, we order the datasets with their data sizes and show their dataset names, input size of each sample, the total number of samples, the number of classes and the default Trojan triggers used for generating poisoned data in each column.
Among these datasets, MNIST~\cite{lecun1998gradient} is widely used for digit classification tasks.
The GTSRB~\cite{stallkamp2012man} dataset is used for traffic sign recognition tasks in the self-driving scenario.
TrojAI~\cite{karra2020trojai} contains the images created by compositing a synthetic traffic sign, with a random background image from the KITTI dataset~\cite{geiger2013vision}.
Other datasets (i.e., CIFAR-10~\cite{krizhevsky2009learning} and ImageNet-10\footnote{https://github.com/fastai/imagenette}) are built for recognizing general objects (e.g., animals, plants and handicrafts). The default triggers (\autoref{fig:triggers} in the main paper) used for each dataset are shown in the last column of~\autoref{tab:dataset}.
All datasets used in the experiments are with MIT license. They are open-sourced and do not contain any personally identifiable
information or offensive content.

\subsection{Attack Details}\label{sec:details_attack}
We first evaluate the performances of \sys against BadNets~\cite{gu2017badnets} on two different settings: single target attack and label specific attack. For the single target attack, we set the label whose index is 0, 0, 1 and 1 as the target label for MNIST, CIFAR-10, GTSRB and ImageNet-10, respectively.
For label specific attack, the target label of each sample is the label whose index is (the label index of this sample plus 1)\%(the number of classes in the dataset).
Then, we evaluate the defense against the label-consistent attack~\cite{turner2019label} and the natural Trojan attack~\cite{liu2019abs}.
We use the same implementation and parameters in original papers to achieve these attacks and compare \sys with other defense methods.
Notice that for the label-consistent attack, the official github repository\footnote{https://github.com/MadryLab/label-consistent-backdoor-code} only provides poisoned CIFAR-10 datasets, and the code for training GAN and generating poisoned samples are not released.
Therefore, we only evaluate \sys on CIFAR-10.
For defending against the hidden trigger Trojan attack~\cite{saha2020hidden}, we follow the parameter settings in original paper and use a pair of image categories (i.e., randomly selected from ImageNet dataset in the previous work~\cite{saha2020hidden}) for testing.

\begin{table}[]
    \centering
    \scriptsize
    \caption{Overview of Datasets.}\label{tab:dataset}
    \vspace{0.1cm}
\begin{tabular}{@{}ccccc@{}}
\toprule
Name        & Input Size & Samples & Classes & Trigger         \\ \midrule
MNIST       & 28*28*1    & 60000   & 10      & Single Pixel    \\
GTSRB       & 32*32*3    & 39209   & 43      & Static          \\
CIFAR-10    & 32*32*3    & 50000   & 10      & Dynamic         \\
ImageNet-10 & 224*224*3  & 9469    & 10      & Watermark       \\
TrojAI      & 224*224*3  & 125000  & 5-25    & Natural Trojans \\ \bottomrule
\end{tabular}
\end{table}

\subsection{Resistance to More Attacks}\label{sec:resistance_more_attacks}

In this section, we evaluate the resistance of \sys to more attacks. Four state-of-the-art poisoning based Trojan attacks (WaNet~\cite{nguyen2021wanet}, SIG attack~\cite{barni2019new}, Filter attack~\cite{liu2019abs} and Blend attack~\cite{chen2017targeted}) are included in the experiments. The dataset and the network used are CIFAR-10 and ResNet18. We report the BA and ASR of undefended model, and the model trained with NAD~\cite{li2021neural}, ABL~\cite{li2021anti} and \sys. Results in \autoref{tab:more_attacks} demonstrates \sys has better performance than baseline methods (i.e., NAD and ABL). On average, \sys has 0.98\% ASR and 
92.52\% BA.
The results indicate that our method is resistance to various Trojan attacks.

\begin{table}[]
\centering
\scriptsize
\caption{Results on More Attacks.}
\vspace{0.1cm}
\setlength\tabcolsep{3pt}
\label{tab:more_attacks}
\begin{tabular}{@{}ccclcccccccccccc@{}}
\toprule
\multirow{2}{*}{Dataset}  & \multirow{2}{*}{Network}  & \multicolumn{2}{c}{\multirow{2}{*}{Attack}} &  & \multicolumn{2}{c}{Undefended} &  & \multicolumn{2}{c}{NAD} &  & \multicolumn{2}{c}{ABL} &  & \multicolumn{2}{c}{NONE} \\ \cmidrule(lr){6-7} \cmidrule(lr){9-10} \cmidrule(lr){12-13} \cmidrule(l){15-16} 
                          &                           & \multicolumn{2}{c}{}           &  & BA             & ASR           &  & BA          & ASR       &  & BA          & ASR       &  & BA          & ASR        \\ \midrule
\multirow{4}{*}{CIFAR-10} & \multirow{4}{*}{ResNet18} & \multicolumn{2}{c}{WaNet}                   &  & 94.39\%        & 96.71\%       &  & 88.81\%     & 1.17\%    &  & 90.79\%     & 2.68\%    &  & \textbf{92.24\%}     & \textbf{0.69\%}     \\
                          &                           & \multicolumn{2}{c}{SIG}                     &  & 94.34\%        & 99.08\%       &  & 88.26\%     & 1.42\%    &  & 91.44\%     & 1.29\%    &  & \textbf{93.79\%}     & \textbf{1.08\%}     \\
                          &                           & \multicolumn{2}{c}{Filter}                  &  & 91.08\%        & 99.34\%       &  & 87.91\%     & 4.38\%    &  & 88.46\%     & 2.24\%    &  & \textbf{89.87\%}     & \textbf{1.20\%}     \\
                          &                           & \multicolumn{2}{c}{Blend}                   &  & 94.62\%        & 99.86\%       &  & 88.24\%     & 1.58\%    &  & 92.72\%     & 1.70\%    &  & \textbf{94.21\%}     & \textbf{0.93\%}     \\ \bottomrule
\end{tabular}
\end{table}

\subsection{Sensitivity to Configurable Parameters}\label{sec:RQ3}
\sys has a few configurable parameters that may affect its
performance: learning rate in training, resetting fraction, number of neurons in each layer used to detect malicious samples (selection threshold) and different thresholds used for the identification of compromised neurons.
We vary the configurable parameters in \sys independently and evaluate the impact of each.
The setting of dataset, models and attack type is the same as evaluation in~\autoref{sec:RQ2} of the main paper.
We use 5\% poisoning rate, 3*3 trigger size as the default attack setting.

\begin{figure}[]
     \centering
     \footnotesize
     \begin{subfigure}[t]{0.44\columnwidth}
         \centering     
         \footnotesize
         \includegraphics[width=\columnwidth]{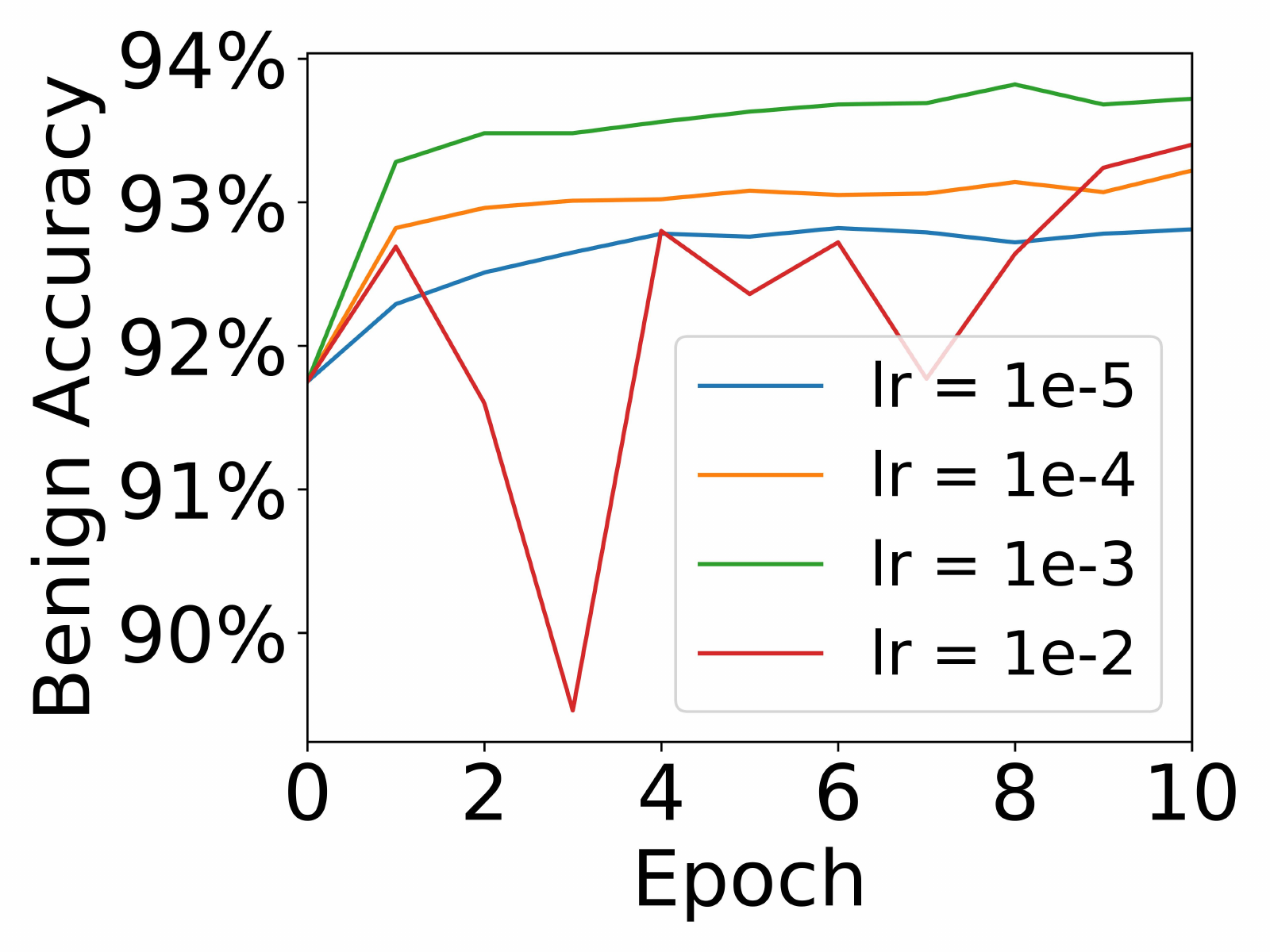}
         \caption{Benign Accuracy}
         \label{fig:lr_ba}
     \end{subfigure}
     \begin{subfigure}[t]{0.44\columnwidth}
         \centering
         \footnotesize
         \includegraphics[width=\columnwidth]{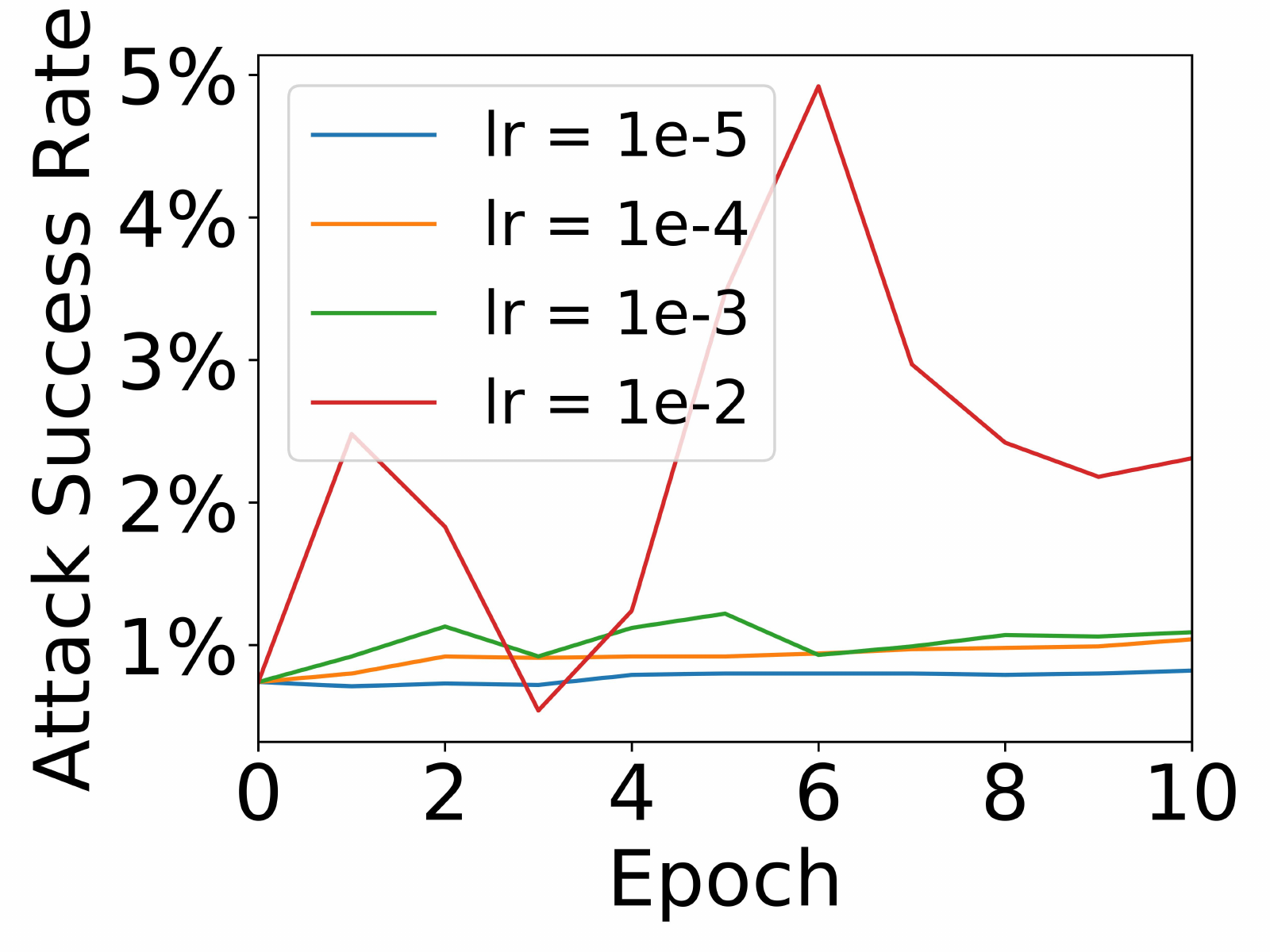}
     	   \caption{Attack Success Rate}
         \label{fig:lr_asr}
     \end{subfigure}
     \caption{Evaluation Results with Different Learning Rates.}
     \label{fig:lr}
\end{figure}

{\bf Learning rate.}
Learning rate usually affects the accuracy and convergence speed of the model during the training process.
To understand how the learning rate impacts the model deployed with \sys, we choose learning rates from 0.01 to 0.00001 and then measure the BA and ASR of models using different learning rates in the training process. 
The results are shown in \autoref{fig:lr}.

Overall, as shown in \autoref{fig:lr_ba}, using a larger learning rate makes the convergence process faster and the BA lower, except for using the learning rate 0.01.
This is because using a larger learning rate can update the weights quickly, but a too large learning rate makes it difficult to find the local optimum and decrease the BA.

In addition, in \autoref{fig:lr_asr}, we find that the final ASR is decreased with the decrease of learning rate after the model is converged.
Using learning rate 0.00001 finally achieves the lowest ASR.
The reason is that increasing the learning rate tends to make the model skip the local optimal value and get a more likely worse value.

Therefore, combining the results in these 2 subfigures, we choose the learning rate 0.001 as the default setting in \autoref{sec:RQ1} because using 0.001 achieves the best BA and ASR.
The epoch number is set to 5 because ASR is not decreased after 5 epochs and the BA is already good.

\begin{figure}[]
     \centering
     \footnotesize
     \begin{subfigure}[t]{0.44\columnwidth}
         \centering     
         \footnotesize
         \includegraphics[width=\columnwidth]{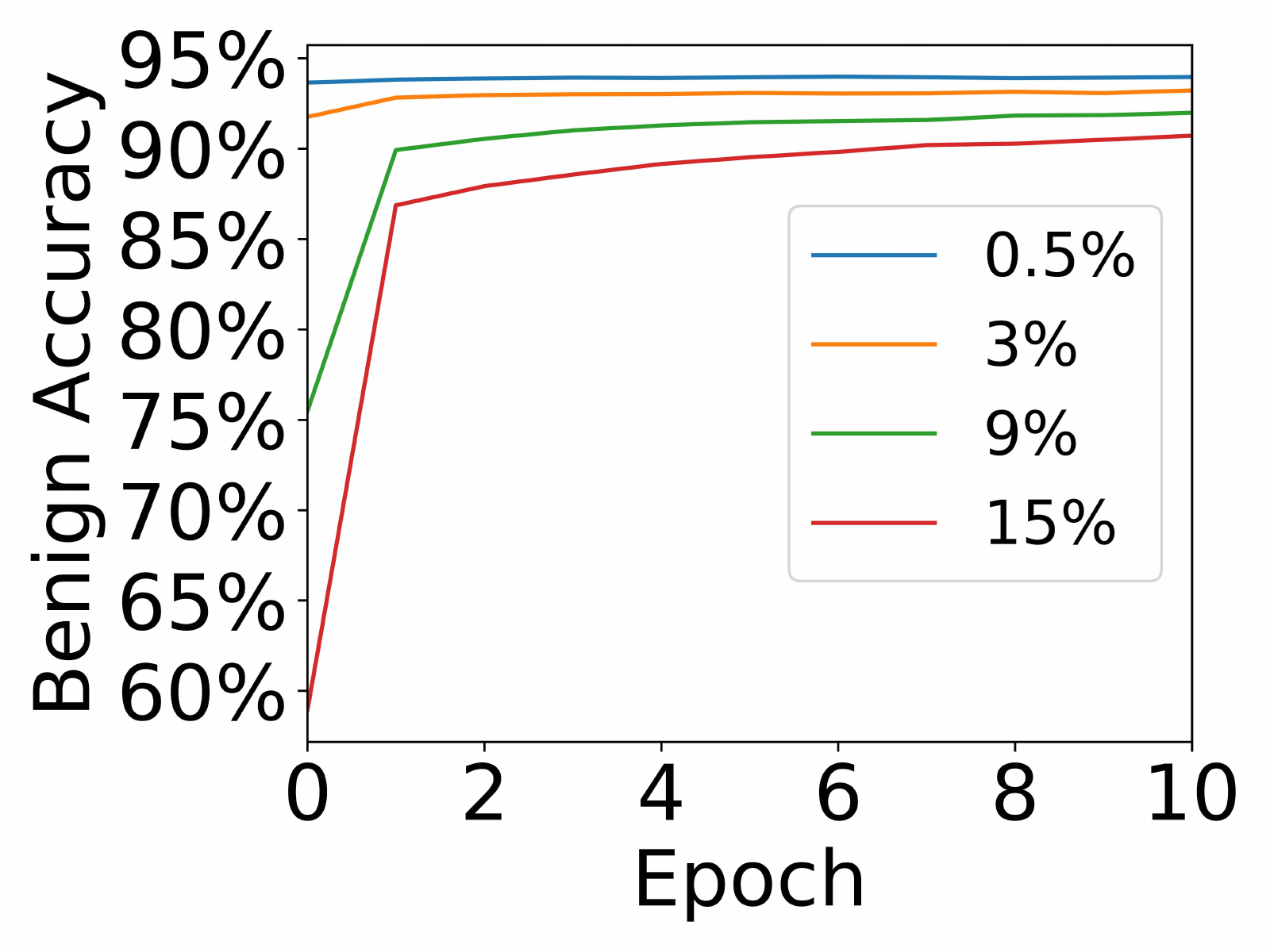}
         \caption{Benign Accuracy}
         \label{fig:r_ba}
     \end{subfigure}
     \begin{subfigure}[t]{0.44\columnwidth}
         \centering     
         \footnotesize
         \includegraphics[width=\columnwidth]{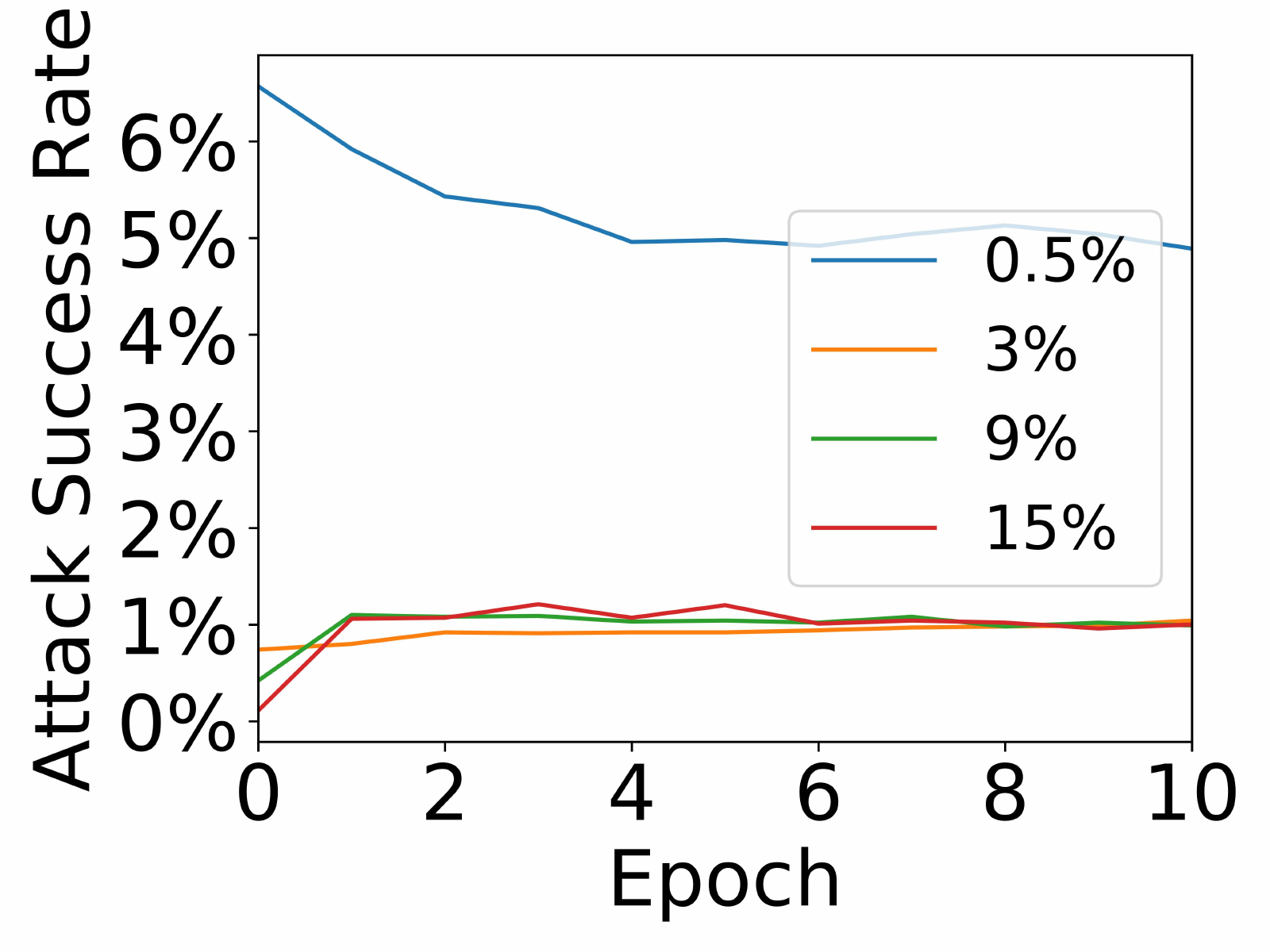}
     	   \caption{Attack Success Rate}
         \label{fig:r_asr}
     \end{subfigure}
     \caption{Evaluation Results with Different Resetting Fractions.}
     \label{fig:r}
\end{figure}

{\bf Resetting fraction.}
Resetting fraction measures the number of neurons that are reset by \sys.
Specifically, \sys first sorts the probabilities that the neuron has activation values larger than 0 and then resets the neurons whose probabilities is in top \(r_1\%\) in each layer.
Using a smaller resetting fraction makes \sys to detect compromised neurons more conservatively (only labeling and resetting the most likely compromised neurons).
To measure the effect of resetting fraction on defense performance of \sys, we obtain the BA and ASR of models at different resetting fractions from 0.5\% to 15\%.
The results are shown in \autoref{fig:r}, where the legend shows different resetting fraction values.

From the results in \autoref{fig:r_ba}, it is obvious that when we use a larger resetting fraction and reset more neurons, the final BA is lower.
The reason is that after we reset neurons, some good features learned by the model are lost, which decreases the final BA.
When we reset more neurons (i.e., using a larger resetting fraction), the model loses more high quality features and decreases more BA.
Therefore, to avoid losing too much BA, the resetting fraction is recommended to be small.

\begin{minipage}[htbp]{\textwidth}
    \begin{minipage}{0.55\textwidth}
        \centering
          \scriptsize
          \setlength\tabcolsep{3pt}
          \captionof{table}{Results on Different Selection Thresholds.}
          \label{tab:different t}
          \scalebox{1}{
            \begin{tabular}{@{}cccccc@{}}
            \toprule
            \multirow{2}{*}{Number of Neurons} & \multicolumn{2}{c}{Single Target Attack} &  & \multicolumn{2}{c}{Label Specific Attack} \\ \cmidrule(lr){2-3} \cmidrule(l){5-6} 
                                               & BA                  & ASR                &  & BA                  & ASR                 \\ \midrule
            top 1                              & 93.11\%             & 1.03\%             &  & 93.31\%             & 0.96\%              \\
            top 0.5\%                          & 93.10\%             & 1.07\%             &  & 93.32\%             & 0.96\%              \\
            top 1\%                            & 93.13\%             & 1.07\%             &  & 93.37\%             & 0.95\%              \\
            top 10\%                           & 93.11\%             & 1.06\%             &  & 93.29\%             & 1.04\%              \\
            top 30\%                           & 93.14\%             & 1.04\%             &  & 93.26\%             & 41.07\%             \\
            top 50\%                           & 93.08\%             & 1.04\%             &  & 93.18\%             & 60.12\%             \\
            top 100\%                          & 93.05\%             & 1.07\%             &  & 93.14\%             & 71.78\%             \\ \bottomrule
            \end{tabular}}
            \end{minipage}
    \begin{minipage}{0.4\textwidth}
        \centering
          \scriptsize
          \captionof{table}{Results on Different $\lambda_l$ and $\lambda_h$.}
          \label{tab:different_lambda_l_h}
          \scalebox{1}{
            \begin{tabular}{@{}crrcrr@{}}
            \toprule
            $\lambda_h$ & \multicolumn{1}{c}{BA} & \multicolumn{1}{c}{ASR} & \multicolumn{1}{r}{$\lambda_l$} & \multicolumn{1}{c}{BA} & \multicolumn{1}{c}{ASR} \\ \midrule
            0.3         & 93.22\%                & 1.07\%                  & 0.1                             & 93.18\%                & 1.13\%                  \\
            0.5         & 93.05\%                & 1.14\%                  & 0.3                             & 93.08\%                & 0.99\%                  \\
            0.7         & 93.13\%                & 1.11\%                  & 0.5                             & 93.18\%                & 1.11\%                  \\
            0.9         & 93.12\%                & 1.11\%                  & 0.7                             & 93.11\%                & 1.03\%                  \\ \bottomrule
            \end{tabular}}
            \end{minipage}
\vspace{0.3cm}
\end{minipage}
Furthermore, in \autoref{fig:r_asr}, we find that different resetting fractions do not affect the ASR of models after a certain threshold (i.e., 3\%).
Because when the resetting fraction is large, \sys can successfully detect almost all compromised neurons.
Increasing the resetting fraction does not help \sys to detect more compromised neurons.

Based on the above conclusions, we set the default resetting fraction as 3\% because using resetting fraction 3\% requires changing fewer neurons, achieving high BA and low ASR.

{\bf Selection threshold.}
When detecting poisoning samples, we only use the neurons whose compromised values are larger than the values of a portion of neurons in the same layer and we call this portion as selection threshold.
To fully understand the impact of this threshold, we vary the threshold from 1 neuron to 100\% neurons in the dataset and collect the corresponding BA and ASR under different attack settings.
We test the single target BadNets attack and the label specific BadNets attack.
We then show the results in \autoref{tab:different t}, where the first column shows the threshold and the following columns show the results against the BadNets.

As the results show, when we increase the selection threshold, the ASR of the label specific BadNets attack significantly increases when the threshold is larger than 10\%.
This is because only a few neurons in the model are compromised.
If the selection threshold is larger than the number of compromised neurons, \sys chooses many benign neurons to identify whether a sample is malicious or not, which introduces more noise and reduces the detection accuracy because benign neurons are not sensitive to Trojan behavior.
Furthermore, the label specific BadNets attack specifies many different labels as target labels, making the attack stealthy and detecting the attack more difficult.
Therefore, with the increase of the selection threshold, the defense performance becomes worse.

However, we observe that the ASR of the single target standard Trojan attack is not correlated with the selection threshold, showing the robustness of \sys to selection threshold against the single target Trojan attack.
This is due to the fact that the single target BadNets attack only focuses on one 
label, making the malicious behavior more obvious, thus reducing the impact of introduced noise and still achieving a low ASR.

For the BA, we find that the BA against both the single target BadNets attack and the label specific BadNets attack is stable.
Although using a lower selection threshold may allow \sys to filter out malicious samples conservatively (i.e., only use the most likely compromised neurons to detect malicious samples), enabling \sys to train the model on most of the data and achieve good BA results.
Choosing a higher selection threshold does not decrease the BA significantly.
Because considering there are a large number of benign samples in the dataset, even a higher selection threshold introduces more benign neurons (i.e., noise) to identify malicious samples and reduces the number of benign samples for finetuning, \sys still has enough benign samples for training and achieves similar BA results as using low selection thresholds.

Therefore, considering both BA and ASR, we set the selection threshold as 10\% to avoid the ASR increasing significantly.

{\bf Parameters in compromised neurons identification.}
As mentioned in \autoref{sec:defense} of the main paper, we use an alternative implementation to evaluate our design.
We first obtain two clusters of samples according to their final layer probability outputs (the value in the probability vector).
Subsequently, we classify the samples whose probability values are lower than a threshold \(\lambda_l\) to the first cluster (i.e., low confidence samples) and classify the samples whose probability values are higher than \(\lambda_h\) to the second cluster (i.e., low confidence samples).
Then, we use the gap between two clusters to measure the linearity of each neuron. If a neuron has high linearity (i.e., top \(r_1\) in a layer), then we consider it as a compromised neuron.
In this process, \(\lambda_h\) and \(\lambda_l\) determine the selection of the high confidence

\begin{table}[]
\centering
\scriptsize
\captionof{table}{Comparisons on Efficiency.}
\label{tab:efficiency}
\scalebox{1}{
\begin{tabular}{@{}ccc@{}}
            \toprule
            Method          & Runtime & Overhead \\ \midrule
            Native training & 2898.4s & N/A      \\
            AC              & 4459.7s & 53.86\%  \\
            ABL             & 3197.4s & 10.31\%  \\
            NONE            & 3149.7s & 8.60\%   \\ \bottomrule
            \end{tabular}}
\end{table}
samples and the low confidence samples that affect the defense performance of \sys.
Therefore, to fully understand the impacts of them, We vary \(\lambda_h\) and \(\lambda_l\) values, and obtain the corresponding ASR and BA.
By default, we use \(\lambda_h\) as 0.9 and \(\lambda_l\) to be 0.1 when the other parameter is changing.

\autoref{tab:different_lambda_l_h} shows the results with different \(\lambda_h\) and \(\lambda_l\) settings.
The results indicate that there is no obvious correlation between the performance of \sys and parameters (i.e., \(\lambda_l\) and \(\lambda_h\)).
As the results show, the ASR of models is always around 1.11\% under different parameter settings.
And the difference between the highest BA and the lowest BA is 0.17\% which is quite small.
Therefore, \sys is not sensitive to \(\lambda_l\) and \(\lambda_h\), which improves the usability of \sys.

\subsection{Efficiency}\label{sec:efficiency}

We compare the total training time of native training, AC~\cite{chen2018detecting}, ABL~\cite{li2021anti}, and our method on the CIFAR-10 dataset with ResNet18. 
The results are shown in \autoref{tab:efficiency}.
The epoch number (i.e., 100) and batch size (i.e., 128) for different methods are the same. 
The ASR and BA are consistent with results in \autoref{tab:standard backdoor attack}.
We run each method with five trails and report the average time. All methods are run on the same device specified in \autoref{sec:evaluation} of the main paper.
Thus, our method is efficient.

\subsection{Adaptive Attack}\label{sec:appendix_adaptive_attack}

In this paper, we assume that attackers can poison the training data but have no control over the training procedure, e.g., the training algorithm, code, and hardware.
This is consistent with existing work~\cite{hong2020effectiveness,chen2018detecting,li2021anti,huang2022backdoor}. 
It is hard for attackers to conduct adaptive attacks under the threat
model because they can not directly control the training of the model, instead, \sys will be in charge of the training process. 
Therefore, we relax the threat model and consider the adaptive attacker in a code-poisoning attack~\cite{bagdasaryan2021blind}, which requires extra capability from the adversary, i.e., modifying the training procedure.

In the considered adaptive code-poisoning attacks, the adversary goal is to train a Trojaned model with low linearity and try to evade the defense of \sys.
However, under our threat model, the adversary can only poison the data but cannot modify the training process of \sys, which makes reducing the model's internal linearity almost impossible.
Therefore, we relax the threat model for attackers and allow the attacker to control the training process of the model. 
We also assume the defender can access both the training data and the trained model. The defender tries to use \sys to eliminate Trojans injected in the model trained by the attacker.

Then, we design an adaptive loss that minimizes the activation difference between benign samples and corresponding Trojan samples to achieve attack goals.
The adaptive loss is defined in \autoref{eq:adaptive}, where \(\bm{x}\) is benign sample and \(\tilde{\bm{x}}\) is the corresponding Trojan sample (i.e., the sample obtained by pasting trigger on \(\bm{x}\)).
\begin{equation}\label{eq:adaptive}
     \mathcal{L}\left(F_\theta(\bm{x}), y\right) + \mathcal{L}\left(F_\theta(\tilde{\bm{x}}), y_t\right) + \alpha \sum (I_i({\bm{x}}) - I_i(\tilde{\bm{x}}))^2
\end{equation}
\(y\) and \(y_t\) are the label of benign sample \(\bm{x}\) and target label respectively.
\(F_{\theta}\) donates the final prediction of the model. \(\mathcal{L}\) means the Cross-Entropy criterion. Meanwhile, \(I_i\) is the feature on the i-th layer, and \(\alpha\)
is the weight that controls the influence of the third loss item. 
By design, the loss function minimizes the distance between activation values of benign samples and the corresponding Trojan samples, making the Trojan decision region more curve and complex.
Trojan models trained with the adaptive loss should have low linearity and may evade the detection of \sys.

\begin{table}[]
\centering
\scriptsize
\caption{Adaptive Attack.}\label{tab:adaptive}
\begin{tabular}{@{}ccccccc@{}}
\toprule
\multirow{2}{*}{$\alpha$} &  & \multicolumn{2}{c}{Undefended} &  & \multicolumn{2}{c}{NONE} \\ \cmidrule(lr){3-4} \cmidrule(l){6-7} 
                          &  & BA            & ASR            &  & BA          & ASR        \\ \midrule
1e-4                      &  & 90.06\%       & 100.00\%       &  & 88.48\%     & 67.89\%    \\
1e-3                      &  & 89.53\%       & 99.97\%        &  & 87.92\%     & 76.78\%    \\
1e-2                      &  & 89.03\%       & 99.91\%        &  & 86.50\%     & 86.20\%    \\
1e-1                      &  & 88.72\%       & 99.98\%        &  & 85.71\%     & 94.92\%    \\ \bottomrule
\end{tabular}
\end{table}
To measure whether the adaptive attack works, we first train a benign model and then fine-tune that model using adaptive loss when attackers use poisoned data to attack the model.
The Trojan trigger we use in the attack is the watermarking trigger and the model is VGG16.
The results are shown in ~\autoref{tab:adaptive}.
The results show that \sys does not always achieve good defense against adaptive attacks. 
For example, when \(\alpha = 1e-1\), the BA and the ASR of \sys is 85.71\% and 94.92\%, respectively.
However, the  BA and ASR of the model trained with \sys are lower than that of the undefended model, showing that \sys helps in training a better model.

\subsection{Comparison with DBD}\label{sec:compare_to_dbd}
Besides existing defenses compared in \autoref{sec:RQ1} (i.e., DP-SGD~\cite{hong2020effectiveness}, NAD~\cite{li2021neural}, AC~\cite{chen2018detecting}, ABL~\cite{li2021anti}), we also compare \sys with another training time defense DBD~\cite{huang2022backdoor}.
DBD defends backdoor attacks by decoupling the end-to-end training process into three stages, i.e., self-supervised learning for the backdoor, supervised training for the fully-connected layers, and semi-supervised fine-tuning of the whole model.
We use six different attacks (i.e., BadNets~\cite{gu2017badnets}, Label-consistent~\cite{turner2019label}, Blend~\cite{chen2017targeted}, SIG~\cite{barni2019new}, Filter~\cite{liu2019abs}, WaNet~\cite{nguyen2021wanet}) and the CIFAR-10 dataset.
We report the BA and ASR of the native training, DBD, and \sys in \autoref{tab:compare_to_dbd}.
The average runtime of DBD and \sys are 18,988.4s and 3,149.7s, respectively.
For all attacks, our method achieves higher BA than DBD.
In addition, in five of six attacks, the ASR of \sys is lower than that of DBD.
The results show that our method is more effective and efficient than DBD.

\begin{table}[H]
\centering
\scriptsize
\caption{Comparison to DBD~\cite{huang2022backdoor}.}
\label{tab:compare_to_dbd}
\vspace{0.1cm}

\setlength\tabcolsep{3pt}
\begin{tabular}{@{}cccccccccc@{}}
\toprule
\multirow{2}{*}{Attack} &  & \multicolumn{2}{c}{Undefended} &  & \multicolumn{2}{c}{DBD} &  & \multicolumn{2}{c}{NONE} \\ \cmidrule(lr){3-4} \cmidrule(lr){6-7} \cmidrule(l){9-10} 
                        &  & BA            & ASR            &  & BA          & ASR       &  & BA          & ASR        \\ \midrule
BadNets                 &  & 94.10\%       & 100.00\%       &  & 91.24\%     & 1.25\%    &  & \bf{93.62\%}     & \bf{1.07\%}     \\
Label-consistent        &  & 94.73\%       & 83.42\%        &  & 91.08\%     & \bf{1.87\%}    &  & \bf{94.01\%}     & 2.14\%     \\
Blend                   &  & 94.62\%       & 99.86\%        &  & 92.03\%     & 1.96\%    &  & \bf{94.21\%}     & \bf{0.93\%}     \\
SIG                     &  & 94.34\%       & 99.08\%        &  & 91.55\%     & 1.51\%    &  & \bf{93.79\%}     & \bf{1.08\%}     \\
Filter                  &  & 91.08\%       & 99.34\%        &  & 88.75\%     & 1.42\%    &  & \bf{89.87\%}     & \bf{1.20\%}     \\
WaNet                   &  & 94.39\%       & 96.71\%        &  & 90.98\%     & 0.95\%    &  & \bf{92.24\%}     & \bf{0.69\%}     \\ \bottomrule
\end{tabular}
\end{table}

\subsection{Comparison to More Defenses on Natural Trojan}\label{sec:compare_to_more_natural_trojan}
Besides the results of comparison to DP-SGD on natural Trojan (\autoref{sec:RQ1}), we compare \sys with more training-time defenses (i.e., DP-SGD~\cite{hong2020effectiveness}, NAD~\cite{li2021neural}, AC~\cite{chen2018detecting}, ABL~\cite{li2021anti}, DBD~\cite{huang2022backdoor}) on natural Trojan~\cite{liu2019abs}.
The dataset used here is CIFAR-10, and DNNs are NiN and VGG16.
As shown in \autoref{tab:compare_to_more_natural_trojan}, the average ASR of \sys is 33.07\%, 2.41 times lower than the undefended model.
However, the average ASR of DP-SGD, NAD, AC, ABL, and DBD are 75.4\%, 80.43\%, 77.45\%, 79.32\%, 77.99\%, respectively.
The results demonstrate all existing methods have high ASR when facing natural backdoors, while our method can reduce the ASR significantly.

\begin{table}[H]
\centering
\scriptsize
\caption{Comparisons to More Defenses on Natural Trojan.}
\vspace{0.1cm}

\label{tab:compare_to_more_natural_trojan}
\setlength\tabcolsep{1.5pt}
\begin{tabular}{@{}cccccccccccccccccccccc@{}}
\toprule
\multirow{2}{*}{Network} &  & \multicolumn{2}{c}{Undefended} &  & \multicolumn{2}{c}{DP-SGD} &  & \multicolumn{2}{c}{NAD} &  & \multicolumn{2}{c}{AC} &  & \multicolumn{2}{c}{ABL} &  & \multicolumn{2}{c}{DBD} &  & \multicolumn{2}{c}{NONE} \\ \cmidrule(lr){3-4} \cmidrule(lr){6-7} \cmidrule(lr){9-10} \cmidrule(lr){12-13} \cmidrule(lr){15-16} \cmidrule(lr){18-19} \cmidrule(l){21-22} 
                         &  & BA             & ASR           &  & BA           & ASR         &  & BA         & ASR        &  & BA         & ASR       &  & BA         & ASR        &  & BA         & ASR        &  & BA          & ASR        \\ \midrule
NiN                      &  & 91.02\%        & 87.62\%       &  & 39.19\%      & 87.22\%     &  & 80.75\%    & 88.10\%    &  & 83.85\%    & 88.28\%   &  & 86.28\%    & 86.33\%    &  & 86.27\%    & 87.54\%    &  & \bf{86.94\%}     & \bf{34.21\%}    \\
VGG16                    &  & 90.78\%        & 71.88\%       &  & 53.40\%      & 63.58\%     &  & 85.20\%    & 72.76\%    &  & 85.69\%    & 66.67\%   &  & \bf{86.46\%}    & 72.32\%    &  & 86.38\%    & 68.45\%    &  & 81.83\%     & \bf{31.49\%}    \\ \bottomrule
\end{tabular}
\end{table}

\subsection{Generalization to Larger Models}\label{sec:generalization_larger_models}
To study \sys's generalization to larger models, we report its BA and ASR on ResNet34~\cite{he2016deep} and Wide-ResNet-16 (WRN16)~\cite{zagoruyko2016wide}.
The results of two baseline methods (i.e., NAD~\cite{li2021neural} and ABL~\cite{li2021anti}) are also reported.
The dataset used is CIFAR-10.
The runtime overhead of \sys on ResNet34 and WRN16 are 10.15\% and 9.73\%, respectively.
For both two models, \sys achieves higher BA and lower ASR than NAD and ABL.
For ResNet34, the BA of \sys is 2.47\% and 2.84\% higher than NAD and ABL.
The ASR of \sys for ResNet34 is also 1.45\% and 0.19\% lower than that of NAD and ABL.
The results show that our method is scalable to larger models.

\subsection{Generalization to Larger Datasets}\label{sec:generalization_to_larger_datasets}
To evaluate the generalization of \sys to larger datasets, we report the performance (i.e., BA, ASR, and Runtime) of native training and \sys on a ImageNet subset (200 classes with 100k images for training and 10k images for testing) from Li et al.~\cite{li2021invisible}.
The results can be found in \autoref{tab:generalization_to_larger_datasets}.
\sys achieves low ASR (i.e., 1.98\%, 50.32 times lower than Native Training) with a high BA (i.e., 1.66\% lower than native training).
In addition, the overheads compared with native training is 13.86\%.

\begin{minipage}[htbp]{\textwidth}
    \begin{minipage}{0.55\textwidth}
        \centering
          \scriptsize
          \captionof{table}{Generalization to Larger Models.}
          \label{tab:generalization_to_larger_models}
          \setlength\tabcolsep{3pt}
          \scalebox{1}{
            \begin{tabular}{@{}cccccccccc@{}}
            \toprule
            \multirow{2}{*}{Networks} &  & \multicolumn{2}{c}{NAD}    &  & \multicolumn{2}{c}{ABL}    &  & \multicolumn{2}{c}{Ours}   \\ \cmidrule(lr){3-4} \cmidrule(lr){6-7} \cmidrule(l){9-10} 
                                      &  & BA           & ASR         &  & BA           & ASR         &  & BA           & ASR         \\ \midrule
            ResNet34                  &  & 90.54\% & 2.67\% &  & 90.17\% & 1.41\% &  & \bf{93.01\%} & \bf{1.22\%} \\
            WRN16                  &  & 86.73\% & 5.96\% &  & 84.70\% & 5.04\% &  & \bf{88.28\%} & \bf{3.88\%} \\ \bottomrule
            \end{tabular}}
            \end{minipage}
    \begin{minipage}{0.42\textwidth}
        \centering
          \scriptsize
          \captionof{table}{Generalization to Larger Datasets.}
          \label{tab:generalization_to_larger_datasets}
          \scalebox{1}{
        \begin{tabular}{@{}cccc@{}}
\toprule
Method          & BA      & ASR     & Runtime \\ \midrule
Native Training & 85.12\% & 99.65\% & 23.8h   \\
NONE            & 83.46\% & 1.98\%  & 27.1h   \\ \bottomrule
\end{tabular}}
        \end{minipage}
\vspace{0.3cm}
\end{minipage}

\end{document}